\PassOptionsToPackage{table,xcdraw}{xcolor}
\documentclass[a4paper]{article}

\usepackage[pages=all, color=black, position={current page.south}, placement=bottom, scale=1, opacity=1, vshift=5mm]{background}
\SetBgContents{
	\tt This work is shared under a \href{https://creativecommons.org/licenses/by-sa/4.0/}{CC BY-SA 4.0 license} unless otherwise noted
}

\usepackage{graphicx}
\usepackage{amsmath}
\usepackage{amsfonts}
\usepackage{multirow}
\usepackage[normalem]{ulem}
\usepackage[nolist]{acronym}
\usepackage[hidelinks]{hyperref}
\usepackage{rotating}
\usepackage{xspace}
\usepackage{numprint}
\usepackage[margin=1in]{geometry} 
\usepackage{utfsym}
\usepackage{pdflscape}
\usepackage{float}
\usepackage{comment}
\usepackage{booktabs}
\usepackage{longtable}
\usepackage{adjustbox}
\usepackage{arydshln}
\usepackage{listings}

\newcommand{\ie}{\textit{i}.\textit{e}.,\xspace}
\newcommand{\eg}{\textit{e}.\textit{g}.,\xspace}
\newcommand{\vs}{\textit{v}\textit{s}\xspace}

\def\totapproaches{88\xspace}
\def\excludedapproaches{8\xspace}
\def\checkedapproaches{45\xspace}
\def\unsupapproaches{49\xspace}
\def\supapproaches{28\xspace}
\def\hybridapproaches{11\xspace}
\def\numlicence{6\xspace}
\def\licenceNotSpecified{46\xspace}
\def\licenceApache{23\xspace}
\def\licenceCCA{2\xspace}
\def\licenceGPL{2\xspace}
\def\licenceMIT{10\xspace}

\begin{document}

\title{Survey on Semantic Interpretation of Tabular Data: Challenges and Directions}

\author{Marco Cremaschi$^1$ \and Blerina Spahiu$^2$ \and Matteo Palmonari$^1$ \and Ernesto Jimenez-Ruiz$^2$}

\date{
	$^1$University of Milano - Bicocca \\ \texttt{\{marco.cremaschi,blerina.spahiu,matteo.palmonari\}@unimib.it}\\%
	$^2$City, University of London \\ \texttt{ernesto.jimenez-ruiz@city.ac.uk}\\[2ex]%
}

\maketitle

\begin{abstract}
  Tabular data plays a pivotal role in various fields, making it a popular format for data manipulation and exchange, particularly on the web. The interpretation, extraction, and processing of tabular information are invaluable for knowledge-intensive applications. Notably, significant efforts have been invested in annotating tabular data with ontologies and entities from background knowledge graphs, a process known as \ac{sti}. \ac{sti} automation aids in building knowledge graphs, enriching data, and enhancing web-based question answering. This survey aims to provide a comprehensive overview of the \ac{sti} landscape. It starts by categorizing approaches using a taxonomy of 31 attributes, allowing for comparisons and evaluations. It also examines available tools, assessing them based on 12 criteria. Furthermore, the survey offers an in-depth analysis of the Gold Standards used for evaluating \ac{sti} approaches. Finally, it provides practical guidance to help end-users choose the most suitable approach for their specific tasks while also discussing unresolved issues and suggesting potential future research directions.
  
  \noindent\textbf{Keywords:} Semantic Table Interpretation, Semantic Annotation, Table, Knowledge Graph, Table-to-KG Matching, Semantic Web
\end{abstract}

\section{Introduction}
\label{sec:introduction}

\begin{figure}[ht]
  \centering
  \includegraphics[width=0.5\textwidth]{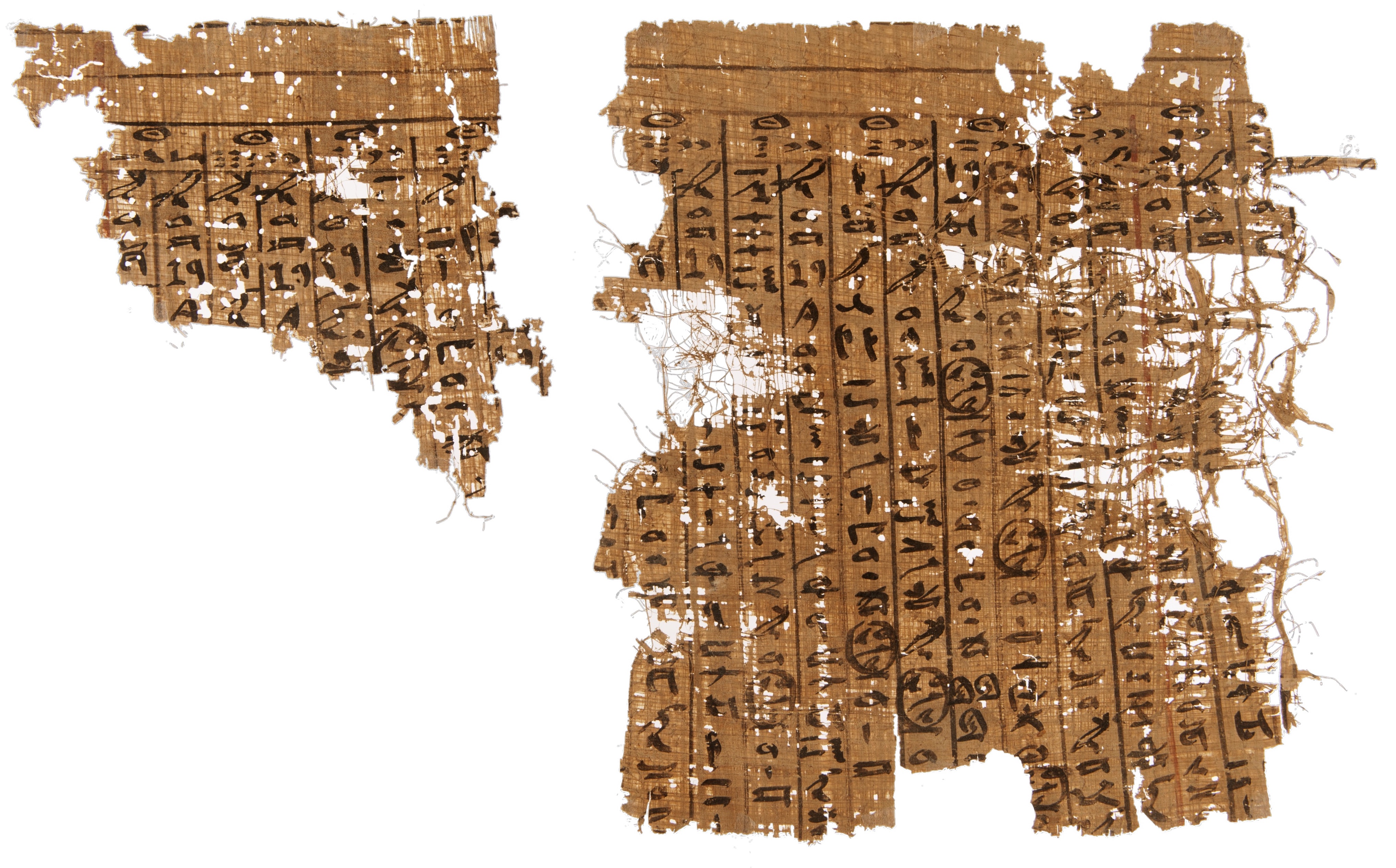}
  \caption{Portion of the diary of Merer (around 2600 BC), an official in charge of a team of workers responsible for transporting limestone blocks from Tura to Giza to construct the Great Pyramid. The document details various aspects of the logistics involved in the transportation process, such as the organisation of labour, the use of boats to navigate the Nile River, and the daily activities of the workers.}
  \label{fig:tallet}
\end{figure}

Tables are widely used and play a crucial role in creating, organising, and sharing information. A notable example of their significance as ways to organise human knowledge can be found in the oldest sample of writing on paper (on papyrus), dating back to around 2500 BC, in which Merer, an Egyptian naval inspector, documents his daily activities in a table (Fig.~\ref{fig:tallet})~\cite{Tallet2021}.

Today, tables are extensively used in both business and scientific sectors, primarily in the form of spreadsheets and other tabular data formats. They frequently appear in documents, including web pages, and are used to publish large amounts of data on the web, especially after the uptake of the Open Data movement. The large amount of tabular data consumed today covers a wide range of domains, such as finance, mobility, tourism, sports, or cultural heritage~\cite{Neumaier2016}. The relevance and diversity of tabular data can be sized by looking at the number of available tables and/or users of tabular data manipulation tools:
\begin{itemize}
  \item Web tables: in 2008 14.1 billion HTML tables were extracted, and it was found that 154 million were high-quality tables (1.1\%). In the Common Crawl 2015 repository, there are 233 million content tables\footnote{\href{https://commoncrawl.org/}{commoncrawl.org}};
  \item Wikipedia tables: the 2022 English snapshot of Wikipedia contains \numprint{2803424} tables from \numprint{21149260} articles~\cite{Marzocchi2022};
  \item Spreadsheets: there are 750 million to 2 billion people in the world who use either Google Sheets or Microsoft Excel\footnote{\href{https://askwonder.com/research/number-google-sheets-users-worldwide-eoskdoxav}{askwonder.com/research/number-google-sheets-users-worldwide-eoskdoxav}}.
\end{itemize}

The heterogeneity of tables and their application domains reflects differences in characteristics, such as the size, cleanliness, and availability of human-interpretable descriptions (headers, metadata, descriptions of natural languages).  
Despite the simplicity of tabular data, understanding their meaning and automating several downstream tasks remains challenging~\cite{fang2024large,pujara2021tables}.

\textit{Semantic Table Interpretation (STI)} 
encompasses various tabular data interpretation tasks that involve labeling an input table using reference knowledge bases and shared vocabularies (these tasks are sometimes also referred to as table annotation or semantic modeling ~\cite{Syed2010}). 
Since \ac{kgs} have become one of the most popular abstractions for knowledge bases and are equipped with shared vocabularies, STI can be understood as a table to \ac{kgs} matching problem. ~\ac{kgs} are used to represent relationships between different entities such as people, 
places,
mountains, 
events,
and so on~\cite{Hogan2021}. They organise knowledge in graph structures where the meaning of the data is encoded alongside the data in the graph.~\ac{rdf}\footnote{\href{https://www.w3.org/RDF/}{www.w3.org/RDF/}} is a data model for representing \ac{kgs} that come with an ecosystem of languages and protocols to foster interoperable data management. 
In \ac{rdf}, most of the graph nodes represent \textit{instances} and \textit{classes} - referred to here as \textit{entities} and \textit{types}, respectively- and are identified by URIs or \textit{literals} (\eg~strings, numbers); most of the edges, each labeled by an RDF \textit{property}, represent relations between nodes, i.e, two entities or an entity and a literal. Some of these edges link entities to their types (\eg~\texttt{dbo:City})  or datatypes (\eg~\texttt{xmls:integer}). Finally, some edges are used to model the \textit{ontologies} that organise the knowledge (e.g., subclass relations between types) and specify the meaning of the terms used in the graph through logical axioms. Labeling the elements of a table with elements of a knowledge graph supports their interpretation, e.g., by disambiguating the meaning of the headers, or of the values that correspond to entities, and it is possible to transform the table into actionable knowledge in different downstream application (see Section~\ref{sec:task_definition_challeneges}). 

\ac{sti} has developed as an active research area attracting the scientific community's attention. An extensive number of approaches have been proposed to tackle STI tasks, from those that use heuristic matching methods to those that use or include feature-based machine learning methods~\cite{Liu2023,Keshvari-Fini2019} to the latest ones that use or are entirely based on Pre-trained Language Models (PLM) such as BERT~\cite{yin2020tabert,huynh2023towards} or generative Large Language Models (LLMs) like Llama~\cite{fang2024large,Zhang2023}. Additionally, studies have examined how users approach reading tables~\cite{cremaschi2022really}. Contributions to \ac{sti} include methods inspired by a variety of \ac{ai} paradigms and have been published in \ac{ai} journals and conferences or in journals and conferences related to the sibling fields of Semantic Web, \ac{nlp}, and Data Management (for more details see Fig.~\ref{fig:conferences-journal})~\cite{Ruiz2019,Ruiz2020,Cutrona2021-1}; this broad scope suggest that the topic is considered relevant across different research sub-communities. Another initiative reflecting the interest in the topic is the international Semantic Web Challenge on Tabular Data to Knowledge Graph Matching (SemTab), which has been proposed since 2019\footnote{\href{https://www.cs.ox.ac.uk/isg/challenges/sem-tab/}{cs.ox.ac.uk/isg/challenges/sem-tab}} and still running in 2024~\cite{Ruiz2019,Ruiz2020,Cutrona2021-1}. The initiative represents a community-driven effort to formalize the different STI tasks and develop shared evaluation protocols to compare different approaches.  

In this paper, we propose a comprehensive survey on approaches proposed to address \ac{sti} tasks, also covering the latest approaches based on PLMs and LLMs. In Section~\ref{sec:task_definition_challeneges}, we present \ac{sti} tasks more precisely and summarise the impact of \ac{sti} on research
and applications and their challenges; In Section~\ref{sec:contribution}, we summarise the contributions of our paper and present its structure.

\subsection{STI: key definitions, impact and challenges}
\label{sec:task_definition_challeneges}

\begin{figure}
  \centering
  \begin{minipage}[h]{.5\textwidth}
    \centering
    \includegraphics[width=.9\textwidth]{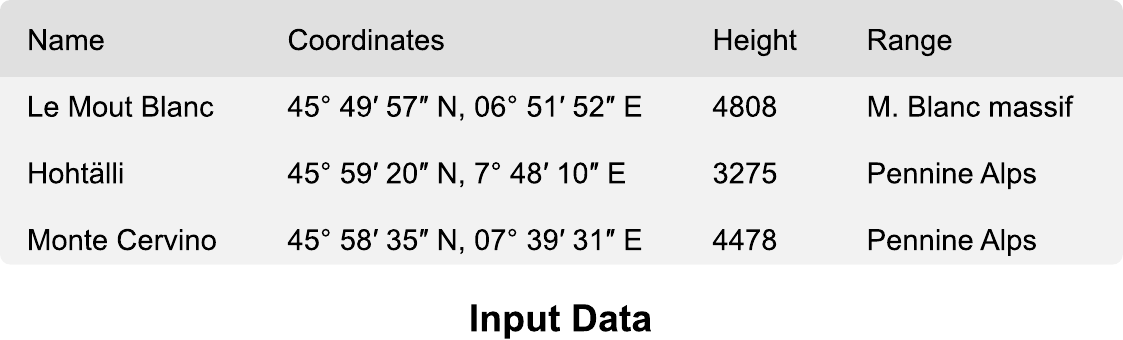}
    \caption{Example of a well-formed relational table.}
    \label{fig:01_def_table}
  \end{minipage}%
  \begin{minipage}[h]{.5\textwidth}
    \centering
    \includegraphics[width=.9\textwidth]{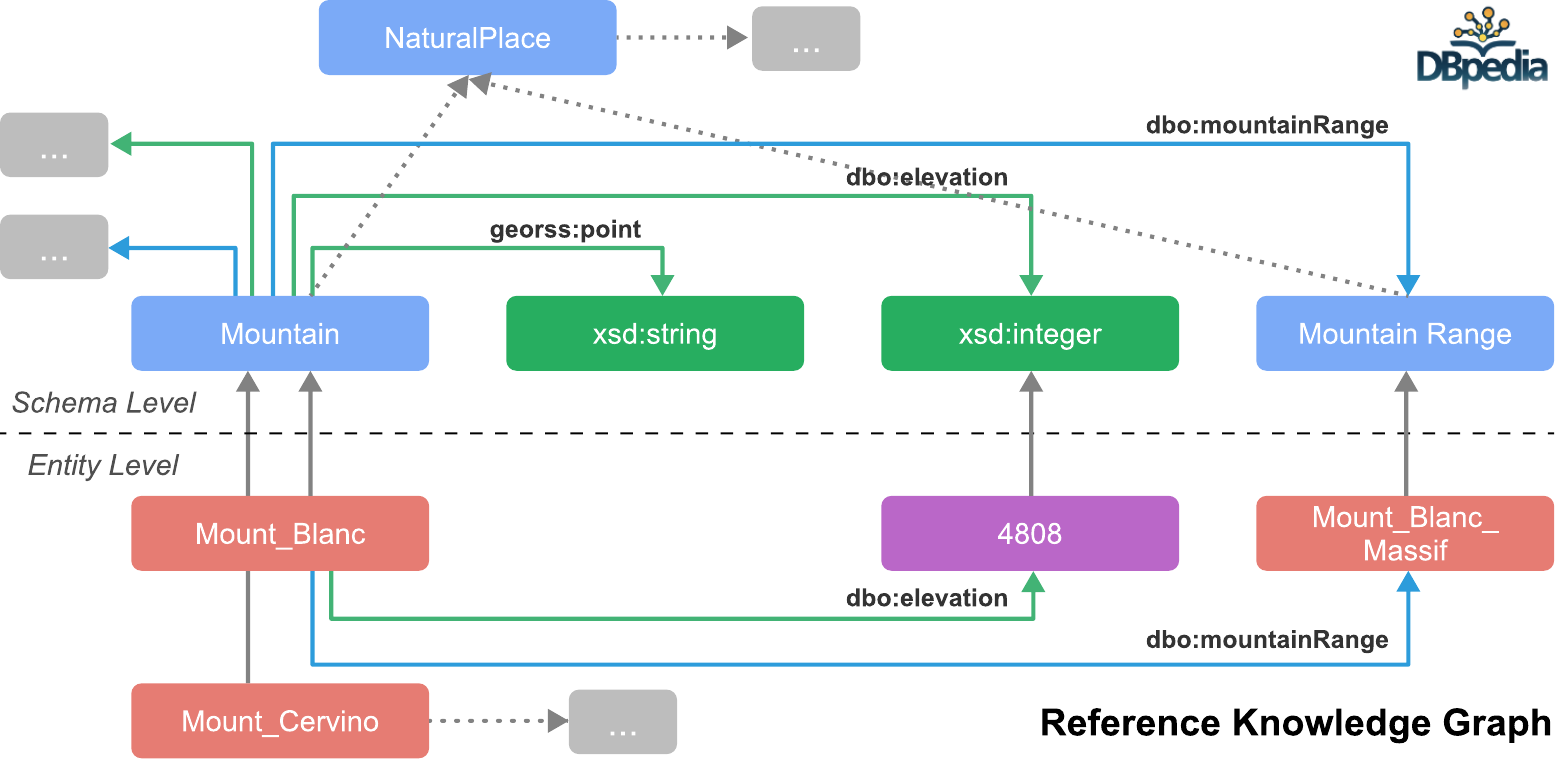}
    \caption{A sample of \acl{kg}.}
    \label{fig:02_def_kg}
  \end{minipage}
\end{figure}

In its most agreed and complete formalisation, the STI process considers two inputs: i) a relational table, which is usually assumed to be \textit{well-formed and normalised} (\ie the table has a grid structure, where the first row may contain the table headers and any other row contains values), as in Fig.~\ref{fig:01_def_table}; and ii) a reference \textit{\ac{kg}} with its vocabulary (\ie a set of symbols denoting concepts, datatypes, properties, instances - also referred to as \textit{entities} in the following) as in Fig.~\ref{fig:02_def_kg}). The output of the \ac{sti} process is a semantically annotated table, \ie a table where its elements, typically values, columns, and column pairs, are annotated with symbols from the \ac{kg} vocabulary.
The exact specification of the annotations expected as the output of the \ac{sti} process may differ in the proposed approaches. Here we discuss a canonical definition of a \textit{semantically annotated table} to provide a first understanding of key \ac{sti} tasks, inspired by the SemTab Challenge, where the annotation process has been better formalised with a community-driven effort. 
 
\begin{figure}[ht]
  \centering
  \includegraphics[width=\textwidth]{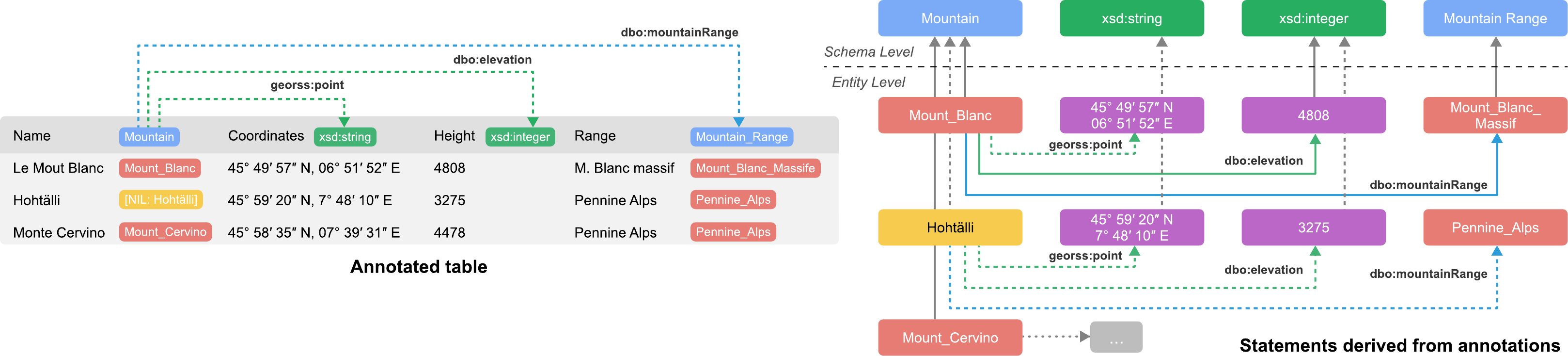}
  \caption{Example of an annotated table.}
  \label{fig:03_def_sti}
\end{figure}

To discuss this canonical definition, we use the example reported in Fig.~\ref{fig:03_def_sti}.


Given:
\begin{itemize}
  \item a relational table $T$ (Fig.~\ref{fig:01_def_table});
  \item a Knowledge Graph and its vocabulary (Fig.~\ref{fig:02_def_kg}).
\end{itemize}
$T$ is annotated when:
\begin{itemize}
  \item each column is associated with one or more types from the \ac{kg} [\ac{cta}]; \eg~the column \textit{Name} in the Fig.~\ref{fig:01_def_table} is annotated with the type \texttt{Mountain}; the column \textit{Height} is annotated with datatype \texttt{xsd:integer};
  \item each cell in ``entity columns'' is annotated with an entity identifier or with \textit{NIL}, if no entity in the KG corresponds to the cell value [\ac{cea}]; \eg~the cell \textit{Le Mout Blanc} is annotated with \texttt{Mont\_Blanc}; the cell \textit{Hohtälli} is annotated with \textit{NIL} since it has not yet been included in the \ac{kg};
  \item some pairs of columns are annotated with a binary property [\ac{cpa}]; \eg~the pairs composed by the columns \textit{Name} and \textit{Hight} is annotated with \texttt{dbo:elevation}.
\end{itemize}
The result of the annotation process for the table considered in the example is shown in Fig.~\ref{fig:03_def_sti}. Observe that each bullet point can be interpreted as a high-level STI task to complete; also, the annotations can identify in the table new entities not included in the reference \ac{kg} (\eg \texttt{Hohtälli}). 

\begin{figure}[ht]
  \centering
  \includegraphics[width=\textwidth]{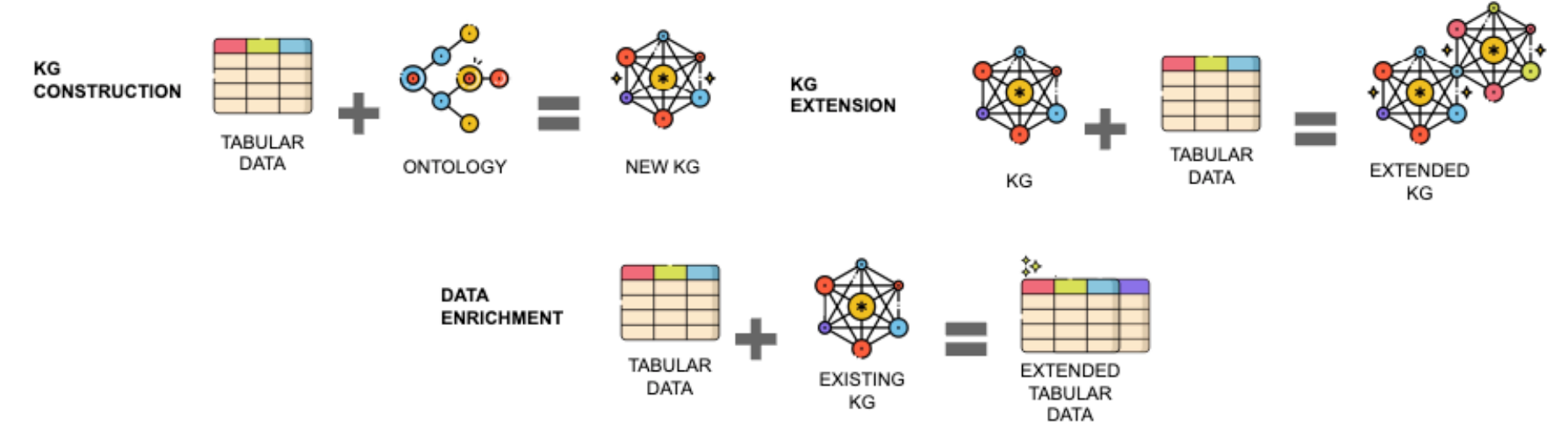}
  \caption{Examples of applications supported by STI.}
  \label{fig:motivations}
\end{figure}

STI plays a relevant role in the \ac{ai} research and applications landscape. From a \textbf{research perspective}, the capability of performing STI tasks such as \ac{cea}, \ac{cta}, \ac{cpa} are considered part of a broader set of tabular data understanding skills~\cite{yin2020tabert,du2021tabularnet,Zhang2023}, which impact the application of \ac{ai} to tabular data. Also, \ac{cea} can be considered a variant of entity linking in texts, while \ac{cta} and \ac{cpa} are not too different from ontology matching when applied to different data formats (entity linking and ontology matching are both considered \ac{ai} tasks~\cite{li2022community,kolitsas-etal-2018-end,huynh2023towards}). From an \textbf{application perspective}, \ac{sti} tasks can support the automation of processes to construct and extend knowledge bases~\cite{Weikum2021,Kejriwal2021} and enrich tabular data, eventually supporting downstream applications to data analysis. To provide an idea of the contributions of \ac{sti} to these automation processes, we refer to Fig.~\ref{fig:motivations}. For \textit{\ac{kg} construction}, \ac{cta} and \ac{cpa} annotations support the automatic or semi-automatic transformation of the data into a graph format with the schema of the reference \ac{kg}~\cite{Gupta2015,Ramnandan2015,Pham2016,Taheriyan2016-1};
\ac{cea} annotations disambiguate values in the input table, thus supporting the reuse of canonical entity identifiers in the generated data~\cite{Cutrona2019,Ciavotta2022}. The same principles can be applied to support \textit{\ac{kg} extension} processes~\cite{yakout2012infogather,Zhang2020}, by adding to the graph only the new entities and triples represented in the table. \ac{sti} annotations can be useful in \textit{data enrichment} tasks where the generation of graph data is not needed, but links to entities in the \ac{kg} can be used to query the \ac{kg}~\cite{harari2022few,bucher2021scikit} or other third-party data sources~\cite{Palmonari2018,cutrona2019semantically,ciavotta2022supporting}, augmenting the input data and extending the features used to develop analytical models. Other applications of \ac{sti} annotations proposed in the literature or potentially impactful on emergent services include the improvement of search engines and recommender systems for tabular data~\cite{bhagavatula2013methods,cafarella2008webtables, Zhang2018,Zhang2020-survey,benjelloun2020google} and question answering~\cite{yin2020tabert,Deng2022}.


Machine interpretation of tabular data is challenging because of the limited context available to resolve semantic ambiguities, the layout of tables that can be difficult to handle, and the incompleteness of \ac{kgs} in general. 

Key challenges involved in the annotation process include:
\begin{itemize}
    \item \textit{Dealing with the heterogeneity of domains and data distributions}: the tables may cover information that refers to very different domains (\eg~Geography \vs Sports); the specificity of the table content may vary significantly (from a table with basic information about most famous mountains, like Fig.~\ref{fig:01_def_table} to table that contains the composition of the rocks of this mountains\footnote {\href{https://en.wikipedia.org/wiki/List_of_rock_formations}{en.wikipedia.org/wiki/List\_of\_rock\_formations}}).
    \item \textit{Dealing with limited contextual information}: if compared with similar interpretation and disambiguation tasks on the textual document, the presence of contextual clues to support the interpretation and annotation of table elements may be limited and very diverse depending on the data sources; for example, table headers 
    are often missing. Tables in open data portals may be described by metadata, while tables published on web pages may have some surrounding text. 
    \item \textit{Detecting the type of columns}: in a table, there can be columns that contain references to named entities (NE-columns)
    and columns that contain strings, numbers, dates, and, in general, instances of specific data types, which we refer to as literals (LIT-columns); distinguishing between the two types of columns is crucial to support the annotation process.
    \item \textit{Matching tabular values against the KG}: matching the values in the table to the data in the KG helps collect evidence to interpret the table. However, the values referring to entities in the table may differ from their labels in the \ac{kg}, \eg because of acronyms, aliases, and typos, while  
    other values representing their features may differ for several reasons, \eg~because outdated, measured differently, and so on.
    \item \textit{Dealing with multiple entities with similar names}: the \ac{kg} may contain many entities with similar or even equal names (homonyms) that have different or even the same types. For example, the mention of the Italo-French mountain Mont\_Blanc in Fig.~\ref{fig:01_def_table}
    matches labels of more than a dozen entities, including a tunnel, a poem, a dessert, and another mountain on the moon\footnote{\href{https://en.wikipedia.org/wiki/Mont\_Blanc\_(disambiguation)}{en.wikipedia.org/wiki/Mont\_Blanc\_(disambiguation)}}. 
    \item \textit{Dealing with \ac{nil}-mentions}: some entities referred to in the table may not exist in the \ac{kg}; while several approaches perform CEA by simply selecting the best candidate, \ie the entity with the highest score according to the approach, recognizing new entities requires a decision whether to link or not to link, which may be subtle.
    \item \textit{Choosing the most appropriate types and properties}: the \ac{kg} may contain hundreds or thousands of types and properties to choose from for annotating columns and column pairs. The features of large \ac{kgs} make the decision even more challenging: entities are classified under multiple types, which may reflect different levels of specificity 
    (\eg 
    Mont Blanc can be classified as a mountain, a summit, a pyramidal peak, and so on, up to a geographic location);
    the specificity of the classification may change depending on the entity; 
    several properties have similar meanings, associated with different levels of specificity or different usage patterns~\cite{Porrini2018}. 
    \item \textit{Aggregate evidence from different tasks}: the annotation of a table is, in principle, a collective decision-making process; for example, the disambiguation of entities in a column helps suggest types to annotate the column (\eg~most of the best candidates for mountain mentions in Fig.~\ref{fig:01_def_table} are mountains), but a type or a set of types associated with a column may help disambiguate entities mentioned therein (\eg~non-mountain candidate entities for ``Mont Blanc''); finding strategies to maximise evidence exchange across tasks requires multiple iterations or sophisticated aggregation mechanisms.
    \item \textit{Dealing with amount and shape of data}: depending on the application scenarios, it may be necessary to process a large number of small tables or very large tables~\cite{Ciavotta2022}, which may imply different constraints on the approaches or introduce slightly different challenges; more scenarios may also become more relevant in the future, such as processing streaming data that can be formatted as tabular data.
\end{itemize}

\subsection{Specific contributions and structure of the manuscript}
\label{sec:contribution}

In the presentation of this comprehensive survey on \ac{sti} we make the following more specific contributions, which highlight\sout{s} the main differences with previous surveys published on the same topic (see also Section~\ref{sec:differences-form-other-sti-surveys} for a more detailed comparison):

\begin{itemize}
  \item A new taxonomy to organise and compare reviewed approaches comprising 31 specific attributes;
  \item A new systematic literature review on \totapproaches \ac{sti} approaches published until October 2024, including latest approaches based on LLMs;
  \item Analysis of existing tools that support \ac{sti} and a comparison between their functionality features;
  \item Analysis of the \ac{gs} used to evaluate \ac{sti} approaches;
  \item A guide that can help researchers and practitioners locate \ac{sti} approaches most suited to their tasks;
  \item Highlight and discuss open issues and future research directions.
\end{itemize}

The paper is organised as follows: Section~\ref{sec:scope-methodology} highlights the differences between this survey and other similar surveys in the \ac{sti} field and discusses the methodology used to collect all approaches reviewed in this paper. Section~\ref{sec:taxonomy} defines a taxonomy composed of 31 attributes used to compare \ac{sti} approaches. Sections from~\ref{sec:sub-tasks} to~\ref{sec:validation} help readers gain a comprehensive understanding of the techniques and solutions proposed so far. Section~\ref{sec:remarks} draws open issues and future directions while Section~\ref{sec:conclusion-future-research} concludes the paper. The appendix section contains valuable and additional information regarding \ac{sti} approaches. Appendix~\ref{appendix:A} details information about the methodology to conduct the survey and comparisons of approaches. Appendix~\ref{appendix:b} discusses tools for the \ac{sti} process while Appendix ~\ref{sec:gold-standards} analyses the \ac{gss} used by \ac{sti} approaches to evaluate their performance. 
Additional information about approaches is provided in Appendix~\ref{sec:additional-material}.

\section{Scope and Methodology}
\label{sec:scope-methodology}

This Section highlights the differences between this survey and other similar surveys in the \ac{sti} field.
The second part describes the methodology we applied to our systematic literature review, based on the well-established PRISMA (Preferred Reporting Items for Systematic Reviews and Meta-Analyses) method\footnote{\href{http://www.prisma-statement.org/}{prisma-statement.org}}. 
Details on the systematic review results serve as a basis for the comprehensive analysis in the following Sections.

\subsection{Differences from other STI surveys}\label{sec:differences-form-other-sti-surveys}

\ac{sti} approaches have been analysed in a few surveys~\cite{Keshvari-Fini2019,Zhang2020-survey,Bonfitto2021,Liu2023};~\cite{Keshvari-Fini2019,Zhang2020-survey,Bonfitto2021} have been published before the explosion in volume of \ac{sti} related-works, also as a consequence of the SemTab challenge. Most recently, Liu et al.~\cite{Liu2023} aims to complement these surveys by providing a new classification of \ac{sti} approaches reflecting the heterogeneity of tabular data and the resulting new challenges.
We aim to update and extend previous surveys by introducing a new classification schema of \ac{sti} approaches and discussing new research directions in improving such systems. Nevertheless, our analysis encompasses not only recent works but also older ones, allowing us to derive comprehensive guidelines (Section~\ref{sec:remarks}) for selecting approaches based on specific user needs. Furthermore, we can identify and highlight the unresolved issues which are yet to be addressed.

To provide clarity, we highlight the following differences with the previously published surveys:
\begin{itemize}
  \item Survey scope: through a rigorous snowballing approach, we collected a comprehensive list of \ac{sti} approaches that allowed us to discover \totapproaches works. Moreover, our survey includes works of a wider timeline (2007-2023);
  \item Taxonomy: considering the comprehensive list of all the works in this field allowed us to specify and classify \ac{sti} systems using different orthogonal dimensions. In this survey, we identify 31 dimensions;
  \item Processes: providing a better understanding of the entire processes of \ac{sti} by shedding light on each step;
  \item Deeper investigation: examining a wider range of approaches enabled us to delve deeper into the field, thus, helping researchers and practitioners to better understand and inspire improved or novel approaches. Similarly to~\cite{Liu2023}, we delve into a more comprehensive comparison of the evaluation process;
  \item Opportunity discovery: uncovering research opportunities of the existing approaches. For instance, unlike~\cite{Zhang2020-survey} and~\cite{Liu2023}, we provide a more detailed analysis of the potentials of the available \ac{sti} approaches;
  \item Additional sections: including other important elements, \eg~delving deep into tools and \ac{gs}, this survey provides the complete landscape of the \ac{sti} process.
\end{itemize}

A comparison of the surveys is presented in Table~\ref{tab:survey_comparison}, which reflects the above attributes and highlights the differences between them.

Two additional surveys~\cite{barlaug2021neural,getoor2012entity}, which are partially relevant to \ac{sti}, were also considered. Unlike~\cite{getoor2012entity}, our work provides a more comprehensive analysis of various approaches to the \ac{el} task, accompanied by an in-depth discussion of the associated challenges. However, in contrast to~\cite{barlaug2021neural}, which delves into the technical specifics of entity resolution using \ac{nn}, our focus is less specific on technical details because it covers all \ac{sti} methodologies.

\begingroup
\dashlinedash=1pt
\dashlinegap=1pt
\setlength{\tabcolsep}{5pt}
\renewcommand{\arraystretch}{1.5}

\begin{table}[ht]
  \begin{adjustbox}{width=\textwidth}
    \begin{tabular}{m{100pt}|m{100pt}:m{100pt}:m{100pt}:m{100pt}:m{100pt}}
      \rowcolor[HTML]{DDDDDD}
      \textbf{Attribute}                                  & \textbf{This survey}                                                                                               & 2019~\cite{Keshvari-Fini2019}                                                                                                  & 2020~\cite{Zhang2020-survey}               & 2021~\cite{Bonfitto2021}                                        & 2023~\cite{Liu2023}                                                                               \\

      \textbf{No.\ of Approaches}                         & 85                                                                                                                 & 16                                                                                                                             & 47                                         & 12                                                              & 42                                                                                                \\
      \rowcolor[HTML]{F3F3F3}
      \textbf{Years range}                                & 2007 --- 2023                                                                                                      & 2009 --- 2017                                                                                                                  & 2002 --- 2019                              & 2010 --- 2019                                                   & 2011 --- 2021                                                                                     \\

      \textbf{Gold Standards \newline analysis}           & \color[HTML]{93C47D} ✓ \color[HTML]{000000} 21                                                                     & \color[HTML]{FF0000} \textbf{✗}                                                                                                & \color[HTML]{FF0000} \textbf{✗}            & {\color[HTML]{93C47D} ✓ \color[HTML]{000000}7 (brief analysis)} & {\color[HTML]{93C47D} ✓ \color[HTML]{000000}8}                                                    \\
      \rowcolor[HTML]{F3F3F3}
      \textbf{Formalisation of \newline the STI pipeline} & {\color[HTML]{93C47D} ✓}                                                                                           & {\color[HTML]{FF0000} \textbf{✗}}                                                                                              & {\color[HTML]{FF0000} \textbf{✗}}          & {\color[HTML]{FF0000} \textbf{✗}}                               & {\color[HTML]{93C47D} ✓}                                                                          \\

      \textbf{Comparison attributes}                      & 31 \newline Table classification\newline NLP Tasks on Table\newline Other Projects\newline Discover and Understand & 6\newline Table expansion\newline Table interpretation\newline Table search\newline Question answering\newline KG augmentation & 6\newline Table expansion\newline
      Table interpretation\newline
      Table search\newline
      Question answering\newline
      KG augmentation                                     & 0                                                                                                                  & 5\newline Lookup based\newline Iterative\newline Feature based\newline KG modelling\newline Table modelling                                                                                                                                                                                                                                       \\
      \rowcolor[HTML]{F3F3F3}
      \textbf{Additional sections}                        & Tools                                                                                                              & ---                                                                                                                            & Table classification\newline Table corpora & ---                                                             & Table classification\newline (extension of~\cite{Zhang2020-survey})\newline Evaluation comparison \\

      \textbf{Note}                                       & ---                                                                                                                & Only Web tables                                                                                                                & Focus on other\newline downstream tasks    & Focus on table\newline understanding                            & ---
    \end{tabular}
  \end{adjustbox}
  \caption{Comparison between surveys on STI.}
  \label{tab:survey_comparison}
\end{table}
\endgroup

\subsection{Methodology}

The objective of this systematic review is to provide a synthesis of the state of knowledge and suggestions for future research. The PRISMA method has been designed to provide detailed reporting guidelines for such reviews to ensure a comparable and comprehensive result. This method typically encompasses three stages: i)~identification, ii)~screening, and iii)~selection. In this section, we provide a short overview of the methodology employed to conduct this survey. Please refer to Appendix \ref{appendix:methodology-long} for more details. 

In the identification stage, to efficiently search in different databases for related works, we defined a set of 16 keywords related to semantic table interpretation. These keywords were ranked based on relevance by five researchers. We conducted searches on platforms including Scopus, Web of Science, DBLP, and Google Scholar, covering the period from 2007 to May 2023. We also employed a snowballing technique to include recent publications referencing key works.

Instead, in the screening stage, two experts manually reviewed the identified papers, focusing on the semantic table interpretation phases of the approaches and their relevance. A categorization process was performed based on title, abstract, and keywords. Specific criteria were used, including generic and specific annotation tags, to determine relevance. 

In the selection stage, publications included in this survey were required to be directly related to semantic table interpretation, published in English, and peer-reviewed. Using the specified keywords defined within the PRISMA method, 134 papers were initially identified, which were reduced to 111 after the screening process. Manual annotation and further screening led to the exclusion of 17 papers, resulting in a total of \totapproaches approaches discussed in the survey. 

\section{Taxonomic analysis of STI Approaches}
\label{sec:taxonomy}

\begin{figure}[H]
  \centering
  \includegraphics[width=0.7\textwidth]{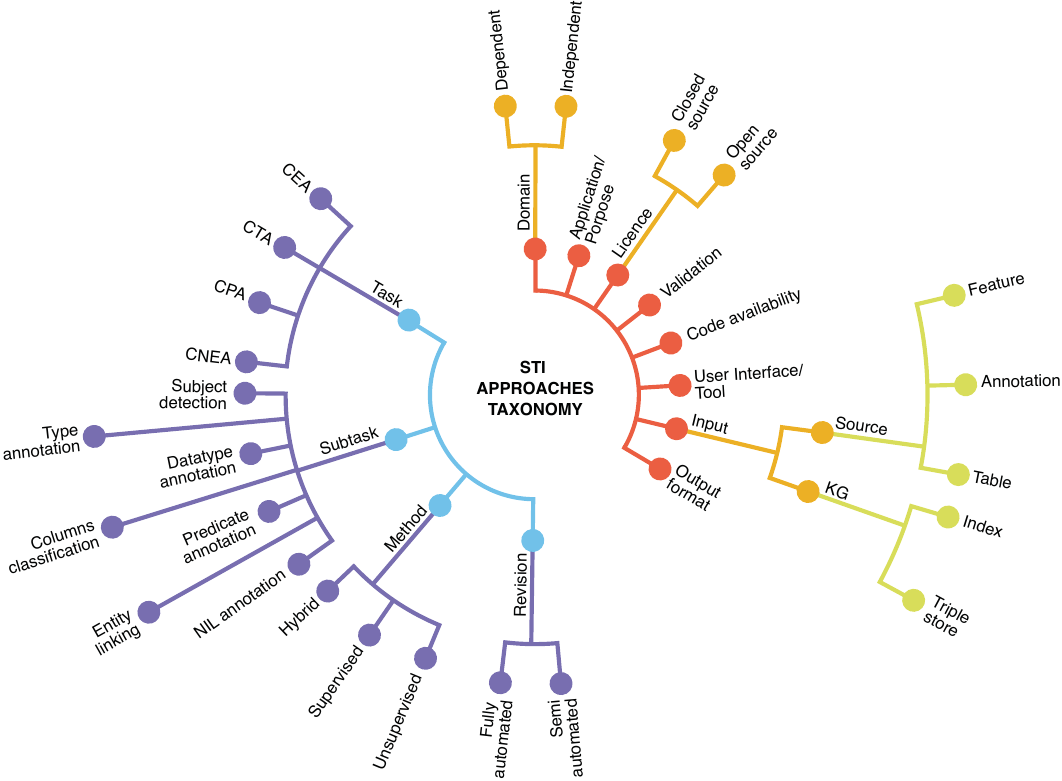}
  \caption{A taxonomy of the \ac{sti} approaches.}
  \label{fig:taxonomy}
\end{figure}

Introducing a taxonomy of features that characterise different \ac{sti} approaches\footnote{\href{https://unimib-datai.github.io/sti-website/approaches/}{unimib-datai.github.io/sti-website/approaches/}} as to main objectives: i)~defined \ac{sti} more precisely by describing the tasks and subtasks, ii)~allows us to comprehensively understand the various approaches and their unique contributions to the field. Fig.~\ref{fig:taxonomy} depicts a high-level taxonomy of the features of \ac{sti} approaches. The features are organised into dimensions. By analysing each dimension individually, we will present the main characteristics of the approaches proposed so far.
The reader should note that many dimensions are orthogonal; thus, an approach may be classified into multiple other ones.
To ensure that the information presented in the survey was verified and complete, we contacted the authors via email and received \checkedapproaches responses.

\paragraph{\textbf{TASKS}} \textit{TASKS} provide the conceptualisation of the output that \ac{sti} approaches are expected to return. TASKS have been defined precisely because the quality of the output of the approaches is usually evaluated (only) against them. In this paper, we consider the TASKS that have been formalised in previous works~\cite{Jimenez2020, Ruiz2020} or have appeared in the latest challenges:

\begin{itemize}
    \item \ac{cta}: the \textit{\ac{cta}} task concerns the prediction of the semantic types (\ie \ac{kg} classes) for every given table column in a table.
    \item \ac{cpa}: the \textit{\ac{cpa}} task concerns the prediction of semantic properties (\ie \ac{kg} properties) that represent the relationship between some pair of columns.
  \item \ac{cea}: the \textit{\ac{cea}} task aims to predict the entity (\ie instances) that a cell in table represents.
  \item \ac{cnea}: the \textit{\ac{cnea}} task aims to predict which cell in the table represents an entity that does not occur in the \ac{kg} and should be therefore labelled as \ac{nil}.
\end{itemize}


Note that \ac{cta} (resp.~\ac{cea}) task focuses on annotating table columns (resp.~cells) that can be represented with a \ac{kg} class/type (resp.~\ac{kg} entity).
The formal definition presented in this Section is more precise but less flexible than the intuitive definition provided in the introduction. The reason for this choice is that most existing approaches in the literature focus on specific tasks that require a more rigorous definition. However, in the last section of this work, we will explore an expanded version of this formalisation that accounts for different application scenarios.

\paragraph{\textbf{SUB-TASK}} The \ac{sti} process involves coordinating multiple specific annotation sub-tasks that contribute to the final results. These sub-tasks are more focused than the overall conceptual tasks mentioned earlier. Approaches may vary in how they coordinate and the algorithms they employ for each sub-task. This aspect evaluates an approach's coverage of the sub-tasks within the \ac{sti} process. We believe that the granularity of the sub-task classification is the best one to report different techniques proposed in the literature so far (Section~\ref{sec:sub-tasks}). Some approaches implement just a few sub-tasks, while others consider the implementation of all sub-tasks listed below.


(i) \textit{Column Classification} considers the content of the cells of each column to mark a column as \textit{Literal column (LIT-column)} if values in cells are literals (\eg~strings, numbers, dates such as 4808, 10/04/1983), or as \textit{Named-Entity column (NE-column)} if values are entities, instances of types (\eg~Mountain, Mountain Range such as Le\_Mout\_Blanc, M.\_Blanc\_massif). The \textit{Column Classification} sub-task is useful, especially, for the \ac{cea} and \ac{cta} tasks because the identification of \textit{NE-columns} helps to concentrate the \textit{Entity linking} task on specific cells and the \textit{Type Annotation} tasks on specific columns;

(ii) \textit{Subject Detection} has the goal of identifying, among the NE-columns, the column that all the others are referring to, also called S-column (\eg~the Name column in Fig.~\ref{fig:03_def_sti}). The \textit{Subject Detection}, in some cases, is useful for the \ac{cpa} task because it allows to find relations between other columns;

(iii) \textit{Type Annotation} pairs NE-columns with concepts of the \ac{kg} (\eg~the column Name is annotated with Mountain in DBpedia). This sub-task represents the final output of the \ac{cta} task;

(iv) \textit{Entity linking} links cell to entity in the \ac{kg} (\eg~the cell Le Mout Blanc is annotated with \texttt{dbo:Mont\_Blanc} in DBpedia). This sub-task represents the final output of the \ac{cea} task;

(v) \textit{Datatype Annotation} pairs LIT-columns with a datatype in the \ac{kg} (\eg~the column Coordinates is of type \texttt{georss:point}). The \textit{Datatype Annotation} sub-task is used in \ac{cpa} because fine-grained types of Lit-columns are easier to match against \ac{kgs};

(vi) \textit{Predicate Annotation} identifies the relations between each pair of columns (\eg~Name \texttt{dbo:elevation} Height). This sub-task represents the final output of the \ac{cpa} task;

(vii) \textit{\ac{nil} Annotation} considers strings that refer to entities for which a representation has not yet been created within the \ac{kg}, namely \textit{\ac{nil}-mentions} (\eg~the mention Hohtälli). This sub-task represents the final output of the \ac{cnea} task.

These sub-tasks are sometimes preceded by a sub-task of \textit{Data Preparation} that is used to normalise the contents (\eg~by standardising the case of letters and the format of numbers) to avoid the presence of syntactical discrepancies that can make annotation techniques ineffective.

Each sub-task is computed by annotating cell values referring to one or more \ac{kg}s. The general approach consists of searching the \ac{kg} with the content of columns to find possible matching. For example, if the majority of entities in the Name column (Fig.~\ref{fig:03_def_sti}) is associated with \texttt{dbo:Mountain}, then all entities in the column can be assumed of type \texttt{dbo:Mountain}. Similarly, if the majority of the concepts of type \texttt{dbo:Mountain} are connected to datatypes of type \texttt{xsd:integer} by the property \texttt{dbo:elevation}, then it can be identified as the property connecting the Name column with the Height column.
However, it is important to note that this approach only applies to trivial cases. Employing advanced methods and techniques to address ambiguous results frequently occurring in real-world tables becomes necessary to tackle more complex scenarios. These methods aim to identify suitable matches for elements that cannot be directly linked to entities within a \ac{kg}.

\paragraph{\textbf{METHOD}} Another crucial dimension in reviewing works on \ac{sti} is the classification based on the main algorithmic idea behind each approach. It is possible to identify three \textit{METHODS}:
\begin{itemize}
  \item SUPERVISED category collects approaches that rely on a training set (\eg~a set of tables already annotated) that learn the annotations before applying them to the target tables;
  \item UNSUPERVISED category collects approaches that do not use annotated data;
  \item HYBRID category instead, collects approaches combining the above two categories.
\end{itemize}

\paragraph{\textbf{REVISION}} Some approaches are fully automated, while others require user intervention to select or validate annotations.

\paragraph{\textbf{DOMAIN}} Approaches can target tables with general or specific data (\eg~bio data, geospatial data).

\paragraph{\textbf{APPLICATION/PURPOSE}} This dimension refers to the use of the approach for particular application purposes such as data enrichment, \ac{kg} construction and \ac{kg} extension. Annotated data might be used as links to find new information for the entities in the table, thus enriching the input, or otherwise to extend and enrich \ac{kgs} as described in Section~\ref{sec:introduction}. Moreover, supposing tables are specific to a given domain, the data might be annotated using specific ontologies or vocabularies to consider the final annotations as newly constructed \ac{kg}.

\paragraph{\textbf{LICENSE}} When it comes to ensuring reproducibility in research, it is crucial to consider the licensing aspect. In this regard, we distinguish between different licensing models that govern the availability and use of approaches:
\begin{itemize}
  \item OPEN SOURCE category collects approaches published under an open source license, facilitating comparison and reuse (\eg~Apache, MIT);
  \item CLOSED SOURCE assemble \ac{sti} approaches, for which the code is not provided, and as such, it is not straightforward to implement or compare it;
  \item NOT SPECIFIED approaches that do not specify any license information.
\end{itemize}

\paragraph{\textbf{VALIDATION}} The approaches might be validated using different \ac{gss}. Some use and validate their annotations using well-known \ac{gss}, and others provide a new \ac{gs} together with the release of the new approach.

\paragraph{\textbf{CODE AVAILABILITY}} From a practical perspective, it is interesting to know the availability of the code of a given \ac{sti} approach.

\paragraph{\textbf{USER INTERFACE/TOOL}} Several \ac{sti} approaches also implement a \ac{ui} so that the \ac{sti} process can be consumed and explored by users.

\paragraph{\textbf{INPUTS}} One of the dimensions for the classification of the approaches is the required \textit{INPUTS}. Among \textit{INPUTS}, we can distinguish between \textit{SOURCES} that are to be annotated and additional resources (\textit{KG}) that support the annotation process.
\begin{itemize}
  \item SOURCES: different systems might need different input sources. Three sources are identified: (i) TABLES are most frequently considered as input (\eg~CSV files, XML files, Spreadsheets, and HTML files), (ii) ANNOTATIONS from already annotated data used as training datasets (supervised approaches), and (iii) FEATURES that support the \ac{sti} approach with additional information such as out-table context (\eg~page title, table caption, texts).
  \item \ac{kg}: the annotation process is facilitated using \ac{kgs}. A \ac{kg} can be stored by a (i) TRIPLE STORE that supports entity searches by lookup APIs (\eg~DBpedia SPARQL Query Editor),
  and by (ii) INDEXING that allows efficient querying of large \ac{kgs} that would otherwise require significant time and resources. 
\end{itemize}

\paragraph{\textbf{OUTPUTS}} The annotated data might be exported into different formats, such as RDF/XML, N3, and CSV.

\section{Sub-tasks}
\label{sec:sub-tasks}

The taxonomy dimension SUB-TASKS refers to the completeness of the approach concerning the sub-tasks of the \ac{sti} process. Some approaches implement just a few sub-tasks, while others consider the implementation of all sub-tasks: i)~Data Preparation (Section~\ref{subsec:datapreparation}), ii)~Column Classification (Section~\ref{subsec:column-classification}), iii)~Datatype Annotation (Section~\ref{subsec:datatype-annotation}), iv)~Subject Detection (Section~\ref{subsec:subject-detection}), v)~Entity Linking (Section~\ref{subsec:entity-linking}), vi)~Type Annotation (Section~\ref{subsec:type-annotation}), vii)~Predicate Annotation (Section~\ref{subsec:property-annotation}), and viii)~\ac{nil} Annotation (Section~\ref{subsec:nil-annotation}).

\subsection{Data Preparation}
\label{subsec:datapreparation}

Data Preparation is usually the first sub-task in an \ac{sti} pipeline. This sub-task transforms the raw data into a format suitable for analysis. Data Preparation plays a crucial role in \ac{sti} as it ensures that the data is appropriately structured and ready for analysis, enabling accurate interpretation and extraction of meaningful insights. It involves transforming the values within cells to a standardised format, ensuring consistency, and facilitating subsequent sub-tasks by eliminating variations in representation~\cite{Ell2017}.

Tables consist of numerous cells with literal values that cannot be directly linked. Literal values encompass various types, including numeric quantities, dates, and coordinates. Multiple data preparation sub-tasks can be employed to clean and format these data types. For instance, numeric quantities within a column can be converted to a joint base unit. For example, values such as 10kg, 100g, and 34t, representing weights, can be interpreted and converted to kilograms. The date is another frequently encountered literal type, often appearing in diverse formats such as ``4 October 1983'', ``4-10-1983'', ``Oct 4, 1983'', ``October 4, 1983'', ``1983/10/4'', or ``1983.10.4.'' Normalising numeric and date values can be challenging. However, it can significantly improve subsequent sub-tasks in the pipeline. Moreover, table cells sometimes contain extraneous values, such as text in brackets or special characters, which can confuse entity-linking algorithms and result in poor annotations. Hence, omitting such values can enhance the reliability and accuracy of the final results.

Indeed, many \ac{sota} approaches in the field recognise the value of the Data Preparation sub-task and incorporate it before proceeding with other sub-tasks in the pipeline~\cite{Goel2012,Cruz2013,Mulwad2013,Munoz2013,Quercini2013,Zhang2013,Ritze2015,Efthymiou2017, Ell2017,Zhang2017,Kacprzak2018,Zhang2018,Chabot2019,Cremaschi2019,Hulsebos2019,Nguyen2019,Steenwinckel2019,Zhang2019,Abdelmageed2020,Azzi2020, Baazouzi2020,Cremaschi2020-1,Cremaschi2020-2,Chen2020,Kim2020,Shigapov2020,Tyagi2020,Yumusak2020,Avogadro2021,Nguyen2021,Abdelmageed2021,Abdelmageed2021-2,Abdelmageed2022,Cremaschi2022,Huynh2020,Huynh2021,Huynh2022,Nguyen2020,Nguyen2021,Baazouzi2021,Gottschalk2022,Wang2021,Zhou2021,avogadro2024feature}. These approaches can also be split into multiple orthogonal categories depending on the type of data preparation technique they perform. The most commonly used techniques are: i)~\textbf{\textit{spell checking}}~\cite{Abdelmageed2020,Abdelmageed2021,Abdelmageed2021-2,Abdelmageed2022,Azzi2020,Chen2020,Kim2020,Yumusak2020}, ii)~\textbf{\textit{units of measurements conversion}}~\cite{Zhang2013,Ritze2015,Ell2017,Cremaschi2019,Baazouzi2020,Cremaschi2020-1}, iii)~\textbf{\textit{cell cleaning}}~\cite{Goel2012,Quercini2013,Munoz2013, Ritze2015,Efthymiou2017,Kacprzak2018,Chabot2019,Cremaschi2019,Cremaschi2022,Cremaschi2020-2,Abdelmageed2021,Abdelmageed2021-2,Abdelmageed2022,Azzi2020,Baazouzi2020,Baazouzi2021,Cremaschi2020-1,Tyagi2020,Avogadro2021,Nguyen2021,Wang2021,avogadro2024feature}, iv)~\textbf{\textit{acronym expansion}}~\cite{Mulwad2013,Ritze2015}, v)~\textbf{\textit{format translation}}~\cite{Cruz2013,Quercini2013,Gottschalk2022}, and vi)~\textbf{\textit{language detection}}~\cite{Ell2017,Nguyen2019}.

The \textbf{\textit{spell checking}} technique involves the automatic detection and correction of spelling errors in the text.
Some approaches employ this technique to clean table content and improve the accuracy and readability of text~\cite{Abdelmageed2020, Azzi2020, Chen2020, Kim2020, Yumusak2020}. The most used method to fix typos in the cells is by invoking autocorrection libraries: JenTab~\cite{Abdelmageed2020,Abdelmageed2021,Abdelmageed2021-2,Abdelmageed2022} invokes Autocorrect library while Azzi et al.~\cite{Azzi2020} invokes Gurunudi and Wikipedia API. In the \ac{sota} there are other libraries for the same purpose:
TextBlob, Spark NLP, Pyspellchecker, Serpapi. LinkingPark~\cite{Chen2020} handles spelling errors by applying a tailored spelling corrector, which performs a one-edit distance check between each cell and a set of candidate entities.~\cite{Kim2020,Yumusak2020} manage errors such as misspellings, incorrect spacing, and omission of special symbols or numbers by crawling through search engines (\eg~Google and Yandex)\footnote{\href{https://pypi.org/project/autocorrect/}{pypi.org/project/autocorrect}, \href{https://github.com/guruyuga/gurunudi}{github.com/guruyuga/gurunudi}, \href{https://wikipedia.readthedocs.io/en/latest/code.html}{wikipedia.readthedocs.io/en/latest/code.html}, \href{https://textblob.readthedocs.io/en/dev/}{textblob.readthedocs.io/en/dev}, \href{https://nlp.johnsnowlabs.com/}{nlp.johnsnowlabs.com}, \href{https://github.com/barrust/pyspellchecker}{github.com/barrust/pyspellchecker}, \href{https://serpapi.com/spell-check}{serpapi.com/spell-check}, \href{https://yandex.com/}{yandex.com}}.

Another Data Preparation sub-step in the \ac{sota} is the \textbf{\textit{units of measurements conversion}} which involves identifying the units in the table and applying appropriate mathematical formulas or conversion factors to convert the values into a standardised unit or a desired unit of measurement. This ensures uniformity and facilitates meaningful data analysis. Several approaches incorporate a numerical conversion~\cite{Zhang2013,Ritze2015,Ell2017,Cremaschi2019,Baazouzi2020,Cremaschi2020-1}. InfoGather+~\cite{Zhang2013} assumes that there is a canonical string for every unit and scale. The system implements a set of conversion rules defined by the system administrator. This process considers three components: left-hand side (LHS), right-hand side (RHS) and $\theta$. LHS and RHS are strings that describe units and scales, while $\theta$ represents the conversion factor. Another approach developed by Ritze et al.~\cite{Ritze2015} normalises units using a set of manually generated conversion rules (around 200). In Ell's approach~\cite{Ell2017}, a conversion process is employed for values that pertain to weights, lengths, volumes, and times. This involves a classic pattern-matching technique, where a unit of measurement follows a numeric value. Also, in MantisTable~\cite{cremaschi2019semantic,Cremaschi2019,cremaschi2019mantistable,Cremaschi2020-1}, unit normalisation is achieved by utilising \ac{regex} based on the rules initially described in InfoGather+~\cite{Zhang2013}, and then it extends them to cover a complete set of units of measurement. The same technique is applied in Kepler-aSI~\cite{Baazouzi2020}.

The most critical and applied technique in the Data Preparation sub-task is \textbf{\textit{cell cleaning}} because it removes or modifies unnecessary or unwanted elements in a cell. In~\cite{Munoz2013}, tables are cleaned and canonicalised (fixing syntax mistakes) using CyberNeko\footnote{\href{https://nekohtml.sourceforge.net/}{nekohtml.sourceforge.net}} while~\cite{Ritze2015,Cremaschi2019,Cremaschi2020-1,Abdelmageed2020,Abdelmageed2021,Abdelmageed2021-2,Abdelmageed2022,Baazouzi2020,Tyagi2020,Nguyen2021} cleans cells by removing HTML artefacts, special characters and additional whitespaces.  Subsequent approaches~\cite{Cremaschi2020-2,Steenwinckel2019,Avogadro2021,Cremaschi2022,Wang2021,avogadro2024feature} use a more straightforward process by removing only parentheses and special characters. During the table loading step, the AMALGAM approach~\cite{Azzi2020} incorporates the capability to clean cells by addressing incorrect encoding through the utilisation of the Pandas library. Several approaches, including JenTab~\cite{Abdelmageed2020}, MTab~\cite{Nguyen2019}, and bbw~\cite{Shigapov2020}, incorporate the ftfy library\footnote{\href{https://github.com/rspeer/python-ftfy}{github.com/rspeer/python-ftfy}} within the cleaning step. This integration allows for the resolution of broken Unicode characters found in various forms, such as transforming ``The Mona Lisa doesnÃƒÂ¢Ã¢â€šÂ¬Ã¢â€žÂ¢t have eyebrows'', which converted to ``The Mona Lisa doesn't have eyebrows''. In the successive MTab4WikiData implementation~\cite{Nguyen2020}, the preprocessing is simpler because of the effectiveness of the fuzzy entity search. Kacprzak et al.~\cite{Kacprzak2018} removes non-numerical chars from numeric columns. In DAGOBAH~\cite{Chabot2019} encoding homogenisation and special characters deletion (parentheses, square bracket and non-alphanumeric characters) are applied to optimise the lookups. The implementation was improved in the subsequent version of DAGOBAH~\cite{Huynh2020,Huynh2021,Huynh2022}.
Some approaches~\cite{Quercini2013,Efthymiou2017,Zhang2017,Cremaschi2019,Cremaschi2020-1,Abdelmageed2020,Abdelmageed2021,Abdelmageed2021-2,Abdelmageed2022,Baazouzi2021} apply stop-word removal in this sub-task.

Tables often include acronyms and abbreviations, shortened terms formed by combining multiple words' initial letters or parts. To address such cases, several approaches employ \textit{\textbf{acronym expansion}}~\cite{Mulwad2013, Ritze2015, Cremaschi2019, Baazouzi2020}. For instance, Mulwad et al.'s approach~\cite{Mulwad2013}, recognises and expands acronyms and stylised literal values like phone numbers. Another approach~\cite{Ritze2015} utilises transformation rules to resolve abbreviations, such as converting ``co.'' to ``company''. MantisTable~\cite{Cremaschi2019} leverages the Oxford English Dictionary\footnote{\href{https://public.oed.com/how-to-use-the-oed/abbreviations/}{public.oed.com/how-to-use-the-oed/abbreviations}} to decipher acronyms and abbreviations. Similarly, Kepler-Asi employs heuristic methods to resolve acronyms and abbreviations~\cite{Baazouzi2020}.

Another possible data preparation technique is \textit{\textbf{format translation}} which consists of translating data into a different structure. For example, Cruz et al.~\cite{Cruz2013} and Quercini et al.~\cite{Quercini2013} translate geographic and temporal information into spatial and time series. In Tab2KG~\cite{Gottschalk2022}, the data is transformed in \ac{rml} format.

Eventually, \textit{\textbf{language detection}} implies detecting the language of the mentions to best address the successive Entity Linking. Ell~\cite{Ell2017} applies some \ac{regex} to detect languages, while~\cite{Nguyen2019} uses pre-trained fastText models\footnote{\href{https://fasttext.cc/docs/en/crawl-vectors.html}{fasttext.cc/docs/en/crawl-vectors.html}} to predict the language of the whole table.

Many approaches do not specify how the data is prepared to be processed~\cite{Hignette2007,Hignette2009,Tao2009,Limaye2010,Mulwad2010,Syed2010,Mulwad2011,Venetis2011,Knoblock2012,Pimplikar2012,Wang2012,Buche2013,Deng2013,Ermilov2013,Zwicklbauer2013,Sekhavat2014,Taheriyan2014,Bhagavatula2015,Ramnandan2015,Ermilov2016,Neumaier2016,Pham2016,Taheriyan2016-1,Taheriyan2016-2,Luo2018,Chen2019,Chen2019learning,Kruit2019,Morikawa2019,Oliveira2019,Takeoka2019,Thawani2019,Chen2020,Eslahi2020,Khurana2020,Guo2020,Li2020,Zhang2020,Heist2021,Steenwinckel2021,Yang2021,Chen2022,Deng2022,Liu2022,Suhara2022}.

\subsection{Column Classification}
\label{subsec:column-classification}

Classifying table columns entails categorising each as a \ac{ne} or a \ac{lit}.~\ac{ne}-columns contain values representing entities such as names, locations, or organisations. In contrast, \ac{lit}-columns contain values representing literal data types such as numbers, dates, or geo-coordinates.
Many existing approaches utilise prior datatype classification to determine the type of columns. By classifying columns, subsequent semantic analysis and data manipulation become more feasible.
These approaches assign specific types (\eg~number, date, geo-coordinate) by employing: i)~\textbf{\textit{\ac{regex} matching}}~\cite{Zhang2017,Efthymiou2017,Cremaschi2019,Cremaschi2020-1,Baazouzi2020,Baazouzi2021,Cremaschi2020-2,Avogadro2021,Cremaschi2022,Nguyen2019,Shigapov2020,avogadro2024feature}, ii)~\textbf{\textit{statistical analysis}}~\cite{Mulwad2013,Kacprzak2018,Guo2020}, or iii)~\textbf{\textit{\ac{ml} techniques}}~\cite{Zhang2020,Deng2022}. Additionally, some approaches prioritise entity linking; in this scenario, unlinked columns are then classified as \ac{lit}-columns.

Some approaches explicitly consider Column Classification as a distinct sub-task~\cite{Mulwad2013,Efthymiou2017,Zhang2017,Kacprzak2018,Cremaschi2019,Baazouzi2020,Baazouzi2021,Cremaschi2020-1,Cremaschi2020-2,Guo2020,Abdelmageed2020,Abdelmageed2021,Abdelmageed2021-2,Abdelmageed2022,Avogadro2021,Nguyen2019,Chen2022,Chabot2019,Huynh2020,Huynh2021,Huynh2022,Cremaschi2022,Zhang2020,Deng2022,avogadro2024feature}, while many others implicitly perform Column Classification by identifying cell or column datatypes~\cite{Hignette2007,Hignette2009,Buche2013,Cruz2013,Quercini2013,Ritze2015,Ramnandan2015,Taheriyan2016-1,Pham2016,Shigapov2020,Kim2020,Alobaid2021,Gottschalk2022,Deng2022,Thawani2019,Nguyen2021,Chen2022}. TableMiner+~\cite{Zhang2017} utilises \textbf{\textit{\ac{regex}}} patterns to identify empty, date, number, and long text columns, categorising them as \ac{lit}-columns, while the rest are considered \ac{ne}-columns. Efthymiou et al.~\cite{Efthymiou2017} applies a column sampling to classify literal columns. Later,~\cite{Cremaschi2019,Cremaschi2020-1,Baazouzi2020,Baazouzi2021,Cremaschi2020-2,Avogadro2021,Cremaschi2022,Nguyen2019,avogadro2024feature} adopt a similar technique, employing additional \ac{regex} patterns and using majority voting to determine column types. In addition, MTab~\cite{Nguyen2019} combines the use of \ac{regex} with SpaCy\footnote{\href{https://spacy.io/}{spacy.io}} pre-trained model to perform column classification. Kim et al.~\cite{Kim2020} consider text, number, and date as possible types while bbw~\cite{Shigapov2020} uses \ac{regex} to predict number, time, name and string datatypes.

Regarding the \textbf{\textit{statistical analysis}}, Mulwad et al.~\cite{Mulwad2013} developed a domain-independent and extensible framework where it is possible to implement components to detect literal values; when all cells in a column are literal, the column is considered as ``No-annotation''.
Kacprzak et al.~\cite{Kacprzak2018} considers as numeric the columns with at least 50\% of numerical values, else the columns are classified as \ac{ne}. Similarly, also~\cite{Guo2020} classifies columns as either ``character'' or ``numeric''.

In the \textbf{\textit{\ac{ml} techniques}}, Zhang et al.~\cite{Zhang2020} proposes a column header classification model which was trained on the T2D dataset. The TURL~\cite{Deng2022} approach uses an additional type embedding vector to differentiate \ac{ne} columns.

Other approaches perform Column Classification, but there are not enough details to categorise them. In DAGOBAH~\cite{Chabot2019} the process aims to identify a first low-level type for each column among five given types (Object, Number, Date, Unit, and Unknown). In the successor~\cite{Huynh2020} numeric values are considered both with or without the corresponding unit of measure. No further changes were applied in consecutive approaches~\cite{Huynh2021,Huynh2022} except adding customised modules for organisation, location and currency detection in 2022~\cite{Huynh2022}. Similarly~\cite{Thawani2019}, considers string, date and numeric types. In~\cite{Takeoka2019} is unclear whether the table columns are already classified as \ac{ne} and \ac{lit}. JenTab~\cite{Abdelmageed2020,Abdelmageed2021,Abdelmageed2021-2,Abdelmageed2022} considers object, date, string and numbers.
LinkingPark~\cite{Chen2022} leverages standard conversion functions for \textit{int}, \textit{float} and \textit{date-time} datatype; otherwise, cells are considered strings. There is no information about the aggregation function, but \ac{lit} and \ac{ne} columns are considered differently.

\subsection{Datatype Annotation}
\label{subsec:datatype-annotation}

In many approaches, the Datatype Annotation of \ac{lit}-columns is tightly linked to the Column Classification. Similarly to the Column Classification sub-task, approaches exploit three methods: i)~\textbf{\textit{statistical analysis}}~\cite{Hignette2007,Hignette2009,Chen2022,Neumaier2016,Ramnandan2015,Pham2016,Venetis2011,Kacprzak2018,Chen2020,Nguyen2020,Ceritli2020}, ii)~\textbf{\textit{\ac{regex} matching}}~\cite{Ritze2015,Quercini2013,Cruz2013,Mulwad2013,Ramnandan2015,Taheriyan2016-1,Pham2016,Zhang2017,Cremaschi2019,Cremaschi2020-1,Cremaschi2020-2,Avogadro2021,Cremaschi2022,Shigapov2020,Baazouzi2020,Baazouzi2021,Nguyen2019,Nguyen2021,Gottschalk2022,avogadro2024feature}, or iii)~\textbf{\textit{\ac{ml} techniques}}~\cite{Takeoka2019,Zhang2013,Thawani2019,Goel2012,Deng2022,Zhang2020,Efthymiou2017,Chen2019,Nguyen2019,Nguyen2021,Gottschalk2022}. In this sub-task it is possible to add an additional category related to approaches that use iv)~\textbf{\textit{other methods}}~\cite{Knoblock2012,Taheriyan2014,Taheriyan2016-1,Taheriyan2016-2}.

Several approaches employ \textbf{\textit{statistical analysis}} by applying a set of rules to classify \ac{lit} and \ac{ne} columns~\cite{Hignette2007,Hignette2009}. Such rules concern how cell content is represented, \ie the amount of text and numbers/units. Inspired by Taheriyan et al.~\cite{Taheriyan2016-1},~\cite{Pham2016} adopts statistical hypotheses as a metric for column annotation. Hierarchical clustering is employed by Neumaier et al.~\cite{Neumaier2016} to construct a background \ac{kg} using DBpedia. The nearest neighbours classification is applied to predict the most probable data type for a given set of numerical values, also considering distribution similarity. NUMBER~\cite{Kacprzak2018} is inspired from previous works~\cite{Neumaier2016,Pham2016}. The evaluation process involves two main aspects. Firstly, the similarity of value distributions is assessed by comparing them to the properties in a target \ac{kg} using a KS test\footnote{The KS test measures the statistical difference between the two distributions, providing insights into their similarity.}. Secondly, the relative difference is computed between numerical values in a column and the numerical values of properties linked to the entities. MTab4Wikidata~\cite{Nguyen2020}, column types are determined after the cells have already been linked to entities. LinkingPark~\cite{Chen2020,Chen2022} uses precomputed statistics for numeric datatypes, such as range, mean, and standard deviation and considers datatypes that match the corresponding ranges as potential candidates. p-type~\cite{Ceritli2020} proposes a model built upon \ac{pfsms}. In contrast to the standard use of \ac{regex}, PFSMs have the advantage of generating weighted posterior predictions even when a column of data is consistent with more than one type of model.

\textbf{\textit{\ac{regex}}} is used to check if the content of the cells can be classified, for instance, as pH, temperature, time, date, number, geo coordinates, iso8601 date, street address, hex colour, URL, image file, credit card, email address, IP address, ISBN (International Standard Book Number), boolean, id, currency, and IATA (International Air Transport Association) codes. If the number of occurrences of the most frequent \ac{regex}Types detected exceeds a given threshold, the column will be annotated as \ac{lit}-column, and the most frequent \ac{regex}Type will be assigned to the column under analysis. Then, to select the datatype to annotate the column, some approaches imply a mapping between \ac{regex}Type and Datatype.~\cite{Quercini2013,Mulwad2013,Cruz2013,Ritze2015,Ramnandan2015,Taheriyan2016-1,Zhang2017,Cremaschi2019,Shigapov2020,Cremaschi2020-1,Cremaschi2020-2,Baazouzi2020,Zhang2020,Avogadro2021,Cremaschi2022,Baazouzi2021,avogadro2024feature} utilise some or the entire list of the \ac{regex} defined above to identify the datatype of the column under analysis.

The last method for Datatype Annotation regards \textbf{\textit{ML techniques}}. Meimei~\cite{Takeoka2019} utilises embeddings by modelling a table with a Markov random field and employing multi-label classifiers to find the correct annotation, similar to InfoGather+~\cite{Zhang2013} which focuses only on numerical values. Another approach~\cite{Goel2012}, models different latent structures within the data and employs a \ac{crf} to perform semantic annotation in different domains (\eg~weather, flight status, and geocoding). The approach ColNet~\cite{Chen2019} utilises a Convolutional Neural Network (CNN) trained on positive and negative samples to differentiate between different column types.

Other approaches combine \textit{\textbf{\ac{regex}}} with \textbf{\textit{\ac{ml} techniques}}; MTab\cite{Nguyen2019} classifies columns as NE or LIT using Duckling \ac{regex}\footnote{\href{https://github.com/facebook/duckling}{github.com/facebook/duckling}} and SpaCy (a pre-trained classificator) with majority voting. Numeric columns are classified using a neural numerical embedding model (EmbNum~\cite{Nguyen2018}) through representation vectors for numerical attributes without prior assumptions on data distribution. In contrast~\cite{Nguyen2021}, used only SpaCy to identify
\ac{lit}-columns. Tab2KG~\cite{Gottschalk2022} uses the Dateparser library\footnote{\href{https://github.com/sisyphsu/dateparser}{github.com/sisyphsu/dateparser}} for classification in numeric, spatial, boolean or text. Later the columns are further classified into more specific categories using \ac{regex} combined with the method described in ~\cite{Alobaid2021}, which implies classifying four kinds of numbers: nominal, ordinal, interval, and ratio. Then fuzzy c-means is used for classification.

Several other approaches in the field of \ac{sti} have been proposed, each with its unique methodology. These approaches, including works such as~\cite{Tao2009,Limaye2010,Mulwad2010,Syed2010,Mulwad2011,Pimplikar2012,Wang2012,Deng2013,Ermilov2013,Munoz2013,Zwicklbauer2013,Sekhavat2014,Bhagavatula2015,Ermilov2016,Ell2017,Luo2018,Zhang2018,Chen2019learning,Hulsebos2019,Morikawa2019,Kruit2019,Oliveira2019,Steenwinckel2019,Zhang2019,Azzi2020,Eslahi2020,Khurana2020,Li2020,Tyagi2020,Yumusak2020,Heist2021,Yang2021,Steenwinckel2021,Wang2021,Zhou2021,Liu2022,Suhara2022}, do not involve semantic classification or Datatype Annotation sub-tasks for the columns. In~\cite{Mulwad2011}, the concept of literal annotation was discussed as a potential area for future works, which was later implemented in~\cite{Mulwad2013}. The approach MAGIC, discussed in~\cite{Steenwinckel2021}, uses entity embeddings with neighbour nodes up to a depth of 2 without explicitly distinguishing between entities and literals. However, it is plausible that this approach can also be extended to include Datatype Annotation. Specific approaches focus on the manual annotation of tables, providing a \ac{ui} that allows users to select the most appropriate datatype manually~\cite{Knoblock2012,Taheriyan2014,Taheriyan2016-1,Taheriyan2016-2}.

Several approaches involve Datatype Annotation; however, they lack sufficient details about how the annotation of the columns with specific datatypes occurs~\cite{Buche2013,Efthymiou2017,Thawani2019,Abdelmageed2020,Abdelmageed2021,Abdelmageed2021-2,Abdelmageed2022,Kim2020,Guo2020,Chabot2019,Huynh2020,Huynh2021,Huynh2022,Deng2022}.

\subsection{Subject Detection}
\label{subsec:subject-detection}

As previously illustrated (Section~\ref{sec:taxonomy}), the S-column is the column among the NE-columns that all the others refer to. Some approaches define it as a ``key'' column that includes entity-based mentions that could potentially be consulted in a \ac{kg}, containing a large number of unique values.

Generally, approaches might employ one of the following techniques for Subject Detection: i)~\textit{\textbf{heuristic approaches}}, ii)~\textit{\textbf{statistical analysis}}, or iii)~\textit{\textbf{\ac{ml} techniques}}.

Regarding \textit{\textbf{heuristic approaches}}, one common method is based on the \textbf{column position} within the table. For example, some approaches designate the leftmost column as the S-column~\cite{Efthymiou2017,Abdelmageed2020,Kim2020,Abdelmageed2021,Abdelmageed2021-2,Abdelmageed2022,Chen2022,Deng2022}.~\cite{Syed2010} instead, involves identifying columns with \textbf{specific labels} as the S-column, such as columns labelled as ``title'' or ``label''.
Another method for Subject Detection is to consider column detection after linking entities. For instance, Zhang et al.~\cite{Zhang2018,Zhang2020} use the column with the highest number of linked entities as an indicator for Subject Detection. Similarly, the approach proposed by Heist et al.~\cite{Heist2021}, considers subject-predicate-object relations between the table's columns and identifies the S-column as the one with the highest number of entities in the subject position.

The use of \textit{\textbf{statistical analysis}} to identify the S-column is found in TableMiner+~\cite{Zhang2017}; it uses a set of rules based on the number of words, the capitalisation, and the mentions of months or days in a week. In MantisTable~\cite{Cremaschi2019, Cremaschi2020-1, Cremaschi2020-2, Avogadro2021,Cremaschi2022,avogadro2024feature} the subject is selected among the NE-columns using a calculated as a set of indicators such as the average number of words in each cell, the fraction of empty cells in the column, the fraction of cells with unique content and the distance from the first NE-column. The same indicators are also used in~\cite{Baazouzi2020}. DAGOBAH2019~\cite{Chabot2019}, TAKCO~\cite{Kruit2019} and MTab2021~\cite{Nguyen2021} consider the fraction of cells with unique content and the column position in the table.

Some works in the literature~\cite{Venetis2011,Ermilov2016} use \textit{\textbf{\ac{ml} techniques}}. TAIPAN~\cite{Ermilov2016} selects SVM and Decision Tree as the best classifiers for this sub-task and uses as features the ratio of cells with disambiguated entities in a column and the number of relations between the columns. It is worth noting that~\cite{Venetis2011} also employs an SVM using features dependent on the name and type of the column and the values in different column cells.

Tab2KG~\cite{Gottschalk2022} uses a graph-based approach for subject detection, but the paper lacks details about the implementation.

Eventually, there are many approaches that do not perform S-column detection~\cite{Hignette2007,Hignette2009,Tao2009,Limaye2010,Mulwad2010,Mulwad2011,Goel2012,Knoblock2012,Pimplikar2012,Wang2012,Buche2013,Cruz2013,Deng2013,Ermilov2013,Mulwad2013,Munoz2013,Quercini2013,Zhang2013,Zwicklbauer2013,Sekhavat2014,Taheriyan2014,Bhagavatula2015,Ramnandan2015,Ritze2015,Neumaier2016,Pham2016,Taheriyan2016-1,Taheriyan2016-2,Ell2017,Kacprzak2018,Luo2018,Chen2019,Chen2019learning,Hulsebos2019,Morikawa2019,Nguyen2019,Oliveira2019,Steenwinckel2019,Takeoka2019,Thawani2019,Zhang2019,Azzi2020,Chen2020,Eslahi2020,Guo2020,Huynh2020,Khurana2020,Li2020,Nguyen2020,Shigapov2020,Tyagi2020,Yumusak2020,Baazouzi2021,Huynh2021,Steenwinckel2021,Wang2021,Yang2021,Zhou2021,Liu2022,Huynh2022,Suhara2022}.

\subsection{Entity Linking}
\label{subsec:entity-linking}

Entity Linking, also known as named entity linking or entity resolution, is an \ac{nlp} task that involves linking named entities mentioned in the text to their corresponding entities in a \ac{kg}. The goal is to identify and disambiguate the entities mentioned in the text and connect them to unique identifiers.

In Entity Linking, a named entity refers to a specific named person, organisation, location, event, or other well-defined entity. For example, in the sentence ``Barack Obama was born in Hawaii'', the named entity ``Barack Obama'' can be linked to the corresponding entry in a \ac{kg}, such as \texttt{dbr:Barack\_Obama} in DBpedia o \texttt{Barack Obama (Q76)} in Wikidata. For this sub-task, approaches can be grouped into three step within entity liking: i)~\textit{\textbf{mention detection}}~\cite{Bhagavatula2015}, ii)~\textit{\textbf{candidate generation}}, and iii)~\textit{\textbf{entity disambiguation}}~\cite{Limaye2010,Syed2010,Mulwad2010,Mulwad2011,Wang2012,Mulwad2013,Zwicklbauer2013,Bhagavatula2015,Ritze2015,Efthymiou2017,Ell2017,Zhang2017,Chabot2019,Chen2019,Cremaschi2019,Kruit2019,Morikawa2019,Nguyen2019,Oliveira2019,Steenwinckel2019,Thawani2019,Abdelmageed2020,Azzi2020,Chen2020,Cremaschi2020-1,Cremaschi2020-2,Eslahi2020,Huynh2020,Kim2020,Nguyen2020,Shigapov2020,Tyagi2020,Zhang2020,Abdelmageed2021,Avogadro2021,Baazouzi2021,Huynh2021,Nguyen2021,Steenwinckel2021,Yang2021,Abdelmageed2022,Chen2022,Cremaschi2022,Huynh2022,Deng2022,Liu2022,avogadro2024feature}.

\textbf{\textit{Mention detection}} refers to the identification and extraction of mentions from tabular data. It involves recognising specific pieces of information within a table representing entities. Detecting mentions in a table can involve various techniques, such as \ac{nlp} methods, \ac{ner} methods, or pattern-matching algorithms. This sub-task analyses the table's content, column headers, and other contextual information to accurately identify and classify the mentions.

The TabEL~\cite{Bhagavatula2015} approach identifies potential mentions within a given cell that can be associated with entities in a \ac{kg}. TabEL identifies the longest phrases within the cell's text content with a non-zero probability of being linked to an entity $e$ according to the probability distribution $P(e|s)$ where $s$ is a phrase. If the length of $s$ is shorter than the length of the cell's text content, TabEL continues searching for the longest phrase. Unfortunately, there are no additional details on this method in the paper.

In the context of \ac{sti}, a crucial role is played by the \textit{candidate generation} sub-task, also known as \textbf{\textit{lookup}}, which refers to the process of identifying potential entities based on the given input or query. When a query is posed, the system must generate a set of candidates for each cell that could potentially satisfy the query. The candidate generation process may utilise various techniques, such as semantic parsing, entity recognition, or information retrieval. It could also leverage \ac{ml} models trained on large data corpora to generate likely candidates based on contextual patterns.
The \textit{Lookup} sub-task can be divided into four methods:
a) \textbf{\textit{custom index}}~\cite{Limaye2010,Syed2010,Efthymiou2017,Ell2017,Chabot2019,Chen2019,Kruit2019,Morikawa2019,Oliveira2019,Thawani2019,Chen2020,Cremaschi2020-2,Huynh2020,Nguyen2020,Huynh2021,Avogadro2021,Nguyen2021,Yang2021,Cremaschi2022,Chen2022,Huynh2022,avogadro2024feature}, b)~\textbf{\textit{external lookup services}}~\cite{Mulwad2010,Mulwad2011,Wang2012,Mulwad2013,Bhagavatula2015,Ritze2015,Zhang2017,Zhang2018,Chabot2019,Cremaschi2019,Cremaschi2020-1,Nguyen2019,Steenwinckel2019,Thawani2019,Abdelmageed2020,Abdelmageed2021,Abdelmageed2022,Azzi2020,Eslahi2020,Chen2020,Kim2020,Shigapov2020,Tyagi2020,Zhang2020,Baazouzi2021,Yang2021,Steenwinckel2021,Deng2022}, c)~\textbf{\textit{hybrid}} (both custom index and external lookup services)~\cite{Chabot2019,Thawani2019,Chen2020,Yang2021}, and d) \textbf{\textit{other}}~\cite{Shigapov2020,Wang2012}.

\textbf{\textit{Custom index}} refers to building a specialised index for specific requirements or use cases. When building a custom index, there is flexibility in defining mappings, analysers, and other configurations based on specific needs. One of the most adopted solutions is Elasticsearch\footnote{\href{https://www.elastic.co/}{www.elastic.co}}, a robust and scalable search and analytics engine. It uses a document-oriented approach, where data is organised and stored as JSON documents. Several approaches rely on Elasticsearch for the lookup sub-task~\cite{Chabot2019,Kruit2019,Huynh2020,Huynh2021,Huynh2022,Cremaschi2020-2,Avogadro2021,Cremaschi2022,Morikawa2019,Oliveira2019,Thawani2019,Chen2020,Chen2022,avogadro2024feature}. The simplest index can incorporate entity labels (\texttt{rdfs:label}) or aliases (\texttt{skos:altLabel}). However, some approaches also add abbreviations (\texttt{dbo:abbreviation}), descriptions (\texttt{rdfs:comment})~\cite{Huynh2021,Huynh2022,Thawani2019}, or, indexes for specific entity types, name (\texttt{foaf:name}), surname (\texttt{foaf:surname}), and given name (\texttt{foaf:givenName})~\cite{Morikawa2019}. The MantisTable team builds a separate system named LamAPI\footnote{\href{https://github.com/unimib-datAI/lamAPI}{github.com/unimib-datAI/lamAPI}}~\cite{Avogadro2022,avogadro2024feature} which is used across multiple versions of this system. LamAPI\footnote{\href{https://lamapi.datai.disco.unimib.it/}{lamapi.datai.disco.unimib.it/}} tool retrieves entities with the highest similarity between the mention in the cell and the entity's label by combining different search strategies, such as full-text search based on tokens, n-grams and fuzzy search. Other approaches build custom indexes using different solutions; Limaye et al.~\cite{Limaye2010} presents a method that utilises a catalogue which comprises types, entities and relations. Entities in the catalogue are associated with lemmas, which are canonical strings extracted from Wikipedia, or synset names from WordNet\footnote{\href{https://wordnet.princeton.edu/}{wordnet.princeton.edu}}. Syed~\cite{Syed2010} develops a hybrid \ac{kg} of structured and unstructured information extracted from Wikipedia augmented by \ac{rdf} data from DBpedia and other Linked Data. The system is called Wikitology and uses an \ac{ir} index (Lucene) to represent Wikipedia articles. The same technique is used by Mulwad et al.\cite{Mulwad2010,Mulwad2011,Mulwad2013}. Ell et al.~\cite{Ell2017} creates an index for each type of resource (entity, property, type) for each language. These indexes contain the names of the resources, according to DBpedia. For properties and classes, the names are obtained from the \texttt{rdfs:label} property in DBpedia. Efthymiou et al.~\cite{Efthymiou2017} utilises a lookup-based method to establish connections. It leverages the limited entity context available in Web tables to identify correspondences with the \ac{kg}. The approach builds its custom search index over Wikidata, called \textit{FactBase}, consisting of entities with corresponding IDs and textual descriptions. Another system named ColNet~\cite{Chen2019} involves two steps in its candidate generation. Firstly, a lookup step is performed to retrieve entities from the \ac{kg} by matching cells based on entity labels and anchors (\eg~Wikipedia link) using a lexical index composed of terminology and assertions from the \ac{kg}. MTab4Wikidata~\cite{Nguyen2020}, another version of MTab~\cite{Nguyen2019}, focuses on annotating cells to Wikidata entities. It starts by downloading and extracting a Wikidata dump to build an index using hash tables. The lookup process is then performed using a fuzzy search. The result is a ranking list of entities based on edit distance scores. In the updated version of MTab 2021~\cite{Nguyen2021}, a WikiGraph index is constructed, combining Wikidata, Wikipedia, and DBpedia. The lookup uses Keyword Search, Fuzzy Search, and Aggregation Search. GBMTab~\cite{Yang2021} tackles candidate entity generation by differentiating the entities extraction from Wikidata and DBpedia. Only for DBpedia, the approach builds an index using hash tables. Then, it uses the Levenshtein distance to calculate a string similarity between mentions and entities.

The second method employed for candidate generation uses \textbf{\textit{external lookup services}}. This process refers to using a separate service or system to perform lookup or queries for retrieving specific information or data. The external lookup service usually utilises entity recognition, entity disambiguation, or semantic matching techniques. It may consider factors like textual similarity, context, or other relevant information. In the \ac{sti}, many services can be used to extract a set of possible entities given a string as input. The choice of the service depends on the specific requirements and context, such as the \ac{kg} used to annotate the entities. Most approaches annotate table cells to DBpedia entities by using related services, such as DBpedia API~\cite{Nguyen2019,Chabot2019}, DBpedia Lookup Service~\cite{Nguyen2019,Steenwinckel2019,Tyagi2020} and DBpedia Spotlight~\cite{Steenwinckel2019,Steenwinckel2021}. The same occurs for Wikidata, for which the following services are employed: Wikidata API~\cite{Chabot2019,Thawani2019,Azzi2020,Shigapov2020,Steenwinckel2021}, Wikidata Lookup Service~\cite{Nguyen2019,Abdelmageed2020,Eslahi2020,Tyagi2020,Abdelmageed2021,Abdelmageed2021-2,Abdelmageed2022,Deng2022} and Wikidata CirrusSearch Engine~\cite{Chabot2019}. Other services used to execute the lookup sub-task are Wikipedia API~\cite{Chabot2019,Nguyen2019,Shigapov2020,Zhang2020}, MediaWiki API~\cite{Chen2020,Yang2021} and Wikibooks~\cite{Shigapov2020}. Instead of explicitly using lookup services, some approaches perform SPARQL queries, a query language used to retrieve and manipulate data stored in \ac{rdf} format. This method is the default way to obtain information from triple stores. For the approaches that do not provide any specific information on the lookup service, it is assumed that SPARQL is employed. For instance, such queries are used to retrieve entities from YAGO~\cite{Bhagavatula2015}, DBpedia~\cite{Ritze2015,Zhang2017,Zhang2018,Cremaschi2019,Cremaschi2020-1,Steenwinckel2019,Abdelmageed2021,Abdelmageed2021-2,Baazouzi2021,Abdelmageed2022} and Wikidata~\cite{Kim2020,Shigapov2020,Abdelmageed2021,Abdelmageed2021-2,Baazouzi2021,Abdelmageed2022}. Furthermore, other sources are used, for instance, SearX~\cite{Shigapov2020} and Probase~\cite{Wang2012} where pattern matching is used to extract triples.

\textbf{\textit{Entity disambiguation}} refers to the process of resolving ambiguous mentions to entities. When tables contain references to entities or mentions, such as names of people, locations, or organisations, there can be ambiguity if the same name refers to multiple entities. Entity disambiguation in \ac{sti} aims to identify and disambiguate these entity mentions, ensuring that each mention is correctly linked to the appropriate entity. This sub-task can be performed by applying multiple techniques: a)~\textbf{\textit{embedding}}~\cite{Efthymiou2017,Chabot2019,Eslahi2020,Zhang2018,Zhang2020,Steenwinckel2021,Liu2022}, b)~\textbf{\textit{similarity}}~\cite{Limaye2010,Zhang2017,Oliveira2019,Kruit2019,Steenwinckel2019,Thawani2019,Cremaschi2019,Cremaschi2020-1,Cremaschi2020-2,Huynh2020,Tyagi2020,Shigapov2020,Avogadro2021,Abdelmageed2021,Huynh2021,Chen2022,Cremaschi2022,avogadro2024feature}, c)~\textit{\textbf{contextual information}}~\cite{Syed2010,Ritze2015,Efthymiou2017,Chen2019,Nguyen2019,Thawani2019,Nguyen2020,Nguyen2021,Morikawa2019,Azzi2020,Chen2020,Cremaschi2020-2,Abdelmageed2020,Huynh2020,Avogadro2021,Baazouzi2021,Huynh2021,Huynh2022,Chen2022,Cremaschi2022,Liu2022,avogadro2024feature}, d)~\textbf{\textit{\ac{ml} techniques}}~\cite{Mulwad2010,Thawani2019,Zhang2020,avogadro2024feature}, e)~\textbf{\textit{language models}}~\cite{Li2020,Huynh2022,Deng2022,pan2024unifying,lee2018pre,ouyang2022training,touvron2023llama}, f)~\textbf{\textit{probabilistic models}}~\cite{Mulwad2011,Mulwad2013,Bhagavatula2015,Kruit2019,Yang2021}, and g)~\textbf{\textit{other}}~\cite{Wang2012,Munoz2013,Ell2017}.

In graphs and natural language, \textbf{\textit{embedding}} refers to representing nodes in a graph, or words in a text, as dense vectors in a continuous vector space. These embeddings capture semantic and structural relationships between nodes or words, allowing \ac{ml} models to perform tasks such as node classification, link prediction, document similarity, sentiment analysis, and more~\cite{pilehvar2020embeddings}.
Some approaches use embedding techniques to create vector representations of entities~\cite{Efthymiou2017,Chabot2019,Eslahi2020,Zhang2020,Steenwinckel2021}. Every approach tries to capture context information about entities in the \ac{kg} and to incorporate that information in the vector representation.~\cite{Efthymiou2017,Zhang2018,Eslahi2020} employ semantic embeddings obtained through Word2Vec~\cite{Mikolov2013word2vec} on \ac{kgs}, while Zhang et al.~\cite{Zhang2018,Zhang2020} use GloVe~\cite{pennington2014glove}, Wikipedia2Vec~\cite{yamada2018wikipedia2vec} and RDF2Vec~\cite{ristoski2016rdf2vec} to obtain a representation of entities. DAGOBAH~\cite{Chabot2019} uses pre-trained Wikidata embedding~\cite{han-etal-2018-openke}, while MAGIC~\cite{Steenwinckel2021} uses INK technique~\cite{steenwinckel2022ink}, which transforms the local neighbourhood of a node in the \ac{kg} into a structured format. Radar Station~\cite{Liu2022} uses the PyTorch-BigGraph~\cite{lerer2019pytorch} framework for training embeddings.

The entity disambiguation sub-task could involve the computation of some \textbf{\textit{similarities}} among textual data. This type of score is usually adopted by lookup services to retrieve a ranked list of candidates. Often, the disambiguation step involves the selection of the winning candidate by considering the string similarity between the entity label and mention. Some approaches use similarity, such as, Levenshtein distance~\cite{Cremaschi2019,Steenwinckel2019,Cremaschi2020-1,Cremaschi2020-2,Oliveira2019,Thawani2019,Huynh2020,Shigapov2020,Avogadro2021,Huynh2021,Chen2022,Cremaschi2022,avogadro2024feature}, Jaccard similarity~\cite{Limaye2010,Kruit2019,Avogadro2021,Cremaschi2022,avogadro2024feature}, Cosine similarity~\cite{Limaye2010,Tyagi2020} and similarity based on \ac{regex}~\cite{Huynh2020}. Limaye et al.~\cite{Limaye2010} in addition to Jaccard applies also the TF-IDF\footnote{The weight assigned to a term in a document vector is the product of its term frequency (TF) and inverse document frequency (IDF).}. TableMiner+~\cite{Zhang2017} measures the similarity between the \ac{bow} representation of the entity and the \ac{bow} representations of different types of cell contexts, such as row content and column content.

\textbf{\textit{Contextual information}} during the \ac{cea} task considers the surrounding context of a table cell, such as neighbouring cells, column headers, or header row. Contextual information provides additional clues or hints about the meaning and intent of the mention. By analysing the context, a system can better understand the semantics of the cell and make more accurate annotations. Contextual information at column and row level is usually provided by \ac{cta} and \ac{cpa} tasks, but some approaches take column types and properties into account to disambiguate entities even if those tasks are not explicitly treated. Column types are used to disambiguate entities by assuming that entities in a column share the same type. Most approaches rely on this assumption to perform this step~\cite{Syed2010,Ritze2015,Efthymiou2017,Chen2019,Nguyen2019,Morikawa2019,Thawani2019,Azzi2020,Baazouzi2021}.~\cite{Syed2010,Chen2019,Baazouzi2021} limit the number of candidate entities by executing a new lookup query that includes predicted types.~\cite{Ritze2015,Efthymiou2017,Nguyen2019,Morikawa2019,Thawani2019,Azzi2020} refine the candidates set by filtering out entities that do not match the predicted type at the column level.
Similarly, the other assumption considered by some approaches is that contextual information from \ac{cpa} at the row level enables understanding the data better within its broader context~\cite{Huynh2020,Nguyen2020,Nguyen2021,Huynh2022}.~\cite{Huynh2020,Huynh2022} consider the semantic relations between columns by boosting the scores for each candidate entity when the relation is found.~\cite{Nguyen2020,Nguyen2021} compute context similarity between candidate triples and table row values by ranking entities based on this score. Eventually, it selects the candidate with the highest context similarity as the final annotation. Most approaches also implement the disambiguation sub-step by adopting a hybrid solution (considering the information provided by \ac{cpa} and \ac{cta} tasks)~\cite{Chen2020,Chen2022,Avogadro2021,Cremaschi2020-2,Cremaschi2022,Abdelmageed2020,Huynh2021,avogadro2024feature}.
Radar Station~\cite{Liu2022} focuses on improving entity linking in the context of \ac{sti} systems. It addresses disambiguation challenges by using graph embeddings to identify similarities between entities, types, and relationships within tables. The method involves constructing a KD tree of context entities for each column and using it to select the K nearest context entities during prediction, ultimately enhancing the ranking of candidates provided by~\cite{Nguyen2020,Huynh2020,Shigapov2020}.

Other methods that can be employed are \textbf{\textit{\ac{ml} techniques}}. These techniques typically involve training a \ac{ml} model on a labelled dataset where cells are annotated with their corresponding entities. The model learns patterns and relationships between the cells content and their associated entities. To predict the most appropriate entity, \ac{ml} techniques consider various cell features, such as the textual content, context, neighbouring cells, and other relevant information. Several \ac{ml} techniques can be employed to perform the disambiguation task, such as \ac{svm}~\cite{Mulwad2010}, \ac{nn}~\cite{Thawani2019,avogadro2024feature} and Random Forest~\cite{Zhang2020}. Mulwad et al.~\cite{Mulwad2010} create a vector of features for each entity, and then an \ac{svm} is used to rank such vectors. Then a second \ac{svm} decides whether to link or not the entity mentioned in the cell. Thawani et al.~\cite{Thawani2019} build a \ac{nn} that learns adaptive weights and relationships from labelled data. They use a 2-layer architecture with ReLU activation to obtain scores for each candidate. Zhang et al.~\cite{Zhang2020} extract two sets of features: lexical similarity (\eg~Levenshtein, Jaccard) and semantic similarity (\eg~Wikipedia search rank). Eventually, a Random Forest is trained to determine if it is possible to link an entity to a mention.


\textbf{\textit{Language models}} might be used for cell entity annotation by leveraging advanced models, such as BERT. Language models are trained on vast amounts of text data and have the ability to interpret natural language. They can capture complex linguistic patterns, semantic relationships, and contextual cues. These models can be utilised to perform various \ac{nlp} tasks, including entity recognition and linking. In the context of \ac{cea}, an \ac{llm} can be fine-tuned or adapted to this specific task. This involves training the model on a labelled dataset where cells are annotated with corresponding entities. Once the \ac{llm} is trained, it can be applied to unlabelled cells in a table to predict the most likely entity annotations. The model considers the cell's textual content, surrounding context, and potentially other relevant features to make these predictions. Recently \ac{llm}s have been employed to find semantic correlations between different cells at column and row level. The main advantage of using \ac{llm}s is having a contextualised representations for each cell, considering the mention and table metadata. The advent of \textbf{\textit{\ac{llms}}} has led to a new category of approaches for table interpretation. Based on the architecture structure of \ac{llms}, these approaches can be categorised into three groups: i) encoder-decoder \ac{llms}, ii) encoder-only \ac{llms}, and iii) decoder-only \ac{llms}~\cite{pan2024unifying}. Indeed, shortly after the first edition of SemTab, some works~\cite{Li2020,Deng2022,Suhara2022} applied encoder-only \ac{llms} to table interpretation. Although they did not participate in or compare with the SemTab challenge, they created a different experimental setting. During the SemTab2022 instead, a BERT-based~\cite{devlin-etal-2019-bert} model was combined with a more traditional approach~\cite{Huynh2022}. 
More recently, after the release of GPT-3.5~\cite{ouyang2022training} and open-source decoder-only \ac{llms} such as LLAMA~\cite{touvron2023llama} and LLAMA 2~\cite{Touvron2023}, some works have begun applying encoder-based \ac{llms} to table interpretation~\cite{Li2023,Zhang2023}. In SemTab2023, a new decoder-only model was presented that uses BERT~\cite{Dasoulas2023}.

Starting from encoder-based approaches, Ditto~\cite{Li2020} utilises Transformer-based language models to perform a slightly different task; in fact, the goal is entity-matching between various tables. TURL~\cite{Deng2022} leverages a pre-trained TinyBERT~\cite{jiao-etal-2020-tinybert} model to initialise a structure-aware Transformer encoder. Doduo~\cite{Suhara2022} performs \ac{cta} using a pre-trained language model, precisely fine-tuning BERT model on serialised tabular data. DAGOBAH SL 2022~\cite{Huynh2022} employs an ELECTRA-based cross-encoder, a variant of the BERT model. The Cross Encoder takes a concatenated input, including left-side table headers, the target table header, right-side table headers, and the entity description. TorchicTab~\cite{Dasoulas2023} is composed of two sub-systems:  TorchicTab-Heuristic and TorchicTab-Classification. The classification model utilises Doduo~\cite{Suhara2022}.

Regarding decoder-based approaches, TableGPT~\cite{Li2023} performs several tasks, including entity linking using GPT. TableLlama~\cite{Zhang2023}, performs \ac{cea}, along with several other tasks, creating a multi-task dataset for tabular data, in which the entity linking sub-dataset derives from the TURL~\cite{Deng2022} dataset, and using it to fine-tune LLama2~\cite{Touvron2023}. The advent of \textbf{\textit{\ac{llms}}}  has also led to the publication of experiments comparing different approaches based on \ac{llm} and \ac{ml}~\cite{belotti2024evaluating}.

\textbf{\textit{Probabilistic models}} are frameworks for representing and reasoning under uncertainty using probability theory. These models vary in their representation of dependencies and use diverse graphical structures. Several \ac{pgm} can be also used to resolve the disambiguation task, such as Markov models~\cite{Mulwad2011,Mulwad2013,Bhagavatula2015,Yang2021} or Loopy Belief Propagation (LBP)~\cite{Kruit2019}. Markov models focus on sequential dependencies, while LBP employ message passing between nodes in the \ac{pgm}s. Bhagavatula et al.~\cite{Bhagavatula2015} adopt a representation for tables as graphical models. Within this approach, every mention in the table is linked to a discrete random variable that represents the possible candidate entities associated with that mention using Independent Component Analysis. Mulwad et al.~\cite{Mulwad2011,Mulwad2013} resolve ambiguities in table cell values by looking at the evidence from other values in the same row. This is achieved by creating edges between each pair of cell values within a specific row. Kruit et al.~\cite{Kruit2019} introduce a \ac{pgm} incorporating label similarities as priors. The model subsequently improves likelihood scoring to enhance the consistency of entity assignments across rows through Loopy Belief Propagation (LBP). Eventually, Yang et al.~\cite{Yang2021} create a disambiguation graph that utilises mentions and their corresponding candidates from the same row or column in a table. The approach scores the semantic connections between nodes using three features in the \ac{pgm}: Prior, Context, and Abstract.

\textbf{\textit{Other}} approaches cannot be categorised in one of the previous groups. For instance, Munoz et al.~\cite{Munoz2013} proposes an approach to extract RDF triples from Wikitables by linking each cell to DBpedia entities. The process involves following internal links within Wikipedia tables, as they can be directly mapped to DBpedia. Ell et al.~\cite{Ell2017} create some hypotheses for entities extracted in the previous phase (Candidate Generation). These hypotheses include the entity type, the URI in DBpedia, and a confidence value. The confidence value is determined by normalising the frequency value of the entity by dividing it by the sum of frequency values for all the candidates. Wang~\cite{Wang2012} describes the process of understanding a table using the Probase knowledge API. Eventually Kim et al.~\cite{Kim2020} remove candidates considering the content unrelated to the annotation.

Some other approaches such as~\cite{Zwicklbauer2013,Hulsebos2019,Zhang2019,Khurana2020,Wang2021,Zhou2021} do not perform Entity Linking tasks specifically.

\subsection{Type Annotation}
\label{subsec:type-annotation}

The Type Annotation sub-task involves assigning a specific type from a reference \ac{kg} to each NE-column. Various approaches have been developed to address this sub-task, focusing on leveraging Entity Linking techniques, partially or comprehensively, to identify the most frequent column type. This is particularly crucial in unsupervised classification scenarios. Additionally, some approaches consider the information provided by column headers to determine the most suitable type.

The most used methods to annotate NE-columns are: i)~\textbf{\textit{majority voting}}~\cite{Hignette2007,Hignette2009,Mulwad2010,Mulwad2011,Venetis2011,Quercini2013,Zwicklbauer2013,Ritze2015,Ermilov2016,Ell2017,Zhang2017,Azzi2020,Baazouzi2020,Chen2020,Eslahi2020,Kim2020,Nguyen2020,Shigapov2020,Tyagi2020,Nguyen2021,Chen2022,Syed2010,Mulwad2013,Cremaschi2019,Steenwinckel2019,Thawani2019,Cremaschi2020-1,Cremaschi2020-2,Avogadro2021,Cremaschi2022,Abdelmageed2020,Abdelmageed2021,Abdelmageed2021-2,Abdelmageed2022,Kruit2019,Zhou2021,avogadro2024feature}, ii)~\textbf{\textit{Term Frequency-Inverse Document Frequency (TF-IDF)}}~\cite{Limaye2010,Pimplikar2012,Ramnandan2015,Taheriyan2016-1,Oliveira2019,Chabot2019}, iii)~\textbf{\textit{statistical methods}}~\cite{Nguyen2019,Huynh2021,Huynh2020,Heist2021,Deng2013,Huynh2022}, iv)~\textbf{\textit{machine learning}}~\cite{Chen2019,Guo2020,Steenwinckel2021,Wang2021,Yang2021,Deng2022,Gottschalk2022,Chen2019learning,Hulsebos2019,Zhang2019,Suhara2022}, or v)~\textbf{\textit{other}}~\cite{Knoblock2012,Zhang2013,Taheriyan2014,Buche2013,Takeoka2019,Baazouzi2021,Wang2012,Morikawa2019,Khurana2020,Zhang2020}.

Most research studies on column-type prediction utilise a common strategy called \textbf{\textit{majority voting}}. This method involves determining the most frequently occurring type in a column and deciding based on the majority. This pure decision-making method is applied by~\cite{Hignette2007,Hignette2009,Mulwad2010,Syed2010,Ermilov2016,Mulwad2011,Venetis2011,Quercini2013,Zwicklbauer2013,Ritze2015,Ell2017,Zhang2017,Azzi2020,Baazouzi2020,Eslahi2020,Chen2020,Kim2020,Nguyen2020,Shigapov2020,Tyagi2020,Nguyen2021,Yang2021,Chen2022}. Some approaches go beyond simple majority voting and incorporate additional mechanisms to address specific situations. For instance,~\cite{Mulwad2013,Cremaschi2019,Steenwinckel2019,Thawani2019,Cremaschi2020-1,Cremaschi2020-2,Avogadro2021,Cremaschi2022,Kruit2019,Zhou2021} set a threshold to prevent annotating a column when there is insufficient confidence about the type. The approach in~\cite{Quercini2013} applies a deduplication process to remove duplicate types within a column. After deduplication, the type frequencies in the column are summed using a logarithmic function which measures the overall frequency and importance of the types present in the column. In Kruit et al.~\cite{Kruit2019}, majority voting across types is also used to select candidate entities for individual rows using Loopy Belief Propagation (LBP). This approach highlights how \ac{cta} and \ac{el} are interconnected. Also, for LinkingPark~\cite{Chen2020,Chen2022}, the primary method used is majority voting to obtain the most common (\ie most frequent) type. In case multiple candidates have the same frequency, the type selected for annotation is the most specific in the \ac{kg}. For bbw~\cite{Shigapov2020}, majority voting is the primary selection algorithm. However, when multiple types have equal frequency, it selects the first common ancestor type in the \ac{kg}. JenTab~\cite{Abdelmageed2020,Abdelmageed2021,Abdelmageed2021-2,Abdelmageed2022} uses various techniques, including the Least Common Subsumer (LCS), direct parents (\ie majority voting), and popularity. Following the ontology hierarchy, the LCS represents the most specific type, obtained by excluding types occurring in less than 50\% of the cells.
Zhou~\cite{Zhou2021} selects the annotation based on level 2 and level 3 DBpedia classes.

Another method for Type Annotation is based on \textbf{\textit{TF-IDF}}, as mentioned by~\cite{Ramnandan2015,Taheriyan2016-1,Oliveira2019}. DAGOBAH~\cite{Chabot2019} highlights the importance of setting a confidence threshold when using TF-IDF to avoid wrong annotations.
There are also some variations; for example, Limaye~\cite{Limaye2010} introduces the Least Common Ancestor (LCA) as a baseline approach against majority voting and the collective approach. The collective approach considers \ac{el} output, features like Inverse Document Frequency (IDF), and the distance calculated by counting edges between the considered entity and the potential column types obtained in LCA method. In the end, a \ac{pgm} is used for choosing final annotations (CEA, CTA, and CPA). Also~\cite{Pimplikar2012}, uses a \ac{pgm} combined with the TF-IDF approach.

There are approaches that use \textbf{\textit{statistical methods}} other than majority voting; for example, Nguyen~\cite{Nguyen2019} introduces the concept of ``type potential'' which is computed as the cumulative probability entities in a column corresponding to a specific type within the \ac{kg}. ``Type potential'' considers statistics from numerical columns, types from the candidate entities in the whole table, SpaCy type predictions and header values to assess the likelihood of different types for the column.
Similarly~\cite{Huynh2020,Huynh2021,Huynh2022} consider various factors such as frequency, accumulated level (ontology hierarchy), and accumulated rank of Wikidata\footnote{\href{https://www.wikidata.org/wiki/Help:Ranking}{www.wikidata.org/wiki/Help:Ranking}}, for all candidate types of a target column. The final column type is determined using majority voting as the deciding factor. Heist et al.~\cite{Heist2021} compute type frequency and relation frequency statistics in the DBpedia \ac{kg} to identify best-suited types using co-occurrence. Deng~\cite{Deng2013} instead employs an implementation similar to the idea of majority voting using an overlap similarity between top-k candidate types for a given column. In the map-reduce like implementation, the overlap corresponds to the count of entities having a given type.

Another method implies using \textbf{\textit{machine learning}}. The approach presented in~\cite{Chen2019learning} focuses on annotating columns that consist of phrases. For instance, the type \texttt{dbo:Company} can annotate a column containing ``Google, Amazon and Apple Inc.''. To achieve this, they propose a method called \ac{hnn} that captures the contextual semantics of a column. The \ac{hnn} model uses a bidirectional \ac{rnn} and an attention layer (Att-BiRNN) to embed the phrases within each cell, allowing for contextual understanding. A similar \ac{cnn} configuration combined with majority voting is also used in~\cite{Chen2019}.
Sherlock~\cite{Hulsebos2019} is a multi-input deep \ac{nn} for detecting types. It is trained on 686,765 data columns retrieved from the VizNet\footnote{\href{https://viznet.media.mit.edu/}{viznet.media.mit.edu}} corpus by matching 78 semantic types from DBpedia to column headers. Each matched column is characterised by 1,588 features describing the statistical properties, character distributions, word embeddings, and paragraph vectors of column values. Inspired by Sherlock, Sato~\cite{Zhang2019} incorporates table context into semantic type detection. It employs a hybrid model that combines ``signals'' from the global context (values from the entire table) and the local context (predicted types of neighbouring columns). Guo et al.~\cite{Guo2020} introduce a \ac{hnn} model for single-column type annotation, which combines \ac{dl} with a \ac{pgm}~\cite{Koller2009PGM}. With a pre-annotated dataset, a co-occurence matrix is built, considering types for each column pair. This co-occurrence measure is used to annotate similar column pairs in other tables.
TCN~\cite{Wang2021} treats a collection of tables as a graph, with cells as nodes and implicit connections as edges. The connections are cells with the same content or position in different tables. Using graph \ac{nn}, TCN learns a table representation for predicting column types and relations. In the TURL approach~\cite{Deng2022}, the column header and the embedding representation of entities linked to the cells within the column are considered to determine the final annotation. Similarly, Tab2KG~\cite{Gottschalk2022} creates a domain profile from the \ac{kg} and uses it together with a table profile to generate mappings using Siamese Networks between the column content and the types in the \ac{kg}. A domain profile associates relations with feature vectors representing data types and statistical characteristics such as value distributions.
Doduo~\cite{Suhara2022} performs Type Annotation using a pre-trained language model, precisely fine-tuning the BERT model on serialised tabular data. Each column is encoded by appending a special token $[CLS]$ at the beginning, and the resulting embedding representation serves as the contextualised column representation. Column types are predicted using a dense layer followed by an output layer with a size corresponding to the number of column types.

\textbf{\textit{Other approaches}} cannot be classified in the previously mentioned methods; for instance, in~\cite{Knoblock2012,Taheriyan2014} a \ac{crf} is employed to make statistical predictions considering the context, such as column name and values. Similarly,~\cite{Takeoka2019} uses a Markov Network with three potential functions: column-content (similarity between observed cells and a candidate column type), column-column (the similarity between the candidate type of a column and the currently assigned type of other columns), and title-column (similarity between the title and column type). Other approaches such as~\cite{Buche2013,Zhang2013} compute some conditional features considering the presence of specific matches in the column title and content. Another approach~\cite{Wang2012} uses a \ac{kg} taxonomy constructed using type-entity extraction patterns. These patterns, such as ``What is the \textit{A} of \textit{I}?'', where A represents the seed attributes to be discovered and I is an entity in type C obtained from the Probase \ac{kg}~\cite{Wang2011probase}. Those patterns are compared against the Probase's large web corpus of 50 terabytes. Similarly, in LOD4ALL~\cite{Morikawa2019} two scores are used to determine column types. One score prioritises the most frequent ancestor types across the table, while the other emphasises the most specific entity types for each entity. On the other hand, C\textsuperscript{2}\cite{Khurana2020}, a patented approach~\cite{Khurana2020patent}, addresses the type mapping by optimising smaller likelihood problems to reduce the number of candidate types considered. The approach starts with independently finding the top candidates for NE-columns. Best candidate entities are used as pivots to narrow the search, while other features such as \ac{bkg}, diversity (reward deduplication during the scoring), tuple validation (type co-occurrence), and belief sharing (column headers from different tables) are used to enhance prediction accuracy. Zhang et al.\cite{Zhang2020} adopts a different strategy by leveraging the column label and column values to train a classifier for each type present in the \ac{kg}. This classifier is then used to predict the most suitable type for the column based on the learned patterns and characteristics.
Eventually, a naive approach is applied in Kepler-aSI~\cite{Baazouzi2021}, for columns with more than one candidate type no annotation is provided.

There are cases like~\cite{Cruz2013,Baazouzi2020,Steenwinckel2021,Nguyen2021,Yumusak2020} where the specific details about the column annotation sub-task are not provided, making it difficult to understand the exact methodology employed. This is in part because their code is not provided in open source as discussed in Section~\ref{sec:licence}.

Finally, some approaches do not deal with annotations on columns specifically, such as~\cite{Tao2009,Goel2012,Ermilov2013,Munoz2013,Sekhavat2014,Bhagavatula2015,Neumaier2016,Pham2016,Taheriyan2016-2,Efthymiou2017,Kacprzak2018,Luo2018,Zhang2018,Li2020,Liu2022}.


\subsection{Predicate Annotation}
\label{subsec:property-annotation}

The Predicate Annotation sub-task can be challenging, primarily due to the incompleteness of public \ac{kgs} such as DBpedia or Wikidata. Additionally, the Predicate Annotation sub-task can be further categorised into two parts: NE relations, which focus on the relationship between the subject column (S-column) and a named entity (NE) column, and LIT relations, which involve the relationship between the subject column (S-column) and a literal (LIT) column.

The approaches can be categorised into: i)~\textbf{\textit{ruleset}}~\cite{Hignette2007, Hignette2009}, ii)~\textbf{\textit{pattern matching}}~\cite{Tao2009,Wang2012,Sekhavat2014,Deng2022}, iii)
\textbf{\textit{majority voting}}~\cite{Limaye2010,Syed2010,Mulwad2010,Mulwad2011,Venetis2011,Knoblock2012,Mulwad2013,Munoz2013,Buche2013,Ritze2015,Ermilov2016,Chabot2019,Cremaschi2019,Morikawa2019,Thawani2019,Kruit2019,Steenwinckel2019,Cremaschi2020-1,Baazouzi2020,Kim2020,Nguyen2020,Shigapov2020,Yumusak2020,Huynh2020,Baazouzi2021,Steenwinckel2021,Heist2021,Huynh2021,Chen2022,Cremaschi2022,Huynh2022,avogadro2024feature}, iv)~\textbf{\textit{statistical }}~\cite{Taheriyan2014,Taheriyan2016-1,Taheriyan2016-2,Zhang2017,Nguyen2019,Chen2020,Abdelmageed2020,Cremaschi2020-2,Abdelmageed2021,Avogadro2021,Zhang2020,Abdelmageed2021-2,Abdelmageed2022},
and v)~\textbf{\textit{embedding}}~\cite{Chen2019learning,Wang2021,Suhara2022}.

The \textbf{\textit{ruleset}} method is mainly applied in the initial approaches~\cite{Hignette2007,Hignette2009}. The ruleset method consists of a set of rules used to calculate a similarity score between the properties in the \ac{kg} and the column types. In case the table title is provided it is taken into consideration for the score.

The \textbf{\textit{pattern matching}} technique, used by~\cite{Tao2009,Wang2012,Sekhavat2014}, consists of searching for exact matches of subject and object pairs in a large corpus to retrieve possible properties. Similarly, TURL~\cite{Deng2022} aims to extract relations between columns without entity linking. It treats the concatenated table metadata as a sentence and considers the headers of the two columns as entity mentions.

The most commonly used approach is \textbf{\textit{majority voting}}. Initially, Limaye et al.~\cite{Limaye2010} defined a feature vector based on the occurrence and frequency of relations between entities. The most frequently occurring relation is the selected one. A similar method is applied in~\cite{Syed2010,Mulwad2010,Mulwad2011,Venetis2011,Buche2013,Mulwad2013,Munoz2013,Chabot2019,Cremaschi2019,Morikawa2019,Steenwinckel2019,Thawani2019,Kruit2019,Chen2020,Cremaschi2020-1,Kim2020,Nguyen2020,Shigapov2020,Baazouzi2021,Heist2021,Cremaschi2022}. In~\cite{Steenwinckel2019} an additional criterion is introduced to handle cases where relations have equal occurrence. The approach uses column types to examine the \ac{rdfs} range and domain in such situations. When multiple relations have a valid range and domain, the approach selects the relations with the most specific range and domain column types.
Another important aspect is handling duplicate cells within the same column, as the frequency count can be misleading in cases with a high number of duplicates. In~\cite{Ermilov2016}, the authors address this issue by considering possible duplicate cells only once. In contrast, the T2K approach~\cite{Ritze2015}, do not perform deduplication. Instead~\cite{Knoblock2012}, takes a different approach by extracting relations for each pair and uses the Stiner Tree Algorithm to compute the minimal tree among them.

Recent approaches have introduced diverse methods for handling NE and LIT relations. Moreover, the adoption of \textbf{\textit{statistical}} or \textbf{approximate matching} methodologies have seen a noticeable increase. In~\cite{Taheriyan2014,Taheriyan2016-1,Taheriyan2016-2}, the authors leverage existing user-defined \ac{kgs}. The relations are annotated using a directed weighted graph constructed on top of known properties, which are expanded using semantic types in the domain ontology. The properties are represented as weighted links between nodes.
TableMiner+~\cite{Zhang2017} aims to enhance relation matching by employing the Dice similarity measure~\cite{Dice1945}. The Dice function calculates an overlap score by comparing the bag-of-words representations of the cell and the object of a triple, and subsequently, the most frequent resulting relation is selected.

When handling NE columns, several approaches employ string similarity functions such as Levenshtein distance (also known as Edit Distance), Jaccard Similarity, letter distance, or bag-of-words overlap. Notable examples of approaches that use these techniques include~\cite{Nguyen2019,Chen2020,Abdelmageed2020,Cremaschi2020-2,Abdelmageed2021,Avogadro2021,Zhang2020,Abdelmageed2021-2,Abdelmageed2022}.

For LIT columns, various techniques are employed in different approaches. TeamTR~\cite{Yumusak2020} and JenTab~\cite{Abdelmageed2020,Abdelmageed2021,Abdelmageed2021-2,Abdelmageed2022} use fixed thresholds for comparing literals in the \ac{kg} with the values in the table. Similarly, other approaches employ custom formulas, such as DAGOBAH~\cite{Huynh2020}, which uses an absolute value formula, while Mantistable SE~\cite{Cremaschi2020-2} and MantisTable V~\cite{Avogadro2021} use an exponential function as threshold for literal comparison.
The MTab approach~\cite{Nguyen2019} approximates comparisons using a threshold and applies a custom formula for numeric values. Similarly, LinkingPark~\cite{Chen2020} uses pre-computed statistics specifically designed for numerical columns. A more detailed description of their process is provided in~\cite{Chen2022}.
When dealing with dates, after parsing, they may be treated similarly to numerical values (\eg~in~\cite{Cremaschi2020-2,Avogadro2021,Abdelmageed2020,Abdelmageed2021}).
In DAGOBAH~\cite{Huynh2020}, the considered types include ID, number, string, and date types. The authors employ different matching techniques for matching properties for each data type. Date matching involves considering various formats. As described in the previous Section, the MAGIC approach~\cite{Steenwinckel2021} employs a comprehensive procedure that simultaneously addresses CEA, CTA, and CPA. Majority voting is used as the selection strategy across columns, similar to the aforementioned approaches.

Some \textbf{embedding-based methods} have recently emerged for the \ac{cpa} task. These methods leverage property features to represent the potential relationships between the target and surrounding columns. In the approach proposed by Chen et al.~\cite{Chen2019learning}, a property Vector algorithm (P2Vec) is introduced for Predicate Annotation. In the 2021 version of DAGOBAH~\cite{Huynh2021}, \ac{kg} embedding is incorporated along with existing features to enhance property disambiguation, while majority voting remains the selection criteria. Another embedding-based method, TCN, is presented in Wang et al.~\cite{Wang2021}. TCN concatenates the embedding of the subject and object columns and passes through a dense layer to generate predictions on their relationship. Lastly, in Doduo~\cite{Suhara2022}, the corresponding embedding representation, as described in the Type Annotation (Section~\ref{subsec:type-annotation}), is extracted for each column. These embedding representations are also used for the \ac{cpa} task.

Some approaches need more details to fully understand their processes. For example, ADOG~\cite{Oliveira2019} uses CEA results to extract properties for NE-columns, but the paper does not provide enough information about the underlying algorithm. Similarly, MTab2021~\cite{Nguyen2021} and DAGOBAH2022~\cite{Huynh2022} also lack adequate methodology description. In the geospatial domain, Cruz et al.~\cite{Cruz2013} mention that geospatial classification schemes can be modelled using a part-of or is-a relationship. However, the paper does not delve into the methodology in detail.

Several other approaches, such as~\cite{Goel2012,Pimplikar2012,Deng2013,Ermilov2013,Quercini2013,Zhang2013,Zwicklbauer2013,Bhagavatula2015,Ramnandan2015,Neumaier2016,Pham2016,Efthymiou2017,Ell2017,Kacprzak2018,Luo2018,Zhang2018,Chen2019,Hulsebos2019,Takeoka2019,Zhang2019,Azzi2020,Baazouzi2020,Guo2020,Eslahi2020,Khurana2020,Li2020,Tyagi2020,Yang2021,Zhou2021,Gottschalk2022,Liu2022} do not address Predicate Annotation.

\subsection{NIL Annotation}
\label{subsec:nil-annotation}

In the context of \ac{nlp}, the \ac{nil}~\cite{Ilievski2020} Annotation refers to the task of finding and linking whether a given input belongs to a particular category, class or type called ``NIL'' or ``None''.
In \ac{sti}, \ac{nil} Annotations indicate cells in tables lacking relevant information. \ac{sti} involves extracting structured data from tables, but some cells may not correspond to entities in a \ac{kg}, leading to \acf{nil} predictions. \ac{nil} Annotations can be helpful in various applications, such as information extraction, question answering, or \ac{kg} population, where the goal is to extract structured information from tables and integrate it into a knowledge representation system. There are four common ways in the \ac{sota} to perform \ac{nil} annotation~\cite{Mehwish2022}: i)~\textbf{\textit{no candidates}}~\cite{Bhagavatula2015,Wang2012,Ritze2015}, ii)~\textbf{\textit{threshold}}~\cite{Mulwad2010}, iii)~\textbf{\textit{separate model}}~\cite{Kruit2019,Zhang2020,Deng2022,Heist2021}, and iv)~\textbf{\textit{\ac{nil} predictor}}. Some approaches do not belong to any of the above categorisations. For the scope of this survey, it is possible to classify them considering the peculiarity of the \ac{nil} annotation process, which use v)~\textbf{\textit{external services}} for searching candidates~\cite{Quercini2013,Sekhavat2014}.

Sometimes in the \ac{el}, a candidate generator does not yield any corresponding entities for a mention (\textbf{\textit{no candidates}}); such mentions are trivially considered unlinkable. Bhagavatula et al.~\cite{Bhagavatula2015} introduce a variation of the \ac{el} task for tables, using a graphical model representation, where each mention in the table is associated with a discrete random variable representing its candidate entities for identifying and disambiguating unlinked mentions in a Wikipedia table. After performing the \ac{cta} task, Wang et al.~\cite{Wang2012} proceed to ``expand entities'' by creating entities that lack corresponding labels within the \ac{kg}. Also~\cite{Ritze2015} explicitly mentions that their T2K approach could fill the missing values to DBpedia and vice-versa (Web Tables), but it is only listed as a potential use and not validated.

A second group of approaches set a \textbf{\textit{threshold}} for the best linking probability (or a score), below which a mention is considered unlinkable. For instance, T2LD approach~\cite{Mulwad2010} links table cells to entities using the results obtained from a \ac{cta} task, and then a query is sent to the \ac{kg} for each cell, which returns the top $N$ possible entities which are then ranked using an SVM classifier. If the evidence is not strong enough, it suggests that the table cell represents a new entity.

To discover \ac{nil} mentions, it is also possible to train an additional binary classifier (\textbf{\textit{separate model}}) that accepts, as input, mention-entity pairs after the ranking phase and several additional features. It makes the final decision about whether a mention is linkable or not. In this group, Kruit et al.~\cite{Kruit2019} employ \ac{kg} entity and relation embeddings to enhance the disambiguation process when label matching is insufficient. This approach contributes to discovering new facts for \ac{kg} completion. The system developed by~\cite{Zhang2020} proposes a method for discovering new entities containing a subset of those entities that can be extended with information features: a neural embedding space (Word2vec representation), a topical space (annotation of other entities) and a lexical space (normalised Levenshtein distance). Deng et al.~\cite{Deng2022} propose an innovative approach that uses a Transformer encoder with masked self-attention to predict the masked entities based on other entities and the table context (\eg~caption/header). This encourages the model to learn factual knowledge from tables and encode it into entity embeddings for annotation use. Eventually, Heist et al.'s algorithm~\cite{Heist2021} considers a data corpus from which co-occurring entities and related relationships can be extracted (\eg~listings in Wikipedia or a collection of spreadsheets). Furthermore, they assume that a \ac{kg} which contains a subset of those entities can be extended with information learned about the co-occurring in the corpus.

Regarding the \ac{nil} Annotation in the \ac{sota}, some models developed for annotation of the free-form text introduce an additional special ``NIL'' entity in the ranking step in the \ac{el} phase, so models can predict it as the best match for the mention~\cite{Mehwish2022}. It should be noted that currently, \ac{sti} approaches do not employ this technique.

Related to the use of \textbf{\textit{external services}},~\cite{Quercini2013} describes an algorithm that uses web search engines to gather information about ``unknown entities'' (not present in the \ac{kg}) and annotate them with the correct type analysing the snippet. Similarly, the approach in~\cite{Sekhavat2014} searches for exact entity matches across the Subject-Verb-Object (SVO)\footnote{\href{http://rtw.ml.cmu.edu/resources/svo/}{rtw.ml.cmu.edu/resources/svo}} Triples of the Never-Ending Language Learning (NELL) project. The process creates a probabilistic model to estimate the posterior probability of a relationship along with entity-pair instances, and then it uses this relation to create new entities.

\section{Method (Supervision)}
\label{sec:method}

This Section delves into the METHOD dimension and its distinctive categories. While we review the approaches proposed to solve individual tasks in Section~\ref{sec:sub-tasks}, here we summarise the overall usage of labelled data to train models that solve one or more specific tasks. We consider three categories of approaches. First, the UNSUPERVISED approaches (Section~\ref{subsec:unsupervised}) do not rely on annotated data during the \ac{sti} process. Secondly, the SUPERVISED approaches (Section~\ref{subsec:supervised}) utilise a training set, such as a collection of pre-annotated tables. Some approaches are characterised by using both unsupervised and supervised techniques; we can define these approaches as HYBRID (Section~\ref{subsec:hybrid}).

\begin{figure}[H]
  \centering
  \includegraphics[width=.5\textwidth, height=.2\textwidth]{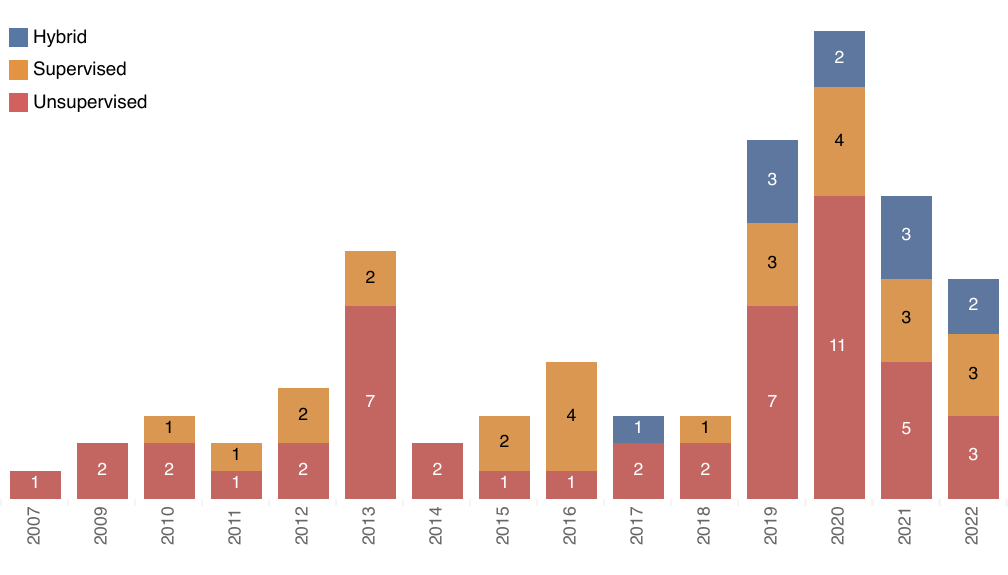}
  \caption{A comparison of Supervised, Unsupervised and Hybrid approaches per year.}
  \label{fig:supunsup}
\end{figure}

The chart in Fig.~\ref{fig:supunsup} displays the yearly distribution of approaches in supervised and unsupervised categories. In the years leading up to 2019, the number of approaches in both categories is roughly the same. However, in 2019, the SemTab challenge's inaugural edition led to a significant uptick in unsupervised methods, peaking at 13 approaches proposed in 2020. As of 2021, the gap between supervised and unsupervised approaches is closing due to the increasing use of \ac{gss} as training and the implementation of \ac{llm} (\eg~BERT) for annotations.

\vspace{-0.7cm}

\paragraph{Unsupervised}
\label{subsec:unsupervised}

Numerous approaches~\cite{Hignette2007,Hignette2009,Tao2009,Limaye2010,Syed2010,Venetis2011,Pimplikar2012,Wang2012,Buche2013,Deng2013,Ermilov2013,Munoz2013,Quercini2013,Zhang2013,Zwicklbauer2013,Sekhavat2014,Taheriyan2014,Ritze2015,Ermilov2016,Ell2017,Zhang2017,Kacprzak2018,Zhang2018,Chen2019,Cremaschi2019,Morikawa2019,Nguyen2019,Oliveira2019,Steenwinckel2019,Thawani2019,Abdelmageed2020,Azzi2020,Baazouzi2020,Chen2020,Cremaschi2020-1,Cremaschi2020-2,Kim2020,Nguyen2020,Shigapov2020,Tyagi2020,Yumusak2020,Abdelmageed2021,Abdelmageed2021-2,Avogadro2021,Baazouzi2021,Nguyen2021,Abdelmageed2022,Chen2022,Cremaschi2022} (\unsupapproaches) prioritise the utilisation of unsupervised methods, driven by the challenges associated with acquiring high-quality datasets for real-world scenarios and the difficulty of modeling annotation problems for \ac{ml} methods. Customised approaches offer greater control over the results and higher annotation precision (as discussed in Appendix~\ref{sec:gold-standards}). Table~\ref{table:unsupervised-approaches} in the Appendix~\ref{appendix-sec:supervised-unsupervised} displays the techniques used in the Candidate Generation and Entity Disambiguation of \ac{el} step.
\vspace{-0.2cm}
\paragraph{Supervised}
\label{subsec:supervised}

While a relatively limited number of supervised approaches has been proposed (if compared to the number of supervised approaches), these approaches (\supapproaches) offer exciting solutions when implemented. Table~\ref{table:supervised-approaches} in the Appendix~\ref{appendix-sec:supervised-unsupervised} displays the techniques used by these approaches. Supervised approaches listed below use a collection of tabular or textual data to learn patterns and relationships between the input data and the corresponding labels. In the following, we briefly summarise the data used to train supervised approaches and the main idea behind these approaches.

Mulwad et al.~\cite{Mulwad2010} query Wikitology\footnote{\href{https://ebiquity.umbc.edu/project/html/id/83/Wikitology}{ebiquity.umbc.edu/project/html/id/83/Wikitology}} to extract entities and use a \ac{svm} model for entity ranking.
Quercini et al.~\cite{Quercini2013} collect label categories from DBpedia and snippets from Bing to train the text classifier.
Ermilov et al.~\cite{Ermilov2016} use a portion of the T2D Gold Standard\footnote{\href{http://webdatacommons.org/webtables/goldstandard.html}{webdatacommons.org/webtables/goldstandard.html}}, where the S-column and column-pairs have been annotated to train an SVM and a \ac{pgm}. T2Dv2 dataset\footnote{\href{http://webdatacommons.org/webtables/goldstandardV2.html}{webdatacommons.org/webtables/goldstandardV2.html}} has been used by Zhang et al.~\cite{Zhang2020} to train a binary classifier and Chen et al.~\cite{Chen2019} to train a \ac{cnn}. The former uses also the WDC Tables dataset\footnote{\href{https://webdatacommons.org/webtables}{webdatacommons.org/webtables}} to train a random forest model. TabEl~\cite{Bhagavatula2015} parses Wikipedia tables to build a corpus containing more than $1.6$ million tables\footnote{\href{http://websail-fe.cs.northwestern.edu/TabEL}{websail-fe.cs.northwestern.edu/TabEL}}. Such tables contain hyperlinks to Wikipedia, and TabEL uses these hyperlinks as probability estimates for training a Markov Network. The same training technique is used also by Takeoka et al.~\cite{Takeoka2019} on a private dataset of $183$ human-annotated tables with $781$ NE-columns and $4109$ LIT-columns\footnote{The paper referenced UCI ML repository, but it was found to be a private dataset upon contacting the author.}. Pham et al.~\cite{Pham2016} use four different datasets: city, weather, museum and soccer\footnote{\href{https://github.com/usc-isi-i2/eswc-2015-semantic-typing}{github.com/usc-isi-i2/eswc-2015-semantic-typing}, the soccer dataset is no longer available.}. Only city, museum and soccer datasets have been used to train random forest and logistic regression models. The weather dataset is only used in semantic labelling because it cannot provide sufficient feature vectors for training classifiers. Neumaier et al.~\cite{Neumaier2016}, propose a hierarchical clustering to build a \ac{bkg} from DBpedia.
Luo et al.~\cite{Luo2018} use English and Chinese dumps of Wikipedia for training word and entity embeddings. In total, it collects $3818$ mentions from $150$ tables. Deng et al.~\cite{Deng2022} use Wikitable corpus to train a BERT model while Gottschalk et al.~\cite{Gottschalk2022} creates a new synthetic dataset automatically extracted from GitHub repositories\footnote{\href{https://github.com/search/advanced}{github.com/search/advanced}}. This dataset is then used for training the Siamese network.
\vspace{-0.2cm}
\paragraph{Hybrid}
\label{subsec:hybrid}

A small number of approaches (\hybridapproaches) opt for solutions involving the use of supervised and unsupervised techniques. Since 2017, an increase in the use of semantic embeddings has been observed. The embeddings exploit a vectorial representation of the rich entity context in a \ac{kg} to identify the table's most relevant subset of entities. This technique is used in conjunction with lookup-focused~\cite{Efthymiou2017,Eslahi2020} or rule-based~\cite{Chabot2019,Huynh2020,Huynh2021} methods. Others adopt a transformer-based embedding in combination with the use of heuristics~\cite{Huynh2022,Liu2022} or embedding technique~\cite{Steenwinckel2021}. ColNet~\cite{Chen2019} utilises a CCN model and a method which automatically extracts samples from the \ac{kg}. Kruit et al.~\cite{Kruit2019} employ a \acl{pgm} and features on T2D and Webaroo\footnote{The dataset is no longer available.} while~\cite{Heist2021} rely on distant supervision to derive rules for CPA and rule-mining techniques.

\section{Domain}

\ac{sti} approaches can be categorised as domain-dependent or domain-independent. Domain-dependent systems address problems and provide solutions specific to the domain they are built for. Instead, domain-independent \ac{sti} approaches do not rely on domain knowledge and provide solutions not tied to a specific area of expertise.
Among all the approaches only~\cite{Hignette2007,Hignette2009,Cruz2013,Luo2018,Buche2013,Neumaier2016,Kacprzak2018} can be classified as domain-dependent. A few papers~\cite{Hignette2007,Hignette2009,Buche2013} are related to the food microbiology domain and present a predefined set of rules specifically built to address the different \textit{numerical} units in Datatype Annotation. Two papers~\cite{Neumaier2016,Kacprzak2018}, are designed to deal only with numerical data labelling. Finally,~\cite{Cruz2013} is designed to address several challenges that come from the geospatial and temporal data manipulation and~\cite{Luo2018} address data and \ac{kg} written in different languages.

The remaining approaches can be considered domain-independent.

Domain-dependent approaches, usually, are affected by the domain ontology used, such as Hignette et al.~\cite{Hignette2007,Hignette2009} that distinguish between symbolic and numeric columns, using some of the knowledge described in the ontology, which has been created ad hoc.

\section{Application/Purpose}
\label{sec:application_purtpose}

This Section discusses the application purpose of approaches ranging from i)~\textit{\textbf{\ac{kg} construction}}, ii)~\textit{\textbf{\ac{kg} extension and tabular data enrichment}}, and iii)~\textit{\textbf{cross-lingual linking}}. For the first two purposes especially, we refer to Fig.~\ref{fig:motivations} and Fig.~\ref{fig:03_def_sti}.

The aim of \textit{\textbf{\ac{kg} construction}} is to derive meaningful information from tabular data, transforming it into a structured and interconnected knowledge representation. This supports cross-domain knowledge integration, discovery, and graph-based data analysis. Approaches~\cite{Heist2021, Ell2017, Ritze2015, Zhang2017, Zhang2013, Kruit2019, Ermilov2016, Kim2020, Deng2022, Taheriyan2016-2, Taheriyan2014} employ \ac{sti} to construct and populate \ac{kgs}. Similarly,~\cite{Munoz2013, Mulwad2013} focus on extracting facts as \ac{rdf} triples from annotated tables.

\ac{sti} significantly enhances \textit{\textbf{\ac{kg} extension and tabular data enrichment}} by extracting structured information from tables, linking it to an existing \ac{kg}, and expanding the \ac{kg} with interpreted graph data. This process, involving automatic identification and annotation of mentions, relationships, and table schema, ensures the extraction of comprehensive and accurate information from tabular data. Approaches such as~\cite{Zhang2013, Heist2021, Zhang2020, Gottschalk2022} discover new entities, types, and properties to enrich the \ac{kg}, including identifying new labels for known entities. Additionally, \cite{Cutrona2019,cutrona2019semantically,Ciavotta2022}, showcase how \ac{sti} supports adding more columns to input tables for downstream data analytics, using annotations as joins to fetch relevant data from the external \ac{kg} and export the output table.
Semantic table interpretation provides a means to bridge the language barrier, thus supporting \textit{\textbf{cross-lingual linking}} of entities mentioned in the tables. A similar, cross-lingual application is proposed in~\cite{Luo2018}, which annotates a table containing entities expressed in one language with entities in a \ac{kg} expressed in another.

Since 2019, the majority of recent approaches have focused more on addressing the tasks outlined in the SemTab challenge~\cite{Chabot2019,Cremaschi2019,Nguyen2019,Morikawa2019,Thawani2019,Oliveira2019,Steenwinckel2019,Abdelmageed2020,Azzi2020,Tyagi2020,Shigapov2020,Nguyen2020,Huynh2020,Baazouzi2020,Chen2020,Cremaschi2020-2,Kim2020,Yumusak2020,Avogadro2021,Abdelmageed2021,Baazouzi2021,Huynh2021,Nguyen2021,Yang2021,Steenwinckel2021,Chen2022,Cremaschi2022,Huynh2022,avogadro2024feature},
and less on downstream applications.
We observe that in some applications of \ac{kg} construction and tabular data enrichment, the most important task to fully automate is \ac{cea} (especially on large tables), while it is assumed that \ac{cpa} and \ac{cta} can be manually performed or revised by a user to ensure a desired level of quality.


\section{License}
\label{sec:licence}

Licensing is vital in protecting intellectual property and establishing the terms under which software, content, or creative works can be used or distributed. Thus, it is very important to review \ac{sti} approaches under such dimension so that users can evaluate which one to use for their specific use case and ensure compliance with legal requirements. There are \numlicence different licenses used by \ac{sti} tools and approaches. 

The most used one is Apache 2.0\footnote{\href{https://www.apache.org/licenses/LICENSE-2.0}{apache.org/licenses/LICENSE-2.0}}, an open-source software license widely used in the development and distribution of software. It allows users to freely use, modify, and distribute the licensed software. Users must include a copy of the Apache 2.0 license, a clear attribution to the original authors, and clearly identifiable modification notices on all altered files. Such licensing is used by \licenceApache approaches~\cite{Knoblock2012,Ramnandan2015,Ritze2015,Neumaier2016,Pham2016,Taheriyan2016-1,Taheriyan2016-2,Ell2017,Zhang2017,Chen2019,Chen2019learning,Cremaschi2019,Zhang2019,Cremaschi2020-1,Cremaschi2020-2,Li2020,Abdelmageed2021,Abdelmageed2021-2,Avogadro2021,Abdelmageed2022,Deng2022,Cremaschi2022,Suhara2022}. The second most used license is MIT\footnote{\href{https://opensource.org/license/mit/}{opensource.org/license/mit}} used by \licenceMIT approaches~\cite{Kacprzak2018,Hulsebos2019,Kruit2019,Thawani2019,Abdelmageed2020,Shigapov2020,Nguyen2021,Chen2022,Gottschalk2022}. The MIT License is a permissive open-source software license that allows users to freely use, modify, and distribute the licensed software. The Orange license\footnote{\href{https://orangedatamining.com/license/}{orangedatamining.com/license}} is used by~\cite{Chabot2019,Huynh2021,Huynh2022,Liu2022}. GPL 3.0\footnote{\href{https://www.gnu.org/licenses/gpl-3.0.html}{gnu.org/licenses/gpl-3.0.html}} is used by \licenceGPL approaches~\cite{Ermilov2016,Heist2021} while \licenceCCA adopts CCA 4.0\footnote{\href{https://creativecommons.org/licenses/by/4.0/}{creativecommons.org/licenses/by/4.0}}~\cite{Zhang2020,Bhagavatula2015}, and eventually,~\cite{Steenwinckel2021} employs a unique licensing by Ghent University (Imec).

Among the approaches reviewed in this survey, \licenceNotSpecified of them lack any specific licensing information.
\vspace{-0.2cm}
\section{Validation}
\label{sec:validation}

The effectiveness of the approaches proposed so far is usually evaluated in terms of annotation quality on different separate computational tasks (\eg~using Precision, Recall, and F1 scores). In the literature, the open-source tool \textbf{STILTool}~\cite{cremaschi2020stiltool} is available for the automated evaluation of the quality of semantic annotations generated by semantic table interpretation methods. For each approach, we specify the datasets used for its evaluation. We report this association in Table~\ref{table:stats-gss}, discussed in Appendix~\ref{sec:gold-standards}, where we discuss the datasets used for table interpretation. A very few approaches also considered other dimensions for evaluation, such as execution times and scalability~\cite{Ciavotta2022,Deng2022}; the latter dimension is relevant, especially for entity linking, which may be inefficient on large tables.

\section{Open Issues and Potential Research Directions}
\label{sec:remarks}

Despite the many contributions to advance \ac{sti} discussed in this paper, we believe that there are some open issues associated with the key challenges listed in Section~\ref{sec:introduction}; these open issues could hinder a broader uptake of \ac{sti} solutions for downstream applications and, at the same time, suggest valuable questions for researchers working in these fields.

\textbf{\textit{i) Heterogeneity of domains and data distributions:}} the lack of labelled data specifically tailored for a domain of interest prevents training and evaluation of domain-specific solutions. Potential solutions to overcome this scarcity could be: a)~involving domain experts in the annotation process, additionally using crowd-sourcing platforms to engage a larger pool of annotators with domain-specific knowledge~\cite{Ritze2017}; b)~using specific data (\eg~\cite{Singh2022,Abdelmageed2021biodivtab,Jiomekong2022}); c)~developing pre-trained models on large \ac{gss}, which can be fine-tuned and/or adapted using a small amount of labelled data (in this way, \ac{sti} approaches would reuse existing annotations and reduce the burden of creating domain-specific \ac{gss} from scratch) (\eg~\cite{Deng2022}).

\textbf{\textit{ii) Limited contextual information:}} missing context can introduce ambiguity, making it challenging to determine the intended meaning of table elements. Most of the \ac{gss} used in recent work do not provide tables associated with extended context; consequently, these aspects have not been emphasised much in recent work. Possible solutions could be: a)~reusing prior datasets based on web tables, which emphasise these challenges, or b)~developing new datasets. In addition, researchers should consider c)~techniques that infer missing context from the available information (\eg~table headers, table metadata, surrounding text~\cite{Deng2022}) or d) employing final-user feedback to enhance the contextual understanding of the table~\cite{Huynh2022}.

\textbf{\textit{iii) Detecting the type of columns:}} the analysis of the \ac{sota} has shown that different approaches effectively manage type annotation. The most critical open challenge is the detection of L-columns. Using \ac{regex} was found effective for identifying L-columns~\cite{Cremaschi2022}, but domain-specific literal values (\eg~for genomics data of biological pattern) are not yet addressed. A potential solution is the definition of new domain-specific regular expressions for the Type Annotation sub-task.

\textbf{\textit{iv) Matching tabular values against the \ac{kg}:}} \ac{sti} approaches work well when the mentions in the NE-columns or literals in the L-columns are similar enough to the values in the \ac{kg}. Regarding annotation of mentions, synonyms, aliases, abbreviations, and acronyms, should be considered to enhance the approach's potential. This remains an open issue, as only a few approaches use indexes with aliases (\eg~\cite{Cremaschi2022,avogadro2024feature}). While this direction seems promising, there is still ample room for improvement. It is also necessary to consider that the mention can contain typos or have syntactic differences from the entities in an \ac{kg}. In such cases, using a)~\textit{indices}~\cite{Cremaschi2022, Nguyen2021,avogadro2024feature} and adequate b)~\textit{similarity measures}~\cite{Zhang2020, Tyagi2020} can increase the results of the candidate generation. The SemTab challenge, which tests \ac{sti} even on corrupted data, has shown that this problem has not been adequately solved yet. Regarding the use of literal values for matching (\ie those in L-columns), the challenge arises due to inconsistencies between the values in the \ac{kg} and those in tabular data. Since \ac{kgs} are known to be incomplete or not updated frequently enough, the correct literal value for a given property may significantly deviate from the one in the table (or in the \ac{kg}). In the \ac{sota}, this challenge is typically addressed by setting c)~thresholds or ranges (\eg~\cite{Neumaier2016}), but these methods introduce the risk of selecting incorrect annotations. A promising research direction would involve leveraging d) \textit{statistical and \ac{ml} methods} to surmount this limitation and achieve even better results.

\textbf{\textit{v) Disambiguation of named entities:}} disambiguation remains a complex and challenging task when the table context is insufficient or unclear. Another aspect that makes disambiguation still challenging is the presence of homonyms in the \ac{kg}, especially when they belong to very similar types. In this case, only the surrounding context can help disambiguate different candidates. However, this remains an open issue as not all approaches consider contextual analysis, or as depicted in challenge ii), sometimes the context should be inferred. In the context of this challenge, homonym management plays a crucial role.
A possible research direction a)~is to include models that consider all elements contributing to the creation of the context (\eg~\cite{Deng2022}). 
Some of the most recent approaches have proposed to use b)~\ac{llm} to capture complex linguistic patterns, semantic relationships, and contextual cues in the tabular data, obtaining some promising results in improving disambiguation (\eg~\cite{Suhara2022}). However, further investigation is needed to explore the use of such models, their efficiency, and comparison with traditional approaches. Some recent approaches have also proposed using latent representations or feature-based neural networks to re-rank candidate entities retrieved with more traditional techniques~\cite{Avogadro2023,Liu2022}: these hybrid solutions are also promising.   

\textbf{\textit{vi) \ac{nil}-mentions:}} Currently, only 10 approaches address \ac{nil} annotation, yielding inconsistent results, as seen in the 2022 and 2023 SemTab challenges. The \ac{sota} has not adequately tackled this crucial aspect despite its significance in \ac{kg} extension and construction for practical applications. Limited \ac{nil} coverage in \ac{gss} biases algorithms toward always selecting the best candidate without deciding whether to link. One solution is to a) develop \ac{gss} that better represent the problem, encouraging solutions that decide on linking the best candidate entity (\eg\cite{Marzocchi2022}). Additionally, b)  utilise techniques and external sources (\eg search engines) for enriched representations of mentions and entities~\cite{Nguyen2021}. Lastly, c)~incorporate domain-specific expert knowledge to enhance \ac{nil}-mention identification.

\textbf{\textit{vii) Choosing the most appropriate types and properties:}} 
more than one type could capture the meaning of one column. This is due especially to the hierarchical organisation of types in ontologies (\eg~an actor is also a person) but also to the presence of very similar types (\eg~in Wikidata). Selecting the types that \textit{better} captures the semantics of a column among all correct types is still an open issue. Possible solutions may come from a)~using contextual information.
Analogous issues affect the selection of properties to annotate pairs of columns, which is even more challenging: predicate annotation is usually performed after other sub-tasks, which increases the risk of error propagation. 

\textbf{\textit{viii) Collective aggregation of evidence from different tasks:}} as described in Section~\ref{sec:introduction}, table interpretation is a collective decision-making process. Finding strategies to maximise evidence exchange across sub-tasks effectively is challenging. a)~Heuristic and b)~\ac{ml} approaches can be useful to overcome this challenge. The heuristic approach depends on expert-derived strategies and fixed rules, making it less adaptable to dynamic environments and potentially hindering its performance with different data inputs~\cite{Cremaschi2022, Chen2022, Abdelmageed2022, Nguyen2021,avogadro2024feature}. \ac{ml} models, instead, offer the capability to make decisions based on learned patterns and relationships between different features, such as table data, annotations, and contexts~\cite{Suhara2022, Deng2022}. In addition, \ac{ml} techniques allow determining feature weights by iterating on data and adapting them incrementally.

\textbf{\textit{ix) Amount and shape of data:}} the amount of data introduces two opposite challenges: data abundance and data scarcity. Data abundance can pose significant challenges for \ac{sti} approaches. a)~Sampling, subset selection, feature selection, or dimensionality reduction can be employed to address data abundance. Data scarcity can be addressed by b)~data augmentation (\eg~\cite{Ritze2015,Machado2022}). c)~Techniques such as synthetic data generation, sampling methods, or data transformations can be used to create additional training instances. Moreover, d) transfer or active learning can help overcome data scarcity (\eg~\cite{Chen2019}). However, this remains an open challenge as none of the reviewed approaches adopts any of these techniques for data augmentation.

\textbf{\textit{x) Annotation of complex formatted tables:}} annotating complex tables introduces unique challenges due to their intricate structures, which often include merged cells, hierarchical data, and varying formats. These complexities can obscure relationships between data points, making it difficult to apply standard annotation methods effectively. Web tables containing complex structures constitute a small population and have not been the focus of research~\cite{Zhang2017}. One solution could be to incorporate a pre-process step that parse complex structures~\cite{zanibbi2004survey}.

In addition to these open issues closely related to the \ac{sti} process, it is possible to identify other open issues related to the \ac{gss}.

\textbf{\textit{i) \ac{gss} Availability:}} Reliable benchmarks are essential for evaluating the effectiveness of \ac{sti} methods. Our review revealed a lack of high-quality benchmarks (see Appendix~\ref{sec:gold-standards}), impeding the development and assessment of \ac{sti} techniques. To enhance robustness evaluations across diverse data distributions, it is crucial to assess \ac{sti} approaches using various \ac{gss}. Additionally, the creation of domain-specific \ac{gss} tailored to specific applications is recommended. 

\textbf{\textit{ii) Multi-lingual \ac{gss}:}} The current limitation of predominantly English \ac{gs} hampers the reproducibility and generalization of \ac{sti} approaches across languages in real-world scenarios. Integrating language detection is crucial to address this issue and enhance \ac{sti} systems. Additionally, creating multi-lingual \ac{gss} is essential to support the training and evaluation of these systems, covering diverse languages, data sources, and domains for comprehensive coverage.

\textbf{\textit{iii) \ac{gss} with \ac{nil}:}} as discussed above, \ac{nil}-mentions are absent or underrepresented in \ac{gss} used in \ac{sota}. We believe that creating \ac{gss} that better cover this annotation type is very important.

\textbf{\textit{iv) Evaluation metrics:}} different approaches use different metrics to evaluate their performance. For instance, in the SemTab challenge, different formulas are used to calculate \textit{Precision}, \textit{Recall} and \textit{F1 measures} in relation to different datasets. For this reason, there is a need for standardised and concrete metrics to effectively test and evaluate various approaches. 

Regarding tools for implementing \ac{sti} approaches, the following open issues can be identified:

\textbf{\textit{i) Transferability:}} while reviewing specific approaches, we observed limitations in their usability in real-world scenarios. In fact, evidence of usage of \ac{sti} approaches in real-world downstream applications is still limited.

\textbf{\textit{ii) Replicability:}} while this survey includes numerous \ac{sti} approaches, a significant portion of them lack publicly available replication code. 
The lack of availability of open-source systems has two main implications: testing and evaluating third-party approaches become a complex, time-consuming, and error-prone task; checking errors and understanding issues and limitations to advance the field is difficult. Better sharing of source code can improve transparency and accelerate advancements in this field. 

\textbf{\textit{iii) Usability:}} Just a handful of tools feature a UI, and among them, only a minority possess a well-crafted UX. To ensure the usability of these solutions, it is imperative to conduct user tests, monitor user behaviour, and employ other techniques tailored to the UI and UX design process.

\textbf{\textit{iv) Adaptability:}} Most of the approaches and tools come with static algorithms. However, when users want to annotate their data, they would like to optimise algorithms for specific data distributions. Improving the support for human-in-the-loop annotation with algorithms that exploit the feedback collected from the users through the UI would provide solutions more helpful in several downstream applications.     






\vspace{-0.2cm}
\section{Conclusions}
\label{sec:conclusion-future-research}

This survey aims to provide a comprehensive and in-depth analysis of available approaches that perform \ac{sti}. It includes approaches from 2007 to the time of writing, resulting in the identification of \totapproaches approaches. Different criteria are used to compare and review \ac{sti} approaches, which are organised into a taxonomy to allow a fair comparison and identify potential future research areas. 
This analysis allowed us to create the Table~\ref{tab:selection_table} in Appendix~\ref{sec:additional-material}, which provides support in selecting approaches in relation to various attributes, such as Method, Tasks, Code availability, License and Triple store.
Also, tools and \ac{gs} have undergone a thorough analysis using specific comparative criteria. As a result of such analysis, open issues have been addressed, and potential research directions have been described. The survey aims to serve as a valuable resource for newcomers, providing an overview of the current \ac{sota} in \ac{sti} and facilitating their exploration of potential directions for enhancing \ac{sti} performance. In future work, open issues for each approach will be identified. Another direction is to review the performance metrics used by each approach.

\bibliographystyle{abbrv}
\bibliography{biblio}

\begin{acronym}
\acro{ai}[AI ]{Artificial Intelligence}
\acro{bow}[BOW]{Bag-Of-Words}
\acro{bkg}[BKG]{Background Knowledge Graph}
\acro{cta}[CTA]{Column-Type Annotation}
\acro{cea}[CEA]{Cell-Entity Annotation}
\acro{cpa}[CPA]{Columns-Property Annotation}
\acro{crf}[CRF]{Conditional Random Field}
\acro{dl}[DL]{Deep Learning}
\acro{el}[EL]{Entity Linking}
\acro{gui}[GUI]{Graphical User Interface}
\acro{gdbt}[GDBT]{Gradient Boosted Decision Tree classification model}
\acro{gs}[GS]{Gold Standard}
\acro{gss}[GSs]{Gold Standards}
\acro{kg}[KG]{Knowledge Graph}
\acro{kgs}[KGs]{Knowledge Graphs}
\acro{kb}[KB]{Knowledge Base}
\acro{kbs}[KBs]{Knowledge Bases}
\acro{lod}[LOD]{Linked Open Data}
\acro{cnea}[CNEA]{Cell-New Entity Annotation}
\acro{nlp}[NLP]{Natural Language Processing}
\acro{regex}[Regex]{Regular Expressions}
\acro{rdf}[RDF]{Resource Description Framework}
\acro{rml}[RML]{RDF Mapping Language}
\acro{rdfs}[RDFS]{Resource Description Framework Schema}
\acro{sota}[SOTA]{state-of-the-art}
\acro{sti}[STI]{Semantic Table Interpretation}
\acro{svm}[SVM]{Support Vector Machine}
\acro{ui}[UI]{User Interface}
\acro{ir}[IR]{Information Retrieval}
\acro{ml}[ML]{Machine Learning}
\acro{ne}[NE]{Named Entity}
\acro{ner}[NER]{Named Entity Recognition}
\acro{nel}[NEL]{Named Entity Linking}
\acro{nil}[NIL]{Not In Lexicon}
\acro{nn}[NN]{Neural Network}
\acro{lit}[LIT]{Literal}
\acro{crf}[CRF]{Conditional Random Field}
\acro{cnn}[CNN]{Convolutional Neural Network}
\acro{rnn}[RNN]{Recurrent Neural Network}
\acro{hnn}[HNN]{Hybrid Neural Network}
\acro{llm}[LLM]{Large Language Model}
\acro{llms}[LLMs]{Large Language Models}
\acro{lms}[LMs]{Language Models}
\acro{lm}[LM]{Language Model}
\acro{hdt}[HDT]{Hybrid Decision Tree}
\acro{pfsms}[PFSMs]{Probabilistic Finite-State Machines}
\acro{pgm}[PGM]{Probabilistic Graphical Model}
\acro{gft}[GFT]{Google Fusion Tables}
\acro{pois}[PoI]{Points of Interest}
\end{acronym}

\newpage

\appendix

\section{Scope and methodology}
\label{appendix:A}

\subsection{Methodology}
\label{appendix:methodology-long}

\paragraph{Identification}
\label{subsec:identification}

In order to enhance the efficiency of our search in publication databases, the authors collaboratively defined and established a set of keywords. The set of keywords is composed of \textit{semantic table interpretation}, \textit{table understanding}, \textit{\ac{sti}}, \textit{table interpretation}, \textit{semantic table analysis}, \textit{semantic table exploration}, \textit{semantic table understanding}, \textit{web tables}, \textit{semantic annotation of tabular data}, \textit{tabular data annotation}, \textit{table annotation}, \textit{semantic interpretation of structured data}, \textit{tabular data semantic labelling}, \textit{tabular data enrichment}, \textit{SemTab challenge,} or simply \textit{tabular data}. Finally, we came up with 16 keywords. Subsequently, these keywords underwent in-depth discussions among five researchers, who assigned scores ranging from 1 (denoting low relevance) to 5 (denoting high relevance) to each keyword. The final score for each keyword is calculated as the average of the scores provided by each researcher. Finally, the list of ranked keywords represented a starting point for an extensive search on several publication platforms. The following search platforms for scientific publications were utilised: i)~Scopus, ii)~Web of Science, iii)~DBLP, and iii)~Google Scholar.

The time period was set from 2007, when the \ac{sti} research field was first approached, until May 2023, when the paper collection process was completed. To complement the extensive search, we incorporated a snowballing technique, which involved exploring additional recent publications that referenced the key works identified within our result corpus. Tracing the citations of central works aimed to capture the field's most up-to-date and relevant literature.

\paragraph{Screening}
\label{subsec:screening}

Two experts manually annotated the papers obtained during the \textit{Identification} step. A key aspect of the screening process was identifying which semantic table interpretation phases were addressed/described in each publication. In addition, the criteria for this \textit{Screening} step was the relevance and comprehensiveness of each publication regarding the \ac{sti} tasks. These criteria were employed to assess how the publications addressed the relevant aspects of \ac{sti} and comprehensively treated the subject matter.
As an additional step, we performed an annotation process with pre-defined categories based on each publication's title, abstract and keywords. If the categorisation based on these three components was impossible, the full text had to be consulted at this stage. The categories for this final step were divided into generic tags (\eg~``semantic table interpretation'', and ``gold standard'') and specific annotation tags (\eg~``supervised'', ``domain independent'' Section~\ref{fig:taxonomy}).

\paragraph{Inclusion}
This Section describes our methods for identifying the final subset of publications to be included in this survey. The first and foremost criteria for inclusion were that publications had to be: i)~directly related to semantic table interpretation, ii)~published in English, and iii)~peer-reviewed. All the paper's authors decided on which publications to report based on their relevance to the assigned category.

\paragraph{Results of the paper collection process}

Through the keywords mentioned above (Section~\ref{subsec:identification}), about 134 papers were grouped; this set further decreased the number to 111 publications after the \textit{Screening} stage (Section~\ref{subsec:screening}), removing unrelated or duplicate publications. This manual annotation first involves assessing whether a paper is relevant~(1), not relevant~(0) or the annotator is unsure about its relevance~(2). In the latter case, a third annotator would determine whether to include the publication. This detailed screening stage led to the exclusion of 17 more papers, 2 of which were superseded by newer publications by the same authors, and \excludedapproaches were finally deemed not closely related to the \ac{sti}. Therefore, \totapproaches approaches were discussed in the following survey, each one described in one or more publications. For this survey's scope, as discussed in Section~\ref{sec:taxonomy}, we identified several criteria for comparing \ac{sti} approaches.

\begin{figure}[ht]
  \centering
  \includegraphics[width=0.7\textwidth]{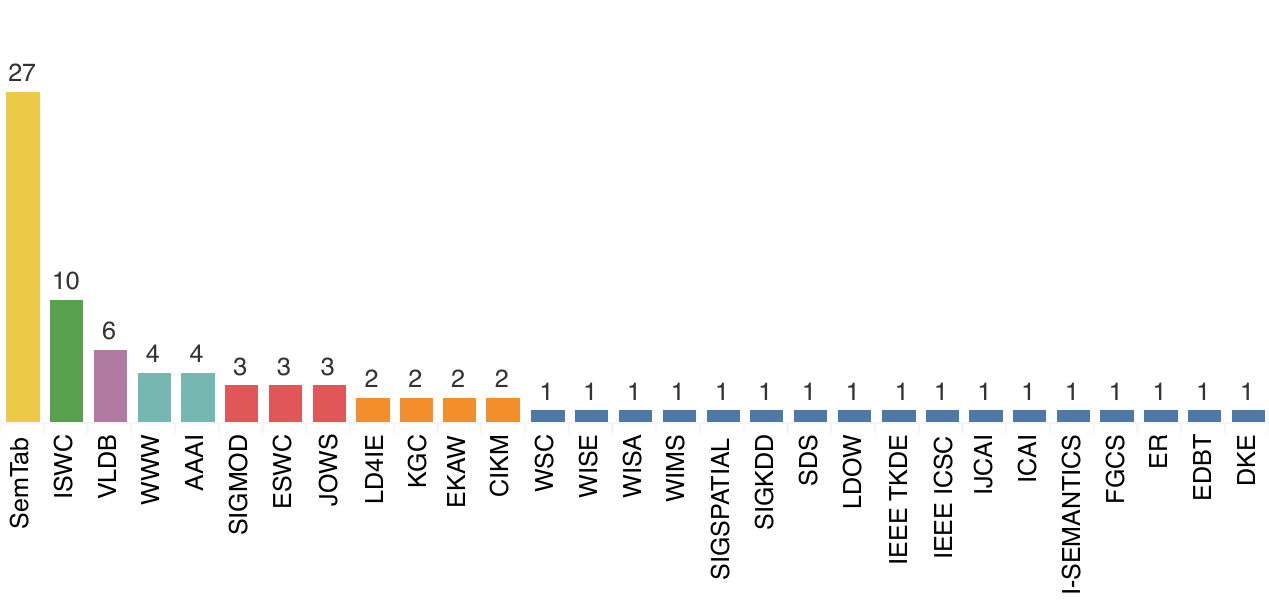}
  \caption{Number of approaches for each conference or journal. The extended version of the acronyms is shown in Table~\ref{tab:conf-acro} in Appendix~\ref{sec:additional-material}.}
  \label{fig:conferences-journal}
\end{figure}

\begin{figure}[H]
  \centering
  \includegraphics[width=0.9\textwidth]{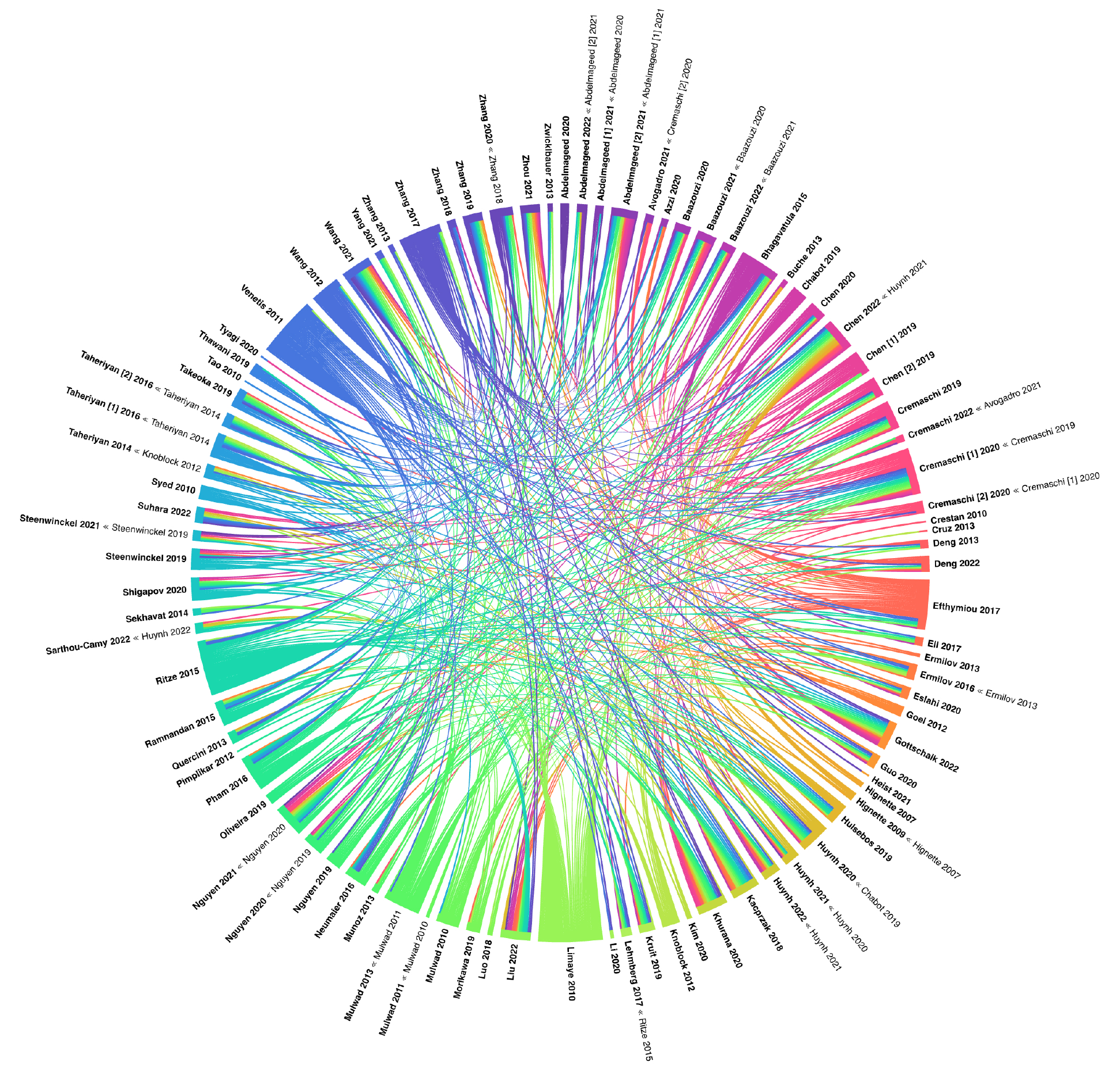}
  \caption{A cross-reference chart for the analysed papers.}
  \label{fig:crossref}
\end{figure}

Fig.~\ref{fig:conferences-journal} summarises the distribution of approaches in conferences and journals; from this analysis, it can be deduced that the \ac{sti} involves multiple research communities like Semantic Web, Data Management, \ac{ai}, and \ac{nlp}.
Fig.~\ref{fig:crossref} shows a graph with the cross-references between the articles\footnote{An online interactive visualisation of the cross-references chart is available at \href{https://observablehq.com/@elia-guarnieri-ws/cross-reference}{observablehq.com/@elia-guarnieri-ws/cross-reference}. A tabular representation is available at \href{https://public.tableau.com/app/profile/marco.cremaschi/viz/ChallengesandDirectionintheAnnotationofTabularData/Crossreference}{public.tableau.com/app/profile/marco.cremaschi/viz/ChallengesandDirectionintheAnnotationofTabularData/Crossreference}}. For some approaches it is indicated whether they are derived from previously published versions.

Table~\ref{table:comparison} provides a detailed comparison of all the approaches analysed.

\begin{landscape}
\begingroup
\scriptsize
  \centering
\begin{longtable}{llllllllllll}

    \rowcolor[HTML]{DDDDDD}
    \textbf{YEAR} & \textbf{AUTHOR}                           & \textbf{METHOD} & \textbf{PUBLICATION} & \rotatebox[origin=c]{90}{\textbf{CTA}} & \rotatebox[origin=c]{90}{\textbf{CPA}} & \rotatebox[origin=c]{90}{\textbf{CEA}} & \rotatebox[origin=c]{90}{\textbf{CNEA}} & \textbf{INDEX}                         & \rotatebox[origin=c]{90}{\textbf{CODE}} & \textbf{LICENCE} & \textbf{TRIPLE STORE}         \\
    \endhead

    2007          & Hignette et al.~\cite{Hignette2007}         & Unsup           & WISE                 & {\color[HTML]{93C47D} \textbf{✓}}      & {\color[HTML]{93C47D} \textbf{✓}}      & {\color[HTML]{FF0000} \textbf{✗}}      & {\color[HTML]{FF0000} \textbf{✗}}       & ---                                    & {\color[HTML]{FF0000} \textbf{✗}}       & ---              & Personal ontologies           \\ \rowcolor[HTML]{F3F3F3}
    2009          & Hignette et al.~\cite{Hignette2009}         & Unsup           & ESWC                 & {\color[HTML]{93C47D} \textbf{✓}}      & {\color[HTML]{93C47D} \textbf{✓}}      & {\color[HTML]{FF0000} \textbf{✗}}      & {\color[HTML]{FF0000} \textbf{✗}}       & ---                                    & {\color[HTML]{FF0000} \textbf{✗}}       & ---              & Personal ontologies           \\
    2009          & Tao et al.~\cite{Tao2009}                   & Unsup           & DKE                  & {\color[HTML]{FF0000} \textbf{✗}}      & {\color[HTML]{FF0000} \textbf{✗}}      & {\color[HTML]{FF0000} \textbf{✗}}      & {\color[HTML]{FF0000} \textbf{✗}}       & ---                                    & {\color[HTML]{FF0000} \textbf{✗}}       & ---              & Personal ontologies           \\ \rowcolor[HTML]{F3F3F3}
    2010          & Limaye et al.~\cite{Limaye2010}             & Unsup           & VLDB                 & {\color[HTML]{93C47D} \textbf{✓}}      & {\color[HTML]{93C47D} \textbf{✓}}      & {\color[HTML]{93C47D} \textbf{✓}}      & {\color[HTML]{FF0000} \textbf{✗}}       & ---                                    & {\color[HTML]{FF0000} \textbf{✗}}       & ---              & Yago                          \\
\end{longtable}
  \endgroup
\end{landscape}

\subsection{Techniques for Supervised and Unsupervised approaches}
\label{appendix-sec:supervised-unsupervised}

Table~\ref{table:unsupervised-approaches} and Table~\ref{table:supervised-approaches} display the techniques used by unsupervised and supervised approaches analysed in this survey.

\begingroup
\setlength{\tabcolsep}{5pt}
\renewcommand{\arraystretch}{1.2}

\begin{table}[ht]
\caption{List of unsupervised techniques and related approaches.}
\resizebox{\columnwidth}{!}{%
\begin{tabular}{lll}
\rowcolor[HTML]{DDDDDD}\textbf{Approach} & \textbf{Candidate Generation} & \textbf{Entity Disambiguation} \\
\rowcolor[HTML]{F3F3F3}
Limaye 2010 \cite{Limaye2010} & YAGO catalog & similarity \\
Syed 2010 \cite{Syed2010} & Wikitology & CTA \\
\rowcolor[HTML]{F3F3F3}
Wang 2012 \cite{Wang2012} & pattern matching & features \\
Munoz 2013 \cite{Munoz2013} & - & redirects \\
\rowcolor[HTML]{F3F3F3}
Ritze 2015 \cite{Ritze2015} & DBpedia lookup service & CTA \\
Ell 2017 \cite{Ell2017} & custom index & features \\
\rowcolor[HTML]{F3F3F3}
Zhang 2017 \cite{Zhang2017} & external lookup & similarity \\
\end{tabular}%
}
\label{table:unsupervised-approaches}
\end{table}
\endgroup

\begingroup
\setlength{\tabcolsep}{5pt}
\renewcommand{\arraystretch}{1.5}

\begin{table}[ht]
  \centering
  \caption{List of supervised techniques and related approaches.}
  \begin{adjustbox}{width=0.6\textwidth}
    \begin{tabular}{m{200pt} m{100pt}}
      \rowcolor[HTML]{DDDDDD}
      \textbf{Technique}                                                       & \textbf{Approaches}                                                                                                                                                                             \\
      \rowcolor[HTML]{FFFFFF}
      \ac{svm} classifier~\cite{Cortes1995SVM}                                 & Mulwad 2010~\cite{Mulwad2010}\newline Quercini 2013~\cite{Quercini2013}\newline Ermilov 2016~\cite{Ermilov2016}                                                                                 \\
      \rowcolor[HTML]{F3F3F3}
      Naive Bayes                                                              & Quercini 2013~\cite{Quercini2013}                                                                                                                                                               \\
      \rowcolor[HTML]{FFFFFF}
      Binary classifier                                                        & Zhang 2020~\cite{Zhang2020}                                                                                                                                                                     \\
      \rowcolor[HTML]{F3F3F3}
      \ac{crf} classifier~\cite{Lafferty2001CRF}                               & Goel 2012~\cite{Goel2012}\newline Knoblock 2012~\cite{Knoblock2012}\newline Zhang 2019~\cite{Zhang2019}\newline Ramnandan 2015~\cite{Ramnandan2015}\newline Guo 2020~\cite{Guo2020}                                                                                   \\
      \rowcolor[HTML]{FFFFFF}
      Markov Network~\cite{Kindermann1980Markov}                               & Mulwad 2011~\cite{Mulwad2010}\newline Mulwad 2013~\cite{Mulwad2013}\newline Zhang 2013~\cite{Zhang2013}\newline Bhagavatula 2015~\cite{Bhagavatula2015}\newline Takeoka 2019~\cite{Takeoka2019} \\
\end{tabular}

  \end{adjustbox}
      \label{table:supervised-approaches}
\end{table}
\endgroup

\begin{figure}[ht]

\includegraphics[width=\textwidth]{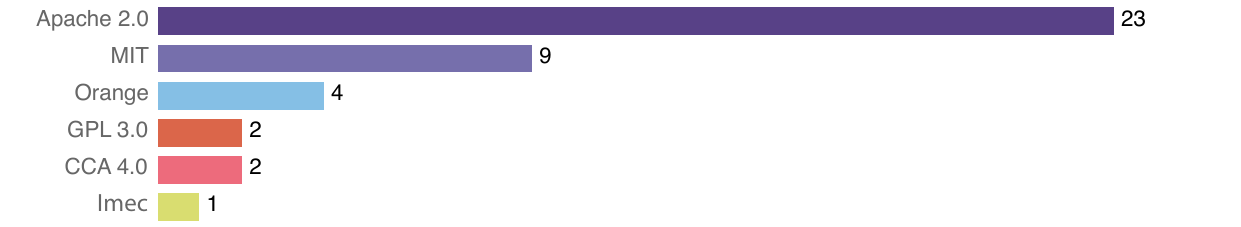}
  \caption{The distribution of the licenses adopted by \ac{sti} approaches.}
 \label{fig:licence}
\end{figure}

\section{STI tools}\label{appendix:b}
\label{appendix-sec:sti_tools_detailded}

\subsection{Tools analysis}
This Section analyses the tools that support \ac{sti}. Several tools can support table interpretation by providing data visualisation, manipulation, and statistical analysis features. This further analysis aims to gather information on the tools available for interpreting a table, or more generally, of a structured data source, to identify their features, strengths, and limitations and assess their suitability for different use cases.

Karma\footnote{\href{https://usc-isi-i2.github.io/karma}{usc-isi-i2.github.io/karma}}~\cite{Gupta2015} is an Open Source (Apache license 2.0) information integration tool that allows users to integrate data coming from different sources quickly and easily. Such sources include databases, spreadsheets, delimited text files, XML, JSON, CSV and Web API. Users can leverage different ontologies to annotate their data with standard vocabularies, ensuring accurate integration. Karma provides a responsive interface, fast processing, and batch mode for large datasets. Additionally, it offers data transformation scripts to convert data into a common format. A demonstration video\footnote{\href{https://www.youtube.com/watch?v=h3_yiBhAJIc}{www.youtube.com/watch?v=h3\_yiBhAJIc}} is available.

TableMiner+~\cite{Mazumdar2016} consists mainly of two components: a Java library that implements the homonym approach~\cite{Zhang2017} and an extension that constitutes a user-friendly \ac{ui} for the semantic annotation of Web tables. The current version of the tool corresponds to the alpha 1.0 development phase. Although the source code is available\footnote{\href{https://github.com/ziqizhang/sti}{github.com/ziqizhang/sti}}, the use is limited as the queries refer to Freebase. Even after applying modifications to exclude calls to the Freebase API and refer to DBpedia instead, an HTTPException error with code 500 prevented the \ac{sti} phase. Such an error probably indicates an incorrect formulation of the SPARQL queries within the TableMiner+ algorithm. Consequently, the analysis is based on the information provided in the paper~\cite{Mazumdar2016}.

The MAGIC tool\footnote{\href{https://github.com/IBCNServices/MAGIC}{github.com/IBCNServices/MAGIC}}~\cite{Steenwinckel2021} supports users to annotate data by following a structured pipeline to augment the semantics of a given table and provides a user-friendly graphical interface for column augmentation. However, it is important to note that the GUI lacks feedback on the produced annotations during table processing. The Instance Neighboring using Knowledge (INK) embedding technique can also enrich the table with information from the same dataset, semantically enriching the overall dataset with external linked data. A demonstration video of the MAGIC tool is available\footnote{\href{https://www.youtube.com/watch?v=ZhTKxcTBZNE}{www.youtube.com/watch?v=ZhTKxcTBZNE}}.

The MTab tool\footnote{\href{https://mtab.app/mtab}{mtab.app/mtab}}~\cite{Nguyen2021} is designed to automatically annotate data using \ac{kg}s. It enriches the original table data by adding schema and instance-level annotations. The tool supports multilingual tables and various formats such as Excel, CSV, and markdown tables. The system operates through a series of steps: preprocessing the tabular data and then enriching the table with semantic annotations using prediction and search functionality. Notably, the system achieved first place in the usability track of the SemTab challenge. It includes a \ac{ui} that offers features like table upload and an annotate button to initiate the process comprising the mentioned tasks. Additionally, the \ac{ui} allows users to search for entities in popular \ac{kg}s like Wikidata and DBpedia. However, it is important to note that the search functionality operates independently and does not assist users in other aspects of the annotation process. The MTab tool is accessible solely through an online web interface, with no available source code about the final version\footnote{\href{https://github.com/phucty/mtab_dev}{github.com/phucty/mtab\_dev} - dev version}.

MantisTable tool\footnote{\href{https://bitbucket.org/disco\_unimib/mantistable-tool}{bitbucket.org/disco\_unimib/mantistable-tool}}~\cite{Cremaschi2019} is a user-friendly \ac{ui} that facilitates the exploration of the annotation steps within the \ac{sti} process. Specifically, the tool enables users to visually explore and execute annotations through all the sub-tasks of the \ac{sti} process. It features a convenient right sidebar that offers additional information about each annotation in an info mode and allows manual editing of annotations using an edit mode widget. To support the annotation editing process, MantisTable leverages \ac{kg} summary profiles provided by ABSTAT~\cite{Spahiu2016}.

OpenRefine\footnote{\href{https://openrefine.org}{openrefine.org}} is an Open Resource tool able to support different formats such as TSV, CSV, Excel, XML, RDF/XML, JSON, N3 and LOG. It offers a workspace with several features, including the ability to export projects in different formats, explore data through filters and faceted exploration, apply clustering for grouping cells, modify cells individually or in groups using transformation rules, modify columns by renaming, deleting, or adding new ones, modify rows by filtering and flagging, display numerical value distributions, and uses extensions for additional functionality. OpenRefine offers functionality to reconcile against user-edited data on Wikidata or other Wikibase instances or reconcile against a local dataset\footnote{\href{https://openrefine.org/docs/manual/reconciling}{openrefine.org/docs/manual/reconciling}}.

Trifacta\footnote{\href{https://www.trifacta.com/}{trifacta.com}} is a collection of software used for data exploration and self-service data preparation for analysis. Trifacta works with cloud and on-premises data platforms. It is designed for analysts to explore, transform, and enrich raw data into clean and structured formats using techniques in \ac{ml}, data visualisation, human-computer interaction, and parallel processing for non-technical users to prepare data for various business processes such as analytics. It is composed of three main products: i)~Trifacta Wrangler - a connected desktop application used to transform data for downstream analytics and visualisation; ii)~Wrangler Pro - support for large data volumes, deployment options for both cloud and on-premises environments, and the capability to schedule and operationalise data preparation workflows, iii)~Wrangler Enterprises - self-service functionalities to explore and transform data with centralised management of security, governance and operationalisation.

Odalic addresses the limitations of TableMiner+ and is an Open Source tool. The code is available in Github\footnote{\href{https://github.com/odalic}{github.com/odalic}} and it can be easily installed via a Docker image\footnote{\href{https://github.com/odalic/odalic-docker}{github.com/odalic/odalic-docker}}. It provides a \ac{ui} for table interpretation, data export as linked data, and results review through user feedback. It supports CSV input and manual specification of relationships between columns. Odalic can work with any \ac{kg} accessible via SPARQL and perform \ac{sti} using query results from different \ac{kg} interrogations.

DataGraft+ASIA\footnote{\href{https://datagraft.io/}{datagraft.io}} refers to the integration of ASIA and Datagraft: ASIA is a tool to assist users in annotate tables and enrich their content using discovered links~\cite{Cutrona2019}, and Datagraft~\cite{Dumitru2016},
a cloud-based data transformation and publishing platform that supports the design and execution of transformations on tabular data.
DataGraft+ASIA refers to the integration of ASIA and Datagraft: ASIA is a tool to assist users in annotate tables and enrich their content using discovered links~\cite{Cutrona2019}, and Datagraft~\cite{Roman2018},
a cloud-based data transformation and publishing platform that supports the design and execution of transformations on tabular data.
Transformations in Datagraft include data cleaning functionalities but also RDF data generation based on table to RDF mappings (implemented with Grafterize framework\footnote{\href{https://www.eubusinessgraph.eu/grafterizer-2-0/}{eubusinessgraph.eu/grafterizer-2-0}}). ASIA supports the annotation of a table: it exploits vocabulary suggestions from the knowledge graph profiling tool ABSTAT~\cite{Spahiu2016} to annotate properties and column types, and entity linking algorithms (executed as services) to annotate cells. In addition, it uses data extension services to fetch data from third party sources adding new columns to the original table. The users control these operations from the \ac{gui} and check the results. As a consequence, DataGraft+ASIA supports two main applications: \ac{kg} generation and tabular data enrichment. ASIA-suported annotations are traduced to data transformations specifications; these specifications can executed, making the transformations repeatable, shareable, and extensible. Data can be exported in several tabular and RDF formats and published in the DataGraft platform.

DAGOBAH \ac{ui} is a web interface designed to visualise, validate, enrich, and manipulate the results of the \ac{sti} process through DAGOBAH API\footnote{\href{https://developer.orange.com/apis/table-annotation}{developer.orange.com/apis/table-annotation}}. The tool allows table data extracted from various pre-loaded benchmarks and additional files. It utilises DAGOBAH-SL, a RESTful API that implements pre-processing and \ac{sti} functionalities. DAGOBAH \ac{ui} addresses the problem of missing data by providing the possibility of adding additional columns using the background knowledge provided by the \ac{kg}s. The tool is only accessible through an online web interface, while the source code is unavailable.

\begingroup
\dashlinedash=1pt
\dashlinegap=1pt
\setlength{\tabcolsep}{5pt}
\renewcommand{\arraystretch}{1.5}

\begin{table}
    \centering
  \begin{adjustbox}{width=0.7\textwidth}
    \begin{tabular}{l:c:c:c:c:c:c:c:c:c:c:c:c:c}
      \rowcolor[HTML]{DDDDDD}
      \textbf{Functionalities}               & \rotatebox[origin=c]{90}{\textbf{Karma}} & \rotatebox[origin=c]{90}{\textbf{TableMiner+}} & \rotatebox[origin=c]{90}{\textbf{Magic}} & \rotatebox[origin=c]{90}{\textbf{MTab}} & \rotatebox[origin=c]{90}{\textbf{MantisTable}} & \rotatebox[origin=c]{90}{\textbf{STAN}} & \rotatebox[origin=c]{90}{\textbf{OpenRefine}} & \rotatebox[origin=c]{90}{\textbf{Trifacta}} & \rotatebox[origin=c]{90}{\textbf{Odalic}} & \rotatebox[origin=c]{90}{\textbf{DataGraft}} & \rotatebox[origin=c]{90}{\textbf{Dagobah}} & \rotatebox[origin=c]{90}{\textbf{SemTUI}} & \rotatebox[origin=c]{90}{\textbf{TableLLama}} \\
      \rowcolor[HTML]{FFFFFF}
      Import of tables                       & \color[HTML]{93C47D} \textbf{✓}          & \color[HTML]{FF0000} \textbf{✗}                & \color[HTML]{93C47D} \textbf{✓}          & \color[HTML]{93C47D} \textbf{✓}         & \color[HTML]{93C47D} \textbf{✓}                & \color[HTML]{93C47D} \textbf{✓}         & \color[HTML]{93C47D} \textbf{✓}               & \color[HTML]{93C47D} \textbf{✓}             & \color[HTML]{FF0000} \textbf{✗}           & \color[HTML]{93C47D} \textbf{✓}              & \color[HTML]{93C47D} \textbf{✓}            & \color[HTML]{93C47D} \textbf{✓}           & \color[HTML]{FF0000} \textbf{✗} \\
      \rowcolor[HTML]{F3F3F3}
      Import of tables  via API              & \color[HTML]{93C47D} \textbf{✓}          & \color[HTML]{FF0000} \textbf{✗}                & \color[HTML]{93C47D} \textbf{✓}          & \color[HTML]{93C47D} \textbf{✓}         & \color[HTML]{93C47D} \textbf{✓}                & \color[HTML]{93C47D} \textbf{✓}         & \color[HTML]{FF0000} \textbf{✗}               & \color[HTML]{93C47D} \textbf{✓}             & \color[HTML]{FF0000} \textbf{✗}           & \color[HTML]{93C47D} \textbf{✓}              & \color[HTML]{93C47D} \textbf{✓}            & \color[HTML]{93C47D} \textbf{✓}           & \color[HTML]{FF0000} \textbf{✗} \\
      \rowcolor[HTML]{FFFFFF}
      Import of ontologies                   & \color[HTML]{93C47D} \textbf{✓}          & \color[HTML]{FF0000} \textbf{✗}                & \color[HTML]{FF0000} \textbf{✗}          & \color[HTML]{FF0000} \textbf{✗}         & \color[HTML]{FF0000} \textbf{✗}                & \color[HTML]{FF0000} \textbf{✗}         & \color[HTML]{FF0000} \textbf{✗}               & \color[HTML]{FF0000} \textbf{✗}             & \color[HTML]{FF0000} \textbf{✗}           & \color[HTML]{93C47D} \textbf{✓}              & \color[HTML]{FF0000} \textbf{✗}            & \color[HTML]{FF0000} \textbf{✗}           & \color[HTML]{FF0000} \textbf{✗} \\
      \rowcolor[HTML]{F3F3F3}
      Definition of  personalised ontologies & \color[HTML]{FF0000} \textbf{✗}          & \color[HTML]{FF0000} \textbf{✗}                & \color[HTML]{FF0000} \textbf{✗}          & \color[HTML]{FF0000} \textbf{✗}         & \color[HTML]{FF0000} \textbf{✗}                & \color[HTML]{93C47D} \textbf{✓}         & \color[HTML]{FF0000} \textbf{✗}               & \color[HTML]{FF0000} \textbf{✗}             & \color[HTML]{FF0000} \textbf{✗}           & \color[HTML]{93C47D} \textbf{✓}              & \color[HTML]{FF0000} \textbf{✗}            & \color[HTML]{FF0000} \textbf{✗}            & \color[HTML]{FF0000} \textbf{✗} \\
      \rowcolor[HTML]{FFFFFF}
      Semi-automatic annotation/HITL         & \color[HTML]{93C47D} \textbf{✓}          & \color[HTML]{93C47D} \textbf{✓}                & \color[HTML]{93C47D} \textbf{✓}          & \color[HTML]{93C47D} \textbf{✓}         & \color[HTML]{93C47D} \textbf{✓}                & \color[HTML]{FF0000} \textbf{✗}         & \color[HTML]{FF0000} \textbf{✗}               & \color[HTML]{FF0000} \textbf{✗}             & \color[HTML]{93C47D} \textbf{✓}           & \color[HTML]{FF0000} \textbf{✗}              & \color[HTML]{93C47D} \textbf{✓}            & \color[HTML]{FF0000} \textbf{✗}           & \color[HTML]{FF0000} \textbf{✗} \\
      \rowcolor[HTML]{F3F3F3}
      Annotation suggestions                 & \color[HTML]{93C47D} \textbf{✓}          & \color[HTML]{FF0000} \textbf{✗}                & \color[HTML]{93C47D} \textbf{✓}          & \color[HTML]{FF0000} \textbf{✗}         & \color[HTML]{93C47D} \textbf{✓}                & \color[HTML]{93C47D} \textbf{✓}         & \color[HTML]{FF0000} \textbf{✗}               & \color[HTML]{FF0000} \textbf{✗}             & \color[HTML]{FF0000} \textbf{✗}           & \color[HTML]{93C47D} \textbf{✓}              & \color[HTML]{FF0000} \textbf{✗}            & \color[HTML]{93C47D} \textbf{✓}           & \color[HTML]{FF0000} \textbf{✗} \\
      \rowcolor[HTML]{FFFFFF}
      Auto-complete support                  & \color[HTML]{93C47D} \textbf{✓}          & \color[HTML]{FF0000} \textbf{✗}                & \color[HTML]{FF0000} \textbf{✗}          & \color[HTML]{FF0000} \textbf{✗}         & \color[HTML]{93C47D} \textbf{✓}                & \color[HTML]{FF0000} \textbf{✗}         & \color[HTML]{93C47D} \textbf{✓}               & \color[HTML]{FF0000} \textbf{✗}             & \color[HTML]{FF0000} \textbf{✗}           & \color[HTML]{93C47D} \textbf{✓}              & \color[HTML]{FF0000} \textbf{✗}            & \color[HTML]{93C47D} \textbf{✓}            & \color[HTML]{FF0000} \textbf{✗} \\
      \rowcolor[HTML]{F3F3F3}
      Subject column detection               & \color[HTML]{93C47D} \textbf{✓}          & \color[HTML]{93C47D} \textbf{✓}                & \color[HTML]{FF0000} \textbf{✗}          & \color[HTML]{93C47D} \textbf{✓}         & \color[HTML]{93C47D} \textbf{✓}                & \color[HTML]{FF0000} \textbf{✗}         & \color[HTML]{FF0000} \textbf{✗}               & \color[HTML]{FF0000} \textbf{✗}             & \color[HTML]{93C47D} \textbf{✓}           & \color[HTML]{FF0000} \textbf{✗}              & \color[HTML]{93C47D} \textbf{✓}            & \color[HTML]{93C47D} \textbf{✓}           & \color[HTML]{FF0000} \textbf{✗} \\
      \rowcolor[HTML]{FFFFFF}
      CEA                                    & \color[HTML]{FF0000} \textbf{✗}          & \color[HTML]{93C47D} \textbf{✓}                & \color[HTML]{93C47D} \textbf{✓}          & \color[HTML]{93C47D} \textbf{✓}         & \color[HTML]{93C47D} \textbf{✓}                & \color[HTML]{FF0000} \textbf{✗}         & \color[HTML]{93C47D} \textbf{✓}               & \color[HTML]{FF0000} \textbf{✗}             & \color[HTML]{93C47D} \textbf{✓}           & \color[HTML]{93C47D} \textbf{✓}              & \color[HTML]{93C47D} \textbf{✓}            & \color[HTML]{93C47D} \textbf{✓}           & \color[HTML]{93C47D} \textbf{✓} \\
      \rowcolor[HTML]{F3F3F3}
      CTA                                    & \color[HTML]{FF0000} \textbf{✗}          & \color[HTML]{93C47D} \textbf{✓}                & \color[HTML]{93C47D} \textbf{✓}          & \color[HTML]{93C47D} \textbf{✓}         & \color[HTML]{93C47D} \textbf{✓}                & \color[HTML]{FF0000} \textbf{✗}         & \color[HTML]{93C47D} \textbf{✓}               & \color[HTML]{FF0000} \textbf{✗}             & \color[HTML]{93C47D} \textbf{✓}           & \color[HTML]{93C47D} \textbf{✓}              & \color[HTML]{93C47D} \textbf{✓}            & \color[HTML]{93C47D} \textbf{✓}           & \color[HTML]{93C47D} \textbf{✓} \\
      \rowcolor[HTML]{FFFFFF}
      CPA (NE columns)                       & \color[HTML]{FF0000} \textbf{✗}          & \color[HTML]{93C47D} \textbf{✓}                & \color[HTML]{93C47D} \textbf{✓}          & \color[HTML]{93C47D} \textbf{✓}         & \color[HTML]{93C47D} \textbf{✓}                & \color[HTML]{FF0000} \textbf{✗}         & \color[HTML]{93C47D} \textbf{✓}               & \color[HTML]{FF0000} \textbf{✗}             & \color[HTML]{93C47D} \textbf{✓}           & \color[HTML]{93C47D} \textbf{✓}              & \color[HTML]{93C47D} \textbf{✓}            & \color[HTML]{93C47D} \textbf{✓}           & \color[HTML]{93C47D} \textbf{✓} \\
      \rowcolor[HTML]{F3F3F3}
      CPA (LIT columns)                      & \color[HTML]{FF0000} \textbf{✗}          & \color[HTML]{93C47D} \textbf{✓}                & \color[HTML]{93C47D} \textbf{✓}          & \color[HTML]{93C47D} \textbf{✓}         & \color[HTML]{93C47D} \textbf{✓}                & \color[HTML]{FF0000} \textbf{✗}         & \color[HTML]{93C47D} \textbf{✓}               & \color[HTML]{FF0000} \textbf{✗}             & \color[HTML]{93C47D} \textbf{✓}           & \color[HTML]{93C47D} \textbf{✓}              & \color[HTML]{93C47D} \textbf{✓}            & \color[HTML]{93C47D} \textbf{✓}           & \color[HTML]{FF0000} \textbf{✗} \\
      \rowcolor[HTML]{FFFFFF}
      Table manipulation                     & \color[HTML]{93C47D} \textbf{✓}          & \color[HTML]{FF0000} \textbf{✗}                & \color[HTML]{FF0000} \textbf{✗}          & \color[HTML]{FF0000} \textbf{✗}         & \color[HTML]{93C47D} \textbf{✓}                & \color[HTML]{FF0000} \textbf{✗}         & \color[HTML]{93C47D} \textbf{✓}               & \color[HTML]{93C47D} \textbf{✓}             & \color[HTML]{FF0000} \textbf{✗}           & \color[HTML]{93C47D} \textbf{✓}              & \color[HTML]{FF0000} \textbf{✗}            & \color[HTML]{93C47D} \textbf{✓}           & \color[HTML]{FF0000} \textbf{✗} \\
      \rowcolor[HTML]{F3F3F3}
      Automatic table extension              & \color[HTML]{FF0000} \textbf{✗}          & \color[HTML]{FF0000} \textbf{✗}                & \color[HTML]{93C47D} \textbf{✓}          & \color[HTML]{FF0000} \textbf{✗}         & \color[HTML]{FF0000} \textbf{✗}                & \color[HTML]{FF0000} \textbf{✗}         & \color[HTML]{93C47D} \textbf{✓}               & \color[HTML]{93C47D} \textbf{✓}             & \color[HTML]{FF0000} \textbf{✗}           & \color[HTML]{93C47D} \textbf{✓}              & \color[HTML]{FF0000} \textbf{✗}            & \color[HTML]{93C47D} \textbf{✓}           & \color[HTML]{FF0000} \textbf{✗} \\
      \rowcolor[HTML]{FFFFFF}
      Visualisation of annotations           & \color[HTML]{FF0000} \textbf{✗}          & \color[HTML]{FF0000} \textbf{✗}                & \color[HTML]{93C47D} \textbf{✓}          & \color[HTML]{93C47D} \textbf{✓}         & \color[HTML]{93C47D} \textbf{✓}                & \color[HTML]{FF0000} \textbf{✗}         & \color[HTML]{FF0000} \textbf{✗}               & \color[HTML]{FF0000} \textbf{✗}             & \color[HTML]{FF0000} \textbf{✗}           & \color[HTML]{93C47D} \textbf{✓}              & \color[HTML]{93C47D} \textbf{✓}            & \color[HTML]{93C47D} \textbf{✓}           & \color[HTML]{FF0000} \textbf{✗} \\
      \rowcolor[HTML]{F3F3F3}
      Auto save                              & \color[HTML]{93C47D} \textbf{✓}          & \color[HTML]{FF0000} \textbf{✗}                & \color[HTML]{FF0000} \textbf{✗}          & \color[HTML]{FF0000} \textbf{✗}         & \color[HTML]{FF0000} \textbf{✗}                & \color[HTML]{FF0000} \textbf{✗}         & \color[HTML]{93C47D} \textbf{✓}               & \color[HTML]{93C47D} \textbf{✓}             & \color[HTML]{FF0000} \textbf{✗}           & \color[HTML]{93C47D} \textbf{✓}              & \color[HTML]{FF0000} \textbf{✗}            & \color[HTML]{FF0000} \textbf{✗}           & \color[HTML]{FF0000} \textbf{✗} \\
      \rowcolor[HTML]{FFFFFF}
      Export mapping                         & \color[HTML]{93C47D} \textbf{✓}          & \color[HTML]{93C47D} \textbf{✓}                & \color[HTML]{93C47D} \textbf{✓}          & \color[HTML]{93C47D} \textbf{✓}         & \color[HTML]{93C47D} \textbf{✓}                & \color[HTML]{93C47D} \textbf{✓}         & \color[HTML]{FF0000} \textbf{✗}               & \color[HTML]{FF0000} \textbf{✗}             & \color[HTML]{93C47D} \textbf{✓}           & \color[HTML]{93C47D} \textbf{✓}              & \color[HTML]{93C47D} \textbf{✓}            & \color[HTML]{93C47D} \textbf{✓}           & \color[HTML]{FF0000} \textbf{✗} \\
      \rowcolor[HTML]{F3F3F3}
      Export RDF triplets                    & \color[HTML]{93C47D} \textbf{✓}          & \color[HTML]{93C47D} \textbf{✓}                & \color[HTML]{93C47D} \textbf{✓}          & \color[HTML]{93C47D} \textbf{✓}         & \color[HTML]{93C47D} \textbf{✓}                & \color[HTML]{93C47D} \textbf{✓}         & \color[HTML]{93C47D} \textbf{✓}               & \color[HTML]{FF0000} \textbf{✗}             & \color[HTML]{93C47D} \textbf{✓}           & \color[HTML]{93C47D} \textbf{✓}              & \color[HTML]{93C47D} \textbf{✓}            & \color[HTML]{93C47D} \textbf{✓}           & \color[HTML]{FF0000} \textbf{✗} \\
      \rowcolor[HTML]{FFFFFF}
      Open Source                            & \color[HTML]{93C47D} \textbf{✓}          & \color[HTML]{93C47D} \textbf{✓}                & \color[HTML]{93C47D} \textbf{✓}          & \color[HTML]{FF0000} \textbf{✗}         & \color[HTML]{93C47D} \textbf{✓}                & \color[HTML]{93C47D} \textbf{✓}         & \color[HTML]{93C47D} \textbf{✓}               & \color[HTML]{FF0000} \textbf{✗}             & \color[HTML]{93C47D} \textbf{✓}           & \color[HTML]{93C47D} \textbf{✓}              & \color[HTML]{FF0000} \textbf{✗}            & \color[HTML]{93C47D} \textbf{✓}           & \color[HTML]{FF0000} \textbf{✗}  \\
      \rowcolor[HTML]{F3F3F3}
    \end{tabular}
  \end{adjustbox}
    \caption{table}{Comparison of semantic table interpretation tools.}\label{tab:ComparisonTools}
\end{table}
\endgroup

SemTUI is an Open Source\footnote{\href{https://github.com/I2Tunimib}{github.com/I2Tunimib}} web-based application composed of a frontend module built with React and Redux, and a backend server. SemTUI focuses on tabular data annotation and extension tasks and is decoupled from DataGraft. It implements a ``link \& extend'' paradigm, inspired by linked open data, but more general and supported by several data linking and data extension services (e.g., geocoding services, data services, and so on).
Compared to its first version ASIA, it provides a better \ac{gui}, and more functionalities to support human-in-the-loop tabular data annotation and extension. It is integrated with end-to-end \ac{sti} algorithms (improved from~\cite{Cremaschi2022}) that provide a first annotation of an input table, which users are expected to revise and manipulate. Particular attention is given to the revision of entity linking, which exploits a recent confidence-aware algorithm~\cite{Avogadro2023}.


Table~\ref{tab:ComparisonTools} provides a comparison between tools.

\subsection{Comparison of tools with GUI}
\label{sec:tools}

In this Section, we compare tools for \ac{sti} or supporting \ac{sti} tasks that also provide a \ac{gui}; these tools are introduced to assist users in applications listed in Section~\ref{sec:application_purtpose}. We found twelve tools with these features: Karma\footnote{\href{https://usc-isi-i2.github.io/karma}{usc-isi-i2.github.io/karma}}~\cite{Gupta2015},
TableMiner+~\cite{Mazumdar2016},
MAGIC\footnote{\href{https://github.com/IBCNServices/MAGIC}{github.com/IBCNServices/MAGIC}}~\cite{Steenwinckel2021},
MTab tool\footnote{\href{https://mtab.app/mtab}{mtab.app/mtab}}~\cite{Nguyen2021},
MantisTable\footnote{\href{https://bitbucket.org/disco\_unimib/mantistable-tool}{bitbucket.org/disco\_unimib/mantistable-tool}}~\cite{Cremaschi2019},
OpenRefine\footnote{\href{https://openrefine.org}{openrefine.org}},
Trifacta\footnote{\href{https://www.trifacta.com/}{trifacta.com}},
Odalic\footnote{\href{https://github.com/odalic}{github.com/odalic}}~\cite{Knap2017},
DataGraft+ASIA\footnote{\href{https://datagraft.io/}{datagraft.io}}~\cite{Dumitru2016,Cutrona2019},
DAGOBAH \ac{ui}~\cite{Sarthou-Camy2022},
SemTUI\footnote{\href{https://github.com/I2Tunimib}{github.com/I2Tunimib}},
MantisTableUI\footnote{\href{https://mantistable.datai.disco.unimib.it/}{mantistable.datai.disco.unimib.it/}}~\cite{cremaschimantistable}.

A short description for each tool is provided in Section~\ref{sec:tools}. Table~\ref{tab:ComparisonTools} in Appendix~\ref{appendix-sec:sti_tools_detailded} provide a comparison of the tools based on some key features: i)~\textbf{\textit{table import}}, ii)~\textbf{\textit{ontology support}}, iii)~\textbf{\textit{ontology support}}, iv)~\textit{\textbf{semi-automatic semantic annotation/HITL}}, v)~\textbf{\textit{semantic annotation suggestions}}, vi)~\textbf{\textit{auto-complete support for the semantic annotation process}}, vii)~\textbf{\textit{STI sub-task}}, viii)~\textit{\textbf{table manipulation}}, ix)~\textbf{\textit{automatic table extension}}, x)~\textbf{\textit{graphical visualisation of semantic annotations}}, xi)~\textbf{\textit{auto save of current user workspace}}, xii)~\textbf{\textit{API services and SPARQL endpoint}}, xiii)~\textbf{\textit{export mapping and RDF triples}}, and xiv)~\textbf{\textit{open source}}.

\textbf{\textit{Table import}}: table import is a crucial functionality for enhancing usability from the users' perspective. It allows users to work with their tables without requiring manual data transfer, ensuring a smoother user experience. Additionally, users can perform various data operations, such as filtering, sorting, aggregating, or visualising the data. Typically, tools provide two main methods for enabling table import: wizards and APIs. Among the twelve analysed tools, only two (TableMiner+ and Odalic) do not offer wizard functionality, while three (TableMiner+, Odalic, and OpenRefine) lack API functionalities for table import.

\textbf{\textit{Ontology support}}: among tables, ontologies play another important role that would enhance users' satisfaction and usability. They allow working with familiar and domain-specific terminology, ensuring accuracy and consistency in the semantic representation of the table. Moreover, the final annotated tables might be easily integrated with other systems. Tools reviewed in this survey offer users two options: \textit{reusing} existing ontologies, either by importing an ontology or searching vocabularies used in existing \ac{kg}, and \textit{creating} personalised ones. Out of the 12 tools, 10 do not allow importing or reusing ontologies. Only Karma and DataGraft+ASIA provide users with such functionality. Similarly, only DataGraft+ASIA supports users in defining their personalised ontology to annotate the data. Karma supports users in importing different ontologies and combining them.

\textbf{\textit{Semi-automatic semantic annotation/HITL}}: semi-automatic semantic annotation and human-in-the-loop ability allows users to review annotations. Users can correct, judge ambiguous or unclear information and refine automated annotations, improving the accuracy of the annotations. Almost half of the tools lack human-in-the-loop functionality. Such is implemented only in Karma, TableMiner+, MAGIC, MTab, MantisTable, Odalic, and DAGOBAH \ac{ui}.

\textbf{\textit{Semi-automatic semantic annotation/HITL}}: semi-automatic semantic annotation and human-in-the-loop ability allows users to review annotations. Users can correct and judge ambiguous or unclear information and refine automated annotations, improving the accuracy of the annotations. Almost half of the tools lack human-in-the-loop functionality. Such is implemented only in Karma, TableMiner+, MAGIC, MTab, MantisTable, Odalic, and DAGOBAH \ac{ui}.

\textbf{\textit{Semantic annotation suggestions}}: such functionality saves time and effort for users as it empowers tools to automatically generate suggestions for the semantic annotation. In particular, for domains that users do not know about, such functionality can ensure more accurate and consistent annotations, reducing the risk of errors or inconsistencies. Seven tools, Karma, MAGIC, MantisTable, DataGraft+ASIA and SemTUI, provide users with annotation suggestions.

\textbf{\textit{Auto-complete support for the semantic annotation process}}: auto-complete functionality speeds up the semantic annotation process by providing suggestions and/or completions for annotations. It drastically reduces the time and effort required for users to manually enter or search for an appropriate semantic term to use in the annotation process. Moreover, it prevents errors as a result of miss-spelling. Karma, MantisTable, OpenRefine, Datagraft+ASIA, MantisTableUI and SemTUI implement auto-complete functionalities.

\textbf{\textit{STI sub-tasks}}: not every \ac{sti} sub-task is implemented in the available tools reviewed in this Section. Full implementation of the \ac{sti} process would allow users full support to annotate every table element accurately. TableMiner+ performs reconciliation by annotating cells with specific entities within the \ac{kg} and identification of the S-column in a semi-automatic manner. The first feature is common to OpenRefine. Trifacta is the weakest tools concerning the implementation of \ac{sti} sub-tasks, while TableMiner+, MTab, MantisTable, Odalic, MantisTableUI, DAGOBAH \ac{ui} and SemTUI fully implement such functionalities.

\textbf{\textit{Table manipulation}}: table manipulation functionality allows users to clean and preprocess the data before performing semantic annotation. For example, among tools that implement such functionality, Karma and OpenRefine allow users to manipulate tables and refine them by allowing column modification, such as renaming, eliminating, or changing their order. This ensures that the data is in the desired form, removing any inconsistencies or errors that could affect the quality of the annotations. Despite being an important functionality, such is implemented only by almost half of the available tools (\ie Karma, OpenRefine, Trifacta, and DataGraft+ASIA)
OpenRefine has features that are not common to others in our analysis. For example the automatic creation of new columns, and the exploration of the cells through the facets.

\textbf{\textit{Automatic table extension}}: this is an important functionality, especially for data enrichment applications as it automatically retrieves additional data from external sources. Furthermore, such functionality can keep the semantic model up-to-date, reflecting the latest knowledge and insights despite the updated data. Only MAGIC, OpenRefine, Trifacta, DataGraft+ASIA, and SemTUI users might benefit from such functionality.

\textbf{\textit{Graphical visualization of semantic annotations}}: graphical visualisation of semantic annotations supports users with a visual representation of the annotated data, allowing them to understand better the relationships and the structure of the data within the table. Moreover, it allows users to identify inter-dependencies between different table parts, enhancing the overall semantic understanding. TableMiner+, OpenRefine, Trifacta and Odalic do not allow users to visualise annotations.

\textbf{\textit{Auto save of current user workspace}}: auto-saving ensures that the user's work is continuously saved, preventing data loss from system failure. It allows users to perform changes and modifications with the assurance that their progress is automatically saved without worrying about manually saving their work. Such functionality might serve as a form of version control, enabling users to review and revert to previous versions if needed. Karma, OpenRefine, and Trifacta have the automatic saving feature of the current work status.

\textbf{\textit{Export mapping and RDF triples}}: exporting mappings and RDF triples allows the data annotated in the tool to be shared and integrated with other systems and applications, enabling interoperability. Most available tools (Karma, MAGIC, MantisTable, Odalic, MantisTableUI, DataGraft+ASIA and DAGOBAH UI) allow both exports. Karma allows export in RDF format or JSON-LD. Regarding the export of tabular data, Karma uses the R2RML format to highlight the annotations between the table and the ontology. The other tools allow both exports. Only Trifacta does not implement any of these functionalities.

\textbf{\textit{Open Source}}: open-source tools provide transparency in their functionality, allowing users to understand how the tool works and ensuring there are no hidden or proprietary algorithms or biases. All software under such a license might be easily customised or modified. Only MTab, MantisTableUI, Trifacta and DAGOBAH \ac{ui} do not provide the code in an open-source license.
A detailed description of all tools can be found in the Appendix~\ref{appendix-sec:sti_tools_detailded}.

\section{Gold Standards}\label{appendix-sec:gold-standard-statistics}
\label{sec:gold-standards}

\ac{gss} serve as a benchmark to measure the performance of various approaches and systems. Moreover, \ac{gss} allow identifying the strengths and weaknesses of existing methods thus helping in the advancements of the state-of-the-art performance. Although several approaches deal with semantic annotations on tabular data, there are limited \ac{gss} for assessing the quality of these annotations. The main ones are T2Dv2, Limaye, Tough Table and SemTab. Table~\ref{table:stats-gss} in Appendix~\ref{appendix-sec:gold-standard-statistics} shows statistics for the \ac{gss}\footnote{\href{https://unimib-datai.github.io/sti-website/datasets/}{unimib-datai.github.io/sti-website/datasets/}}.

This Section considers only publicly available \ac{gss}.
\ac{gss} for \ac{sti} approaches can be classified based on several dimensions.

\textbf{\textit{Domain:}} \ac{gss} can target a certain domain or cover a broad range of domains. Most of the available \ac{gss} target cross-domain annotations. However, there are also some domain-specific \ac{gs} such as IMDB~\cite{Zhang2017}, MusicBrainz~\cite{Zhang2017}, and BiodivTab~\cite{Abdelmageed2021biodivtab}.

\textbf{\textit{Annotation coverage:}} \ac{gss} differ in the level of granularity at which annotations are provided. This can range from fine-grained annotations capturing detailed semantic information at the cell or column level (classes, entities, predicates, \eg~WebTableStiching~\cite{Ritze2017}, 2T~\cite{Cutrona2020}, and SemTab), to coarse-grained annotations(\eg~LimayeAll~\cite{Limaye2010}, GitTables~\cite{Hulsebos2023}, TURL~\cite{Deng2022}), providing broader semantic context at the table or dataset level.

\textbf{\textit{\ac{nil} annotation:}} The previous Sections described how an approach should consider NIL annotations, which can be used for \ac{kg} extension and construction. However, only three datasets currently consider this type of annotation (\ie~\cite{Marzocchi2022}, SemTab2022 R3 Biodiv, and SemTab 2022 R3 GitTables). This underlines how more significant effort is needed on the part of the scientific community towards this key challenge.

\textbf{\textit{Dataset size:}} in \ac{sti}, \ac{gss} should be composed of tables of varying sizes, from small to very large tables. This would allow systems to measure and evaluate their scalability performance. A significant proportion of \ac{gss} are relatively small (T2Dv2~\cite{Ritze2017}, WebTableStiching~\cite{Ritze2017}, Limaye~\cite{Limaye2010}, MusicBrainz~\cite{Zhang2017}, IMDB~\cite{Zhang2017}, Taheriyan~\cite{Taheriyan2016-1}, 2T~\cite{Cutrona2020}, REDTab~\cite{Singh2022}, BiodivTab~\cite{Abdelmageed2021biodivtab}, and TSOTSA~\cite{Jiomekong2022}). In contrast, only a handful of them are larger (MammoTab\footnote{\href{https://unimib-datai.github.io/mammotab-docs/}{unimib-datai.github.io/mammotab-docs/}}~\cite{Marzocchi2022}, SOTAB~\cite{Korini2022}, Wikary~\cite{Mazurek2022}, GitTables~\cite{Hulsebos2023}, and TURL~\cite{Deng2022}).

\textbf{\textit{Documentation:}} \ac{gss} might be accompanied with documentation. It is important that certain factors, such as the availability of guidelines, code availability, documentation on annotation conventions, and examples that aid in understanding and applying the \ac{gs}, are clearly stated. The list of well-curated and documented datasets is limited (\ie 2T~\cite{Cutrona2020}, MammoTab~\cite{Marzocchi2022}, GitTables~\cite{Hulsebos2023}, REDTab~\cite{Singh2022}, SOTAB~\cite{Korini2022}, and BiodivTab~\cite{Abdelmageed2021biodivtab}).

Table~\ref{table:stats-gss} provides detailed statistics about \ac{gss}.
\begingroup
\dashlinedash=1pt
\dashlinegap=1pt
\setlength{\tabcolsep}{5pt}
\renewcommand{\arraystretch}{1.5}
\newcolumntype{M}[1]{>{\centering\let\newline\\\arraybackslash\hspace{0pt}}m{#1}}
\hyphenchar\font=-1
\makeatletter
\newcommand*{\Xbar}{}%
\DeclareRobustCommand*{\Xbar}{%
  \mathpalette\@Xbar{}%
}
\newcommand*{\@Xbar}[2]{%
  \sbox0{$#1\mathrm{X}\m@th$}%
  \sbox2{$#1X\m@th$}%
  \rlap{%
    \hbox to\wd2{%
      \hfill
      $\overline{%
          \vrule width 0pt height\ht0 %
          \kern\wd0 %
        }$%
    }%
  }%
  \copy2 %
}
\makeatother

\begin{table}[ht]
  \caption{Statistics for the most common datasets. `---' indicates unknown.}
  \centering
    \label{table:stats-gss}
  \begin{adjustbox}{width=\textwidth}
\begin{tabular}{m{2cm}l:c:M{2.5cm}:M{2.5cm}:M{1.3cm}:M{1.3cm}:M{1.3cm}:m{1.8cm}:m{3cm}}
      \rowcolor[HTML]{DDDDDD}
      \multicolumn{2}{l:}{\textbf{GS}}                                                                  & \multicolumn{1}{c:}{\textbf{Tables}}  & \multicolumn{1}{M{2.5cm}:}{\textbf{Cols\newline (min | max | $\Bar{x}$)}}   & \multicolumn{1}{M{2.5cm}:}{\textbf{Rows\newline (min | max | $\Bar{x}$)}}                        & \multicolumn{1}{M{1.3cm}:}{\textbf{Classes}}                                                         & \multicolumn{1}{M{1.3cm}:}{\textbf{Entities}} & \multicolumn{1}{M{1.3cm}:}{\textbf{Pred.}} & \multicolumn{1}{m{1.8cm}:}{\textbf{KG}}  & \multicolumn{1}{m{3cm}}{\textbf{Used for \newline validation by}}                                                                                                                                                                                                                                                                                                                  \\
\end{tabular}
  \end{adjustbox}
\end{table}

\begin{table}[ht]
  \begin{adjustbox}{width=\textwidth}
    \begin{tabular}{m{2cm}l:c:M{2.5cm}:M{2.5cm}:M{1.3cm}:M{1.3cm}:M{1.3cm}:m{1.8cm}:m{3cm}}
      \rowcolor[HTML]{DDDDDD}
      \multicolumn{2}{l:}{\textbf{GS}}      & \multicolumn{1}{c:}{\textbf{Tables}}         & \multicolumn{1}{M{2.5cm}:}{\textbf{Cols\newline (min | max | $\Bar{x}$)}} & \multicolumn{1}{M{2.5cm}:}{\textbf{Rows\newline (min | max | $\Bar{x}$)}}                        & \multicolumn{1}{M{1.3cm}:}{\textbf{Classes}}                                                     & \multicolumn{1}{M{1.3cm}:}{\textbf{Entities}}                 & \multicolumn{1}{M{1.3cm}:}{\textbf{Pred.}} & \multicolumn{1}{m{1.8cm}:}{\textbf{KG}} & \multicolumn{1}{m{3cm}}{\textbf{Used for \newline validation by}}                                                                                       \\
                                                                                                              & \cellcolor[HTML]{F3F3F3}\textbf{~R1~} & \cellcolor[HTML]{F3F3F3}180                                                 & \cellcolor[HTML]{F3F3F3}\numprint{802}\newline (\numprint{1} | \numprint{8} | \numprint{4.46})   & \cellcolor[HTML]{F3F3F3}\numprint{194}K\newline (\numprint{6} | \numprint{15.5}K | \numprint{1.08}K) & \cellcolor[HTML]{F3F3F3}\numprint{539}        & \cellcolor[HTML]{F3F3F3}\numprint{667}K    & \cellcolor[HTML]{F3F3F3}\numprint{56,5}K & Wikidata\newline DBpedia                                                                                                                                                                                                                                                                                                                                                           \\
\end{tabular}
  \end{adjustbox}
\end{table}

\endgroup

\section{Additional Material}
\label{sec:additional-material}

Table~\ref{tab:conf-acro} presents the acronyms with the respective names of the conferences or journals to which the articles analysed in this survey were submitted.

\begingroup
\dashlinedash=1pt
\dashlinegap=1pt
\setlength{\tabcolsep}{5pt}
\renewcommand{\arraystretch}{1.5}

\begin{table}[H]
  \caption{Conferences and journals acronyms.}\label{tab:conf-acro}
  \centering
  \begin{adjustbox}{width=.7\textwidth}
    \begin{tabular}{p{60pt}l}
      \rowcolor[HTML]{DDDDDD}
      \textbf{Acronym} & \textbf{Name}                                                        \\
      AAAI             & Association for the Advancement of Artificial Intelligence                 \\
      \rowcolor[HTML]{F3F3F3}
      CIKM             & The Conference on Information and Knowledge Management                     \\
      DKE             & Data \& Knowledge Engineering \\
      \rowcolor[HTML]{F3F3F3}
      EDBT             &  International Conference on Extending Database Technology \\
      EKAW             & European Knowledge Acquisition Workshop \\
      \rowcolor[HTML]{F3F3F3}
      ER             & International Conference on Conceptual Modeling                                            \\
      ESWC             & Extended Semantic Web Conference                                           \\
\end{tabular}
  \end{adjustbox}
\end{table}
\endgroup

Table~\ref{tab:selection_table} provides support in selecting approaches in relation to various attributes, such as Method, Tasks, Code availability, License and Triple store.

\newcommand{\minitab}[2][l]{\begin{tabular}{#1}#2\end{tabular}}

\dashlinedash=1pt
\dashlinegap=1pt

\begin{table}[ht]
        \centering
        \caption{Table for selecting approaches concerning the attributes Method, Tasks, Code availability, License and Triple store.}
  \begin{adjustbox}{totalheight=\textheight}
\begin{tabular}{l:cccc:c:l:l:l}
                                                                     & \multicolumn{4}{c:}{\textbf{Task}}                         &                                                           &                                                           &                                                           &                                                                                                                                                                   \\
      \multirow[b]{-2.9}{*}[5pt]{{\rotatebox[origin=c]{90}{\textbf{Method}}} } & {\rotatebox[origin=c]{90}{\textbf{CEA}}}                  & {\rotatebox[origin=c]{90}{\textbf{CPA}}}                  & {\rotatebox[origin=c]{90}{\textbf{CTA}}}                  & {\rotatebox[origin=c]{90}{\textbf{CNEA}}}                 &
      \multirow[b]{-4.2}{*}[8pt]{\rotatebox[origin=c]{90}{\minitab[p{40.96312pt}]{\textbf{Code \newline Available}}}} &
      \multirow[b]{-2}{*}{\textbf{Licence}}                             &
      \multirow[b]{-2}{*}{\textbf{Triple Store}}                        &
      \multirow[b]{-2}{*}{\textbf{References}}                                                                                                                                                                                                                                                                                                                                                                                                                                              \\
\end{tabular}
  \end{adjustbox}
        \label{tab:selection_table}
\end{table}

\end{document}